%% file: colm2026_conference.tex
\documentclass{article} % For LaTeX2e
\usepackage[final]{colm2026_conference}
\usepackage{microtype}
\usepackage{hyperref}
\usepackage{url}
\usepackage{booktabs}
\usepackage{multirow}
\usepackage{graphicx}
\usepackage{wrapfig}
\input{macros}
\usepackage{lineno}

\definecolor{darkblue}{rgb}{0, 0, 0.5}
\hypersetup{colorlinks=true, citecolor=darkblue, linkcolor=darkblue, urlcolor=darkblue}

\title{Combee: Scaling Parallel Prompt Learning for Self-Improving LLM Agents}

\author{Hanchen Li$^{1*}$, Runyuan He$^{1*}$, Qizheng Zhang$^{2}$, Changxiu Ji$^{2}$, Qiuyang Mang$^{1}$, \\\textbf{Xiaokun Chen$^3$, Lakshya A Agrawal$^{1}$, Wei-Liang Liao$^{1}$, Eric Yang$^4$, Alvin Cheung$^{1}$,}\\ \textbf{James Zou$^{2}$, Kunle Olukotun$^{2}$, Ion Stoica$^{1}$, Joseph E. Gonzalez$^{1}$}\\[0.2em]$^1$ UC Berkeley \quad $^2$ Stanford University \quad $^3$ Independent \quad $^4$ Gradient Network\\[0.2em]{\small $^*$ Equal contribution}}

\begin{document}

% \ifcolmsubmission
% \linenumbers
% \fi

\maketitle

\input{sections/abstract}
\input{sections/introduction_v2}
\input{sections/background_v3}
\input{sections/design_v4}
\input{sections/results}
\input{sections/related}

\tightsection{Conclusion}
% We presented Combee, a distributed framework for prompt learning at scale that enables parallel agents to efficiently acquire and consolidate knowledge during inference.
% Through the combination of parallel scan aggregation, augmented shuffling, and a dynamic batch size controller, Combee addresses the context overload problem that arises when naively scaling existing prompt learning methods.
% Our experiments on agentic benchmarks (AppWorld, Terminal-Bench 2.0) and domain-specific tasks (FiNER, Formula) demonstrate that Combee achieves significant speedups with comparable or improved quality, while reducing computational costs.
% We believe that context learning should evolve into an era of context learning at scale, and Combee represents a first step toward enabling this transition.
We presented Combee, a plug-and-play framework for scalable context learning that enables parallel agents to acquire and consolidate knowledge efficiently on top of existing prompt learning methods.
By combining parallel-scan aggregation, augmented shuffling, and dynamic batch-size control, Combee addresses the context overload issue that emerges when existing context learning methods are scaled naively.
Across agentic benchmarks (AppWorld and Terminal-Bench 2.0) and domain-specific tasks (FiNER and Formula), Combee substantially improves the quality--delay tradeoff, delivering large speedups at negligible cost variation while preserving most or all of the quality of sequential learning.
We believe prompt learning is entering a new era of scale, and Combee is a first step toward making that possible.

\section*{Ethics Statement}

This paper and the associated research do not raise any specific ethical concerns.

\section*{Reproducibility Statement}

We clearly describe the experimental setup used in our study, including the language models, datasets, and hyperparameters, so that readers with appropriate compute resources should be able to reproduce our results. 
All experiments in this paper are conducted on publicly available benchmarks. 
The source code will be released upon publication.

\section*{Acknowledgment}

We thank the anonymous reviewers and area chair for their constructive feedback, which greatly improved this paper. This research is supported by Accenture, AMD, Anyscale, Broadcom Inc., Google, IBM, Intel, Intesa Sanpaolo, Lambda, Mibura
Inc, Samsung SDS, SAP, NSF award CNS-2211384 and DARPA award TFAWIHR00112520038.
We also thank Matei Zaharia, Changran Hu and Yoonho Lee for discussions.

\bibliography{colm2026_conference}
\bibliographystyle{colm2026_conference}

\appendix
\input{sections/appendix}

\end{document}

%% file: macros.tex
\usepackage{xcolor}
\usepackage{colortbl}
\usepackage{tikz}

\definecolor{darkkhaki}{RGB}{189,183,107}
\definecolor{mygreen}{RGB}{0,128,0}

\newcommand{\mypara}[1]{\paragraph{#1}}

\newcounter{packednmbr}

\newenvironment{packeditemize}{\begin{list}{$\bullet$}{\setlength{\itemsep}{0.5pt}\addtolength{\labelwidth}{-4pt}\setlength{\leftmargin}{2ex}\setlength{\listparindent}{\parindent}\setlength{\parsep}{1pt}\setlength{\topsep}{2pt}}}{\end{list}}

\newcommand{\tightsection}[1]{\vspace{-0.25cm}\section{#1}\vspace{-0.15cm}}
\newcommand{\tightsubsection}[1]{\vspace{-0.35cm}\subsection{#1}\vspace{-0.08cm}}

\usepackage{tcolorbox}
  \usepackage{enumitem}
  \tcbuselibrary{skins,breakable}

  \definecolor{headerblue}{HTML}{1a2744}
  \definecolor{bulletblue}{HTML}{2563eb}
  \definecolor{countgray}{HTML}{888888}

  \newtcolorbox{playbookbox}[1]{
    enhanced, breakable,
    colback=white, colframe=headerblue,
    fonttitle=\bfseries\color{white}\normalsize,
    title={#1}, coltitle=white, colbacktitle=headerblue,
    boxrule=0.6pt, arc=1.5pt,
    left=5pt, right=5pt, top=3pt, bottom=3pt,
    toptitle=4pt, bottomtitle=4pt,
    before upper={\sloppy},
  }

  \newcommand{\pbsection}[1]{%
    \vspace{4pt}{\small\bfseries\color{headerblue}#1}\vspace{1pt}\par%
  }

%% file: sections/abstract.tex
\begin{abstract}
Recent advances in prompt learning allow large language model agents to acquire task-relevant knowledge from inference-time context without parameter changes.
For example, existing methods (like ACE or GEPA) can learn system prompts to improve accuracy based on previous agent runs.
However, these methods primarily focus on single-agent or low-parallelism settings.
This fundamentally limits their ability to efficiently learn from a large set of collected agentic traces. 
It would be efficient and beneficial to run prompt learning in parallel to accommodate the growing trend of learning from many agentic traces or parallel agent executions.
Yet without a principled strategy for scaling, current methods suffer from quality degradation with high parallelism.  
To improve the quality delay tradeoff of prompt learning, we propose Combee, a plug-and-play framework that scales parallel prompt learning on top of existing methods such as ACE and GEPA.
Combee speeds up learning and enables running many agents in parallel while preserving most of the quality of sequential learning from their aggregate traces.
To achieve this, Combee leverages parallel scans and employs an augmented shuffle mechanism; Combee also introduces a dynamic batch size controller to balance quality and delay. 
Evaluations on AppWorld, Terminal-Bench, Formula, and FiNER demonstrate that Combee substantially improves the quality delay tradeoff of prompt learning, achieving up to 17$\times$ speedup over previous methods at equivalent cost while matching or approaching their best accuracy.
\end{abstract}

% (1)Today's most agentic systems iterate sequentially (i.e., on the result from the previous iteration) or on small batches (i.e., on a few results from previous iterations). 
% (2) This significantly limits the ability to scale; when we try to scale batch sizes, they often fail.
% (3)In this paper, we address the problem of learning on very large batches. This would not only significantly speed up the learning, but also allow us to run many agent instances in parallel, and quickly learn from their aggregate traces. 

%% file: sections/introduction_v2.tex
\tightsection{Introduction}

\begin{figure}[h]
    \centering
    \includegraphics[width=\linewidth]{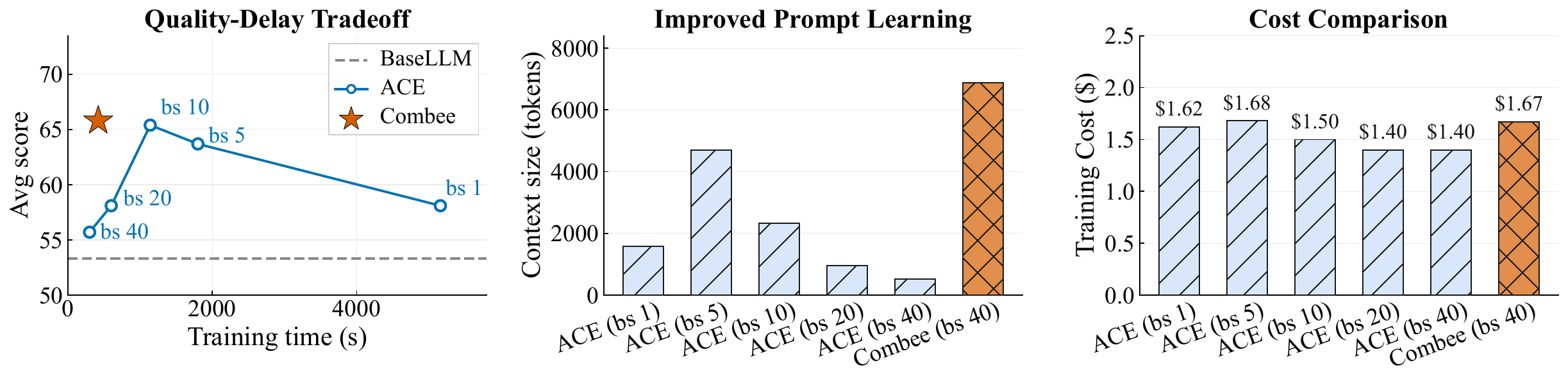}
    \caption{\textbf{Summary of improvement snapshot.} Combee achieves close-to-optimal quality with significantly reduced training time by increasing the content in the prompt learnt under high parallelism. Experiments with DeepSeek-V3.1 on AppWorld.} 
    \label{fig:intro}
\end{figure}

Large language models (LLMs) achieve strong performance in tasks such as mathematics and programming~\citep{lu2024mathvistaevaluatingmathematicalreasoning,jimenez2024swebenchlanguagemodelsresolve,agarwal2024many}. 
However, real-world problem solving usually requires learning from information that is only available at inference time~\citep{cl-bench, mang2025frontiercs}. 
This information is typically provided as \emph{context} (such as documents, examples, tool traces, or execution histories), serving as additional input at inference time that cannot be incorporated through offline training alone.

Recent work has shown that language agents can engage in \emph{prompt learning} to improve current or future task performance: extracting task-relevant knowledge from rich inference-time inputs (trajectories, documents, tool traces) and consolidating it into reusable artifacts such as playbooks or rules, without any weight updates~\citep{khattab2024dspy,ace,gepa,reflexion,voyager}.
% Unlike standard in-context learning~\citep{brown2020languagemodelsfewshotlearners,cl-bench}, which pattern-matches from a handful of examples, prompt learning distills and retains knowledge across interactions.
For example, ACE~\citep{ace} enables agents to adapt during inference by consolidating experience into structured playbooks, while GEPA~\citep{gepa} optimizes prompts based on performance feedback from contextual examples.
These approaches demonstrate that inference-time context can serve as a powerful learning medium without updating parameters.

% However, existing prompt learning methods (including ACE and GEPA) were designed for sequential, single-agent settings and provide no principled strategy for parallel scaling. 
However, existing prompt learning methods (including ACE and GEPA) were designed around sequential or low-parallelism updates, where one or a small number of trajectories are reflected on and consolidated at a time, and thus provide no principled strategy for scaling the reflection-and-aggregation step to high parallelism.
This is increasingly limiting: as agentic systems grow in scale, agents produce large volumes of interaction traces~\citep{wang2024openhands,zhao2024expel,yang2024swe} that ideally would be learned from concurrently, and parallel multi-agent deployments are becoming standard practice in academia~\citep{li2024more,hong2023metagpt,qian2024chatdev} and industry~\citep{cursor_scaling_agents_2026, anthropic2026ccompiler}.
Yet naively increasing parallelism creates a bottleneck: the aggregator LLM responsible for consolidating many reflections must process increasingly long-horizon reflective context at once, and becomes overwhelmed.
We refer to this \emph{context overload}.
Concretely (§\ref{sec:failure}), scaling from batch 1 to batch 100 on the Formula dataset~\citep{wang2025finlora} drops accuracy from 87.0\% to 72.5\%, and qualitative analysis reveals that the aggregator only retained generic patterns while discarding the specific, high-value entries that drive downstream performance. 
Prompt-level mitigations such as summarization and top-K retrieval do not resolve this (\S\ref{sec:results}).
This bottlenecks both efficient learning from large-scale agent traces and timely adaptation in parallel agent scenarios~\citep{snell2024scaling,li2024more}.

To address this problem, we propose Combee, a distributed framework for scalable prompt learning. 
Importantly, Combee is not a new prompt learning method competing with ACE or GEPA: it is a \emph{plug-and-play scaling layer} that sits on top of an existing generate-reflect-update method and improves its quality delay tradeoff under parallelism, leaving the base method's generation and update mechanisms unchanged (\S\ref{sec:design}).
Combee adopts a Map-Shuffle-Reduce paradigm: multiple agents process distinct context shards in parallel (Map), reflections are duplicated and shuffled to prevent information loss (Shuffle), and a hierarchical parallel scan algorithm aggregates local updates into a coherent global context without overloading the LLM context curator (Reduce). 
A dynamic batch size controller further balances quality and training delay automatically across iterations. 
We prototype Combee on both ACE and GEPA, and expect it to generalize to other generate-reflect-update frameworks with minimal changes.
Evaluations on agentic benchmarks, AppWorld~\citep{appworld} and Terminal-Bench~\citep{merrill2026terminal} and domain-specific tasks, FiNER~\citep{loukas2022finer} and Formula~\citep{wang2025finlora} demonstrate that Combee substantially improves the quality delay tradeoff of prompt learning: it achieves up to 17$\times$ speedup over baselines at equivalent cost while preserving most or all of the quality of slow sequential learning.

To summarize, our contributions are to:
\begin{packeditemize}
    \item Identify the problem of efficient scaling for prompt learning and failure of previous methods (\S\ref{sec:failure}).
    \item Design Combee, a plug-and-play framework for scalable prompt learning featuring parallel scan aggregation, augmented shuffling, and a dynamic batch size controller (\S\ref{sec:design}). 
    \item Prototype Combee on top of ACE and GEPA, and expect it to generalize to other generate-reflect-update frameworks with minimal changes.
    \item Perform evaluations on Terminal-Bench 2.0, AppWorld, Formula, and FiNER to show that Combee substantially improves the quality delay tradeoff, achieving up to 17$\times$ speedup at equivalent cost while preserving most or all of the quality of sequential learning (\S\ref{sec:results}).
\end{packeditemize}

% We implemented prototypes of the proposed framework on top of Agentic Context Engineering (ACE)~\citep{ace} and GEPA~\citep{gepa}.
% Through our evaluation on benchmarks in agent tasks~\citep{appworld}, finance~\citep{loukas2022finer, wang2025finlora}, 
% we show that Combee can significantly improve the efficiency and scalability of prompt learning.
% Our results show that Combee achieves up to 13$\times$ speedup with similar quality or 3x higher quality gain from prompt learning. Moreover, Combee is cost efficient as well, achieving 30\% training cost reduction over the original baselines due to reduced repetitive prompt generation steps.

% In conclusion, we argue that prompt learning should transform into the era of prompt learning at scale. By taking the first step to adapt prompt learning system designs for high parallelism, Combee improves upon previous work and opens doors for future research. %% TODO: fill in specific improvement numbers

%% file: sections/background_v3.tex
\tightsection{Background and Motivation}
\label{sec:background}

\subsection{Prompt Learning}
Prompt learning is an emerging inference-time learning paradigm in which an agent extracts task-relevant knowledge from rich inputs (execution trajectories, tool traces, documents) and consolidates it into reusable artifacts such as playbooks, memories, or skill libraries that improve current or future performance without any weight updates~\citep{ace,gepa,dc,asi}. 
% Unlike standard in-context learning (ICL), which pattern-matches from a handful of examples~\citep{agarwal2024many}, prompt learning distills and retains knowledge across interactions, enabling agents to tackle complex tasks beyond what was seen during pre-training~\citep{snell2024scaling, li2026skillsbench}.

In this work, we focus on methods that follow a \emph{generate-reflect-update} loop: an agent executes a task, reflects on its trajectory to extract useful insights, and updates a shared context artifact for future iterations. 
Our goal is to scale this loop to high parallelism, spinning up multiple agents concurrently per iteration, while preserving the quality of the resulting context updates. 
Rather than proposing yet another method in this family, Combee operates as a scaling layer on top of any such method: it intervenes only between the reflection and update steps, and we specify the minimal interface a base method must expose in \S\ref{sec:design}.
This paradigm is broadly adopted across skill libraries~\citep{voyager,asi,memskill,skillrl} and evolving memories or playbooks~\citep{ace,gepa,zhao2024expel,dc,reasoningbank,reflexion,memento}, confirming that scaling generate-reflect-update is a natural and important problem.
Unlike offline prompt engineering~\citep{wei2022chain,zhou2022large}, prompt learning here treats the context artifact as a living object that evolves during deployment.

\subsection{The Problem: Context Overload from Naive Parallel Scaling}
\label{sec:failure}

\begin{figure}[t!]
    \centering
    \includegraphics[width=0.9\linewidth]{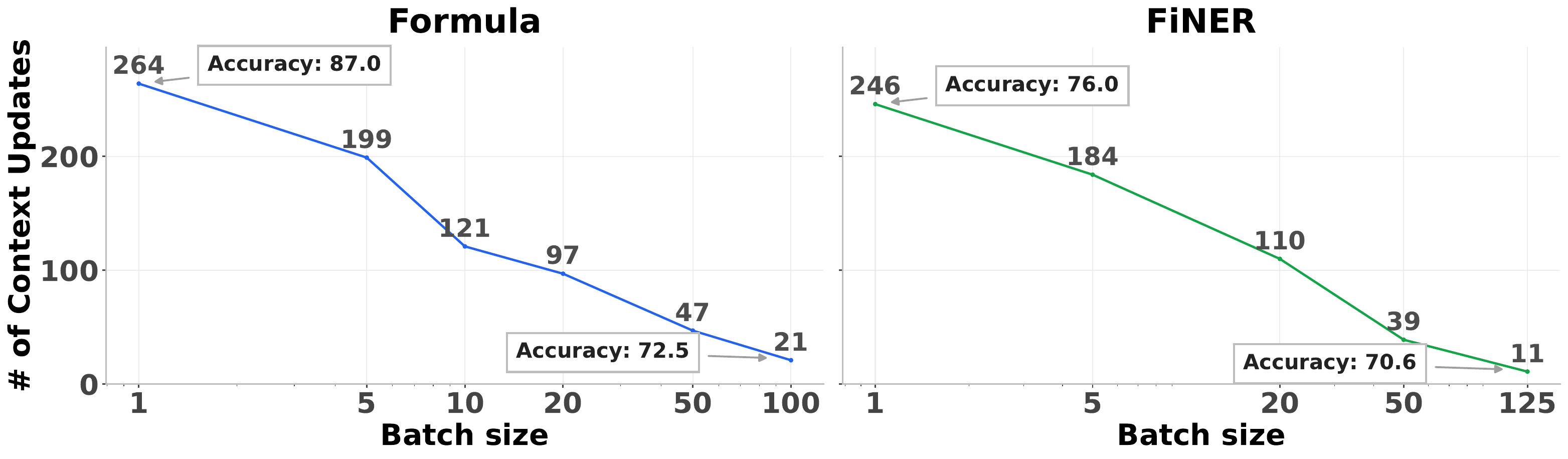}
    \caption{\textbf{Context overload from naive scaling.} As batch size increases, the aggregator LLM produces monotonically fewer and less useful context updates, directly degrading final accuracy across benchmarks.}
    \vspace{-10pt}
    \label{fig:naive_failing}
\end{figure}

% \qizheng{the current version lacks "why" context overload happens}

% \qizheng{if space allows, add a trade-off figure on accuracy v.s. level of parallelism}\lhc{dont we already have that?}

A natural approach to scale prompt learning for the general \emph{generate-reflect-update} paradigm mentioned above is to increase the \emph{batch size} of reflections before generating context updates, which can aggregate more feedback signals before updating the context. 
Here, \emph{batch size} refers to the number of parallel agent trajectories or reflections aggregated before producing one context update in an iteration.
This is appealingly simple, but fails in practice due to a phenomenon we call \emph{context overload}: as batch size grows, the aggregator LLM must distill an increasingly large volume of reflections into a single context update, producing far fewer and lower-quality entries. 
Critically, this degradation occurs even when all reflections fit within the model's context window (we use DeepSeek-V3.1 with 128K context), ruling out simple truncation as the cause. 
Instead, the aggregator appears to perform \textbf{lossy compression}: when presented with many reflections simultaneously, it defaults to retaining broad, generic patterns while discarding the specific, high-value insights that disproportionately drive downstream accuracy.

\mypara{Quantitative Evidence} Figure \ref{fig:naive_failing} demonstrates the information loss from naive scaling across Formula (numerical reasoning)~\citep{wang2025finlora} and FiNER (financial entity recognition)~\citep{loukas2022finer}. 
In both settings, the number of context updates drops monotonically with batch size: on Formula from 264 (batch 1) to 21 (batch 100), on FiNER from 246 to 11. 
Accuracy follows the same trend: Formula drops from 87.0\% to 72.5\%, FiNER from 76.0\% to 70.6\%. 
The same pattern holds on agentic tasks: on AppWorld~\citep{appworld}, scaling from batch 1 to batch 40 reduces accuracy from 58.1 to 55.7, approaching the no-context-learning baseline of 53.3 (Table \ref{tab:appworld_main}).

\mypara{Qualitative Evidence} The degradation goes beyond quantity. 
In ACE, the final system prompt learnt is a playbook with many entries. Each entry is marked helpful (h) or harmful (r) during inference, providing a measure of entry utility. 
Under sequential learning (batch 1), the Formula playbook accumulates 174 total helpful hits across 264 entries, with 19 entries reaching $h\ge3$ and a maximum of $h=16$; the FiNER playbook accumulates 331 helpful hits across 246 entries, with 38 entries reaching $h\ge3$. 
Under naive scaling, these high-value entries vanish entirely: the batch~100 Formula playbook retains zero entries with $h\ge3$ (total helpful hits: 5), and the batch~125 FiNER playbook retains zero (total helpful hits: 4). 
Appendix \ref{app:overload_examples} provides concrete playbook snapshots illustrating how task-specific strategies (e.g., formula-specific edge-case handling, precise rounding protocols) \textbf{collapse} into \textbf{generic reminders} under high parallelism.

% \mypara{Naive Fix} One might hypothesize that context overload can be addressed by improving the aggregation prompt.
% For example, instructing the aggregator to produce more entries, or process reflections via summarization or top-K retrieval before aggregation. 
% We evaluated both strategies in \S\ref{sec:results} and find that these methods fail to recover the quality of sequential learning: summarization and top-K retrieval baselines consistently underperform Combee and often even naive batched ACE (Figures \ref{fig:gepa_fig} and \ref{fig:ace_fig}). 
% This confirms that context overload is not a prompting artifact but a structural limitation of single-pass aggregation over large reflection sets.

This reveals a fundamental tension: increasing parallelism reduces wall-clock training time, but naive aggregation destroys the fine-grained knowledge that makes prompt learning effective. 
The sweet spot for naive scaling, small batch sizes that partially avoid overload, yields only modest speedups, while the large batch sizes needed for meaningful acceleration collapse quality toward the no-context-learning baseline.

%% file: sections/design_v4.tex
\begin{figure}[t]
\centering
\includegraphics[width=0.9\textwidth]{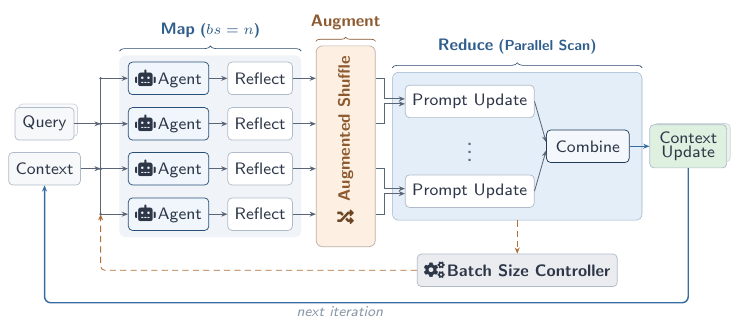}
\caption{\textbf{Overall design of Combee.} Map-Shuffle-Reduce pipeline: parallel reflection (Map), augmented shuffling (Shuffle), and parallel scan aggregation (Reduce). Combee plugs into existing methods (e.g., ACE or GEPA) between reflection and update; naive scaling aggregates all reflections in one pass.}
\label{fig:design}
\vspace{-4pt}
\end{figure}

\tightsection{Design of Combee}
\label{sec:design}

We present Combee, a framework that enables scalable prompt learning through parallel generation and adaptation.
Combee extends generate-reflect-update methods~\citep{ace, gepa, li2024more} to high parallelism via three components: \emph{parallel scan aggregation}, \emph{augmented shuffling}, and a \emph{dynamic batch size controller} (Figure~\ref{fig:design}).

\mypara{Interface with Downstream Prompt Learning Methods}
Combee is a scaling layer, not a standalone prompt learning method. 
It requires only that the base method follow a generate-reflect-update loop producing \emph{textual reflections} aggregated into a shared context artifact; Combee replaces the single aggregation call with its shuffle-and-scan pipeline and returns the consolidated update (Figure~\ref{fig:design}). 
We prototype on ACE (curation-based) and GEPA (optimization-based) because they cover the two dominant update mechanisms.

\tightsubsection{Parallel Scan Aggregation}
\label{sec:scan}
One core design inside Combee is the \emph{parallel scan aggregation} algorithm, 
which is designed to efficiently aggregate learning experience from multiple parallel trajectories while avoiding the context overload problem observed in \S\ref{sec:failure}.

In order to solve this problem, Combee employs a multi-level parallel scan algorithm to aggregate learning experience from multiple trajectories in a way that prevents overloading the aggregator.
Given the $n$ generated trajectories, Combee can choose to first separate them into $k$ mini-batches, each containing $n/k$ trajectories.
Instead of directly feeding all the reflections from the $n$ trajectories to the aggregator, 
Combee can first aggregate the reflections within each mini-batch of trajectories into context updates.
Then, Combee can further aggregate the $k$ context updates into a single update for this round. 
Conceptually, this is similar to the parallel scan algorithm used in parallel computing for performing prefix sum operations~\citep{blelloch1990prefix} and recently adopted in sequence modeling~\citep{gu2024mamba}. This approach also draws inspiration from MapReduce-style decomposition for LLM processing of long documents~\citep{zhou2024llm}.
Parallel scan is a natural fit for our setting: it is the classical technique for combining a sequence under an associative operator with logarithmic rather than linear dependency depth, and we treat reflection aggregation as the combining operator.
This structurally bounds the number of reflections any single aggregator call must process, directly targeting the lossy compression identified in \S\ref{sec:failure}; prompt-level mitigations that keep single-pass aggregation (summarization, retrieval, or instructing the aggregator to be specific) do not achieve the same effect (\S\ref{sec:results}).
By default, Combee sets $k$ equal to $\lfloor\sqrt{n}\rfloor$ so that each level of the aggregation tree processes similar count of entries
: the first level generates context updates based on  $n/k = \sqrt{n}$ reflections per group, 
while the second level aggregates $k = \sqrt{n}$ updates.
We note that the empirically optimal mini-batch size is not always exactly $\sqrt{n}$, but it consistently lies close to the optimum in our experiments (Figure~\ref{fig:worker_ablation}); we therefore adopt $\sqrt{n}$ as a principled approximation that avoids exhaustively searching the mini-batch dimension, consistent with Combee's goal of reducing learning time with minimal quality loss.

\tightsubsection{Augmented Shuffling}
Popular context engineering methods, including GEPA~\citep{gepa} and ACE~\citep{ace}, all incorporate reflection steps to extract past insights from rollouts. 
These reflections usually have higher information density: 
although they consist of a small number of tokens, they contain crucial information necessary for the agent's improvement. 
To fully leverage this dense information during parallel learning without losing vital insights, 
Combee introduces an \emph{augmented shuffling} mechanism. 
Specifically, given a set of $x$ generated reflections, 
Combee duplicates each reflection $p$ times (default $p=2$) and shuffles the augmented set before issuing them to the worker nodes. 
By giving each reflection multiple opportunities to contribute to the learning process, echoing the principle behind self-consistency~\citep{wang2022self}, Combee increases the chances that the aggregator can learn from the reflections even under large batch sizes.
Duplication also exposes each reflection to multiple different groupings, so co-dependent insights are more likely to be extracted together rather than lost to a single unlucky partition or ordering. 
This improves the robustness of the parallel learning pipeline despite increased batch size, at low overhead: our ablation in \S\ref{sec:results} (Table~\ref{tab:dup_ablation}) shows that the benefit is almost fully captured at $p=2$, while further duplication mostly adds redundant copies that crowd out distinct reflections within each mini-batch and yields diminishing or even negative returns.

\tightsubsection{Dynamic Batch Size Controller}
Parallel scan aggregation and augmented shuffling ensure that learning quality is maintained across a wide range of batch sizes.
The batch size selection therefore primarily reduces to a speed question: as the batch size increases, per-epoch delay decreases (more samples are processed in parallel); but with diminishing returns, analogous to the critical batch size concept from distributed training~\citep{mccandlish2018empirical}.
That said, excessively large batch sizes may still degrade learning quality, so we would like to stay within a reasonable range.
We therefore select the largest batch size that still yields meaningful delay reduction, while enforcing an upper bound to avoid unnecessary risk of quality degradation~\citep{smith2017don,goyal2017accurate}.

To find this point, we profile the delay by running trial iterations at a set of default candidate batch sizes $\{bs_1, bs_2, \ldots, bs_k\}$.
For each candidate $bs_i$, we run one iteration to measure the delay $d(bs_i)$ and convert it to estimated epoch time:
\[
T_{\mathrm{epoch}}(bs) = d(bs) \cdot \frac{N_{\mathrm{train}}}{bs},
\]
where $N_{\mathrm{train}}$ is the training set size. We fit a power-law delay curve through measurements:
\[
T_{\mathrm{epoch}}(bs) = A \cdot bs^{-\alpha}.
\]

Given the fitted curve, we select the batch size at which the marginal delay reduction falls below a fixed threshold $\tau$\footnote{In our experiments, 
we set $\tau$ to $1.6\%$ of the peak slope size. This means we stop increasing batch size once each new unit reduces epoch time by less than $1.6\%$ of the steepest improvement rate.}.
Solving $\left|\frac{dT_{\mathrm{epoch}}}{d\,bs}\right| = \tau$ yields:
\[
\mathrm{plateau\_bs} = \left(\frac{\alpha A}{\tau}\right)^{\frac{1}{\alpha+1}}.
\]
% clamped to the range $[bs_{\min},\, Upper\_Bound\_Const(200)]$.

We emphasize that the controller's goal is \emph{not} to maximize the batch size: larger batches are faster but degrade quality beyond a point.
Instead, it conservatively locates a sweet spot on the quality delay tradeoff, choosing a batch size near the largest one that captures most of the available speedup before quality starts to fall; $\tau$ is deliberately set conservatively so the controller stops before measurable quality loss. 
We use a single profiling iteration per candidate batch size because additional profiling runs would consume a meaningful fraction of the training budget on our relatively small training sets; larger-volume deployments can afford multiple iterations for more stable estimates.

%% file: sections/results.tex
\tightsection{Results}
\label{sec:results}
\definecolor{aceblue}{HTML}{0072B2}
\definecolor{aceorange}{HTML}{D55E00}
\definecolor{acegray}{HTML}{888888}

Our main takeaways from evaluating Combee are:
\begin{packeditemize}
    \item Combee integrates plug-and-play with existing prompt learning methods such as ACE and GEPA to enable efficient learning at scale, substantially improving the quality delay tradeoff: it preserves most or all of the quality of sequential learning while reducing training time by up to 17$\times$.
    \item Combee's specialized design including parallel scan aggregation and augmented shuffle 
    prevents context overload and improves on previous parallel methods.
    \item Combee remains robust across different models, tasks, and learning settings with cost comparable to previous methods.  
\end{packeditemize}

% \vspace{-5pt}
\subsection{Experiment Setup}
\mypara{Tasks and Datasets} We evaluate Combee on agentic and domain-specific benchmarks. 
\begin{packeditemize}
    \item \textbf{Agentic Benchmarks:} 
    AppWorld~\citep{appworld} evaluates multi-step API tasks via Task Goal Completion (TGC) and Scenario Goal Completion (SGC). 
    We reuse the training set of 90 tasks and evaluate on the held out Test-Normal dataset.
    Terminal-Bench 2.0~\citep{merrill2026terminal} contains 89 command-line tasks testing software engineering capabilities. We train on 60 Deepseek 3.2 trajectories released on huggingface~\citep{lee2026terminalbench_trajectories} 
    and evaluate average Accuracy@1 across three runs on 29 held-out tasks.
    \item \textbf{Domain-Specific Benchmarks:} We use two finance NLP datasets: FiNER~\citep{loukas2022finer} for fine-grained entity typing in XBRL documents, and Formula~\citep{wang2025finlora} for numerical reasoning over structured filings.
\end{packeditemize}

\mypara{Frameworks and Baselines}
For majority of experiments, we use \texttt{DeepSeek-V3.1} provided by Together AI as the base LLM.
Combee is agnostic to the base prompt learning method: We implement Combee on top of two prompt learning: ACE~\citep{ace}, which accumulates strategies into text-based playbooks, and GEPA~\citep{gepa}, which optimizes system prompts via evolutionary search. Both follow a generate-reflect-update loop that Combee extends for parallel scaling. 
We also compared with three aggregation baselines on top of ACE and GEPA: Top-K Retrieval, Summarization, and Hierarchical Summarization. 
Top-K Retrieval embeds reflections, clusters them into K groups, and feeds one representative reflection from each group to the curator. 
We use clustering rather than standard query-based retrieval because, unlike RAG, the context update step has no external query to rank candidates against; selecting one representative per cluster yields a diverse subset of reflections in this query-less setting. 
Summarization summarizes reflections before feeding them into the curator. 
Hierarchical Summarization first summarizes each reflection individually, then further condenses these summaries with a second summarizer before passing the result to the curator, using the same batch size as Combee.

\subsection{Results on Agent Benchmarks}

\begin{table}[tbh!]
\centering
\setlength{\tabcolsep}{6pt}
\renewcommand{\arraystretch}{1.15}

\resizebox{\columnwidth}{!}{%
\begin{tabular}{l c c c c c c c}
\toprule
\textbf{Method}  & \textbf{Batch}
& \textbf{Playbook}
& \textbf{Training}
& \textbf{Training}
& \multicolumn{3}{c}{\textbf{Test-Normal}} \\
\cmidrule(lr){6-8}
 & \textbf{Size}
  & \textbf{Size (tokens)}
& \textbf{Time (min)}
& \textbf{Cost}
& \textbf{TGC$\uparrow$} & \textbf{SGC$\uparrow$} & \textbf{Avg$\uparrow$} \\
\midrule

ReAct
& --
& --
& 0
& \$0
& 63.7
& 42.9
& 53.3 \\

ReAct + ACE
& 1
& 1,578
& 86
& \$1.62
& 66.1
& 50.0
& 58.1 \\

\midrule
\rowcolor{gray!12}
\multicolumn{8}{c}{\textbf{Parallel Prompt Learning}} \\
\midrule

ReAct + ACE
& 5
& 4,697
& 30
& \$1.68
& 70.2
& 57.1
& 63.7 \\

ReAct + ACE
& 10
& 2,329
& 19
& \$1.50
& 72.0
& 58.9
& 65.4 \\

ReAct + ACE
& 20
& 954
& 10
& \$1.40
& 67.9
& 48.2
& 58.1 \\

ReAct + ACE
& 40
& 526
& 5
& \$1.40
& 66.7
& 44.6
& 55.7 \\

ReAct + Combee
& 40
& 6,887
& 7
& \$1.67
& 70.8
& 60.7
& 65.8 \\

\bottomrule
\end{tabular}%
}

\caption{
\textbf{Parallel prompt learning results with ReAct agent for AppWorld.}
}
\label{tab:appworld_main}
\end{table}

Table~\ref{tab:appworld_main} shows results on AppWorld.
Naive parallel scaling trades quality for speed: batch~10 is the sweet spot, but batch~40 drops near the no-context-learning baseline.
Combee at batch~40 achieves the highest average score and SGC with a 12$\times$ speedup over sequential ACE, retaining 6,887 playbook tokens versus 526 for naive batch~40.

Table~\ref{tab:terminal_bench} shows the same pattern on Terminal-Bench 2.0: Combee at batch~30 recovers most sequential quality (35.6\% vs.\ 37.9\%) with over 17$\times$ lower training time and a playbook size much closer to sequential ACE (8,023 vs.\ 9,067 tokens).
Appendix~\ref{app:cost_breakdown} reports a per-stage cost breakdown showing only 12 extra LLM calls and roughly \$0.04 additional curation cost.

\begin{table}[t!]
\centering
\setlength{\tabcolsep}{6pt}
\renewcommand{\arraystretch}{1.15}

\resizebox{\columnwidth}{!}{%
\begin{tabular}{l c c c c c}
\toprule
\textbf{Method}  & \textbf{Batch} & \textbf{Playbook}
& \textbf{Training}
& \textbf{Training}
& \textbf{Average} \\
 & \textbf{Size} & \textbf{Size (tokens)}
& \textbf{Time (min)}
& \textbf{Cost}
& \textbf{Accuracy @ 1}\\
\midrule

Terminus-2
& --
& --
& 0
& \$0
& 32.2\% \\

Terminus-2 + ACE
& 1
& 9,067
& 42.4
& \$0.24
& 37.9\% \\

\midrule
\rowcolor{gray!12}
\multicolumn{6}{c}{\textbf{Parallel Prompt Learning}} \\
\midrule

Terminus-2 + ACE
& 5
& 4,983
& 10.2
& \$0.17
& 29.9\% \\

Terminus-2 + ACE
& 10
& 3,967
& 5.6
& \$0.15
& 33.3\% \\

Terminus-2 + ACE
& 30
& 3,150
& 2.1
& \$0.13
& 31.0\% \\

Terminus-2 + Combee 
& 30
& 8,023
& 2.4
& \$0.17
& 35.6\% \\

\bottomrule
\end{tabular}%
}
\caption{
\textbf{Parallel prompt learning results with Terminus-2 agent for Terminal-Bench 2.0.} 
We trained on existing open-source traces instead of generating trajectories on the fly. We report the average accuracy over three runs.
}
\label{tab:terminal_bench}
\end{table}

% \vspace{-5pt}

\subsection{Results on Domain Specific Benchmarks}
\begin{figure}[tbhp!]
    \centering
    % ---- legend rows ----
    \footnotesize
    \begin{minipage}{\linewidth}\centering
    \tikz[baseline=-0.6ex]{
        \draw[gray,dashed,thick] (0,0)--(1.6em,0);
    }\ Base LLM \quad
    \tikz[baseline=-0.6ex]{
        \draw[color={rgb,255:red,17;green,119;blue,51},thick,dash dot] (0,0)--(1.6em,0);
    }\ Summarization + GEPA \quad
    \tikz[baseline=-0.6ex]{
        \draw[color={rgb,255:red,136;green,34;blue,85},thick,dotted] (0,0)--(1.6em,0);
    }\ TopK + GEPA \\[3pt]
    \tikz[baseline=-0.6ex]{
        \draw[color={rgb,255:red,136;green,204;blue,238},thick,dashed] (0,0)--(1.6em,0);
    }\ Hierarchical Summarization + GEPA \quad
    \tikz[baseline=-0.6ex]{
        \draw[color={rgb,255:red,0;green,114;blue,178},thick] (0,0)--(1.6em,0);
        \draw[color={rgb,255:red,0;green,114;blue,178},fill=white,thick]
            (0.8em,0) circle (0.18em);
    }\ GEPA \quad
    \textcolor[HTML]{D55E00}{$\star$}\ Combee + GEPA
    \end{minipage}

    \includegraphics[width=0.8\columnwidth]{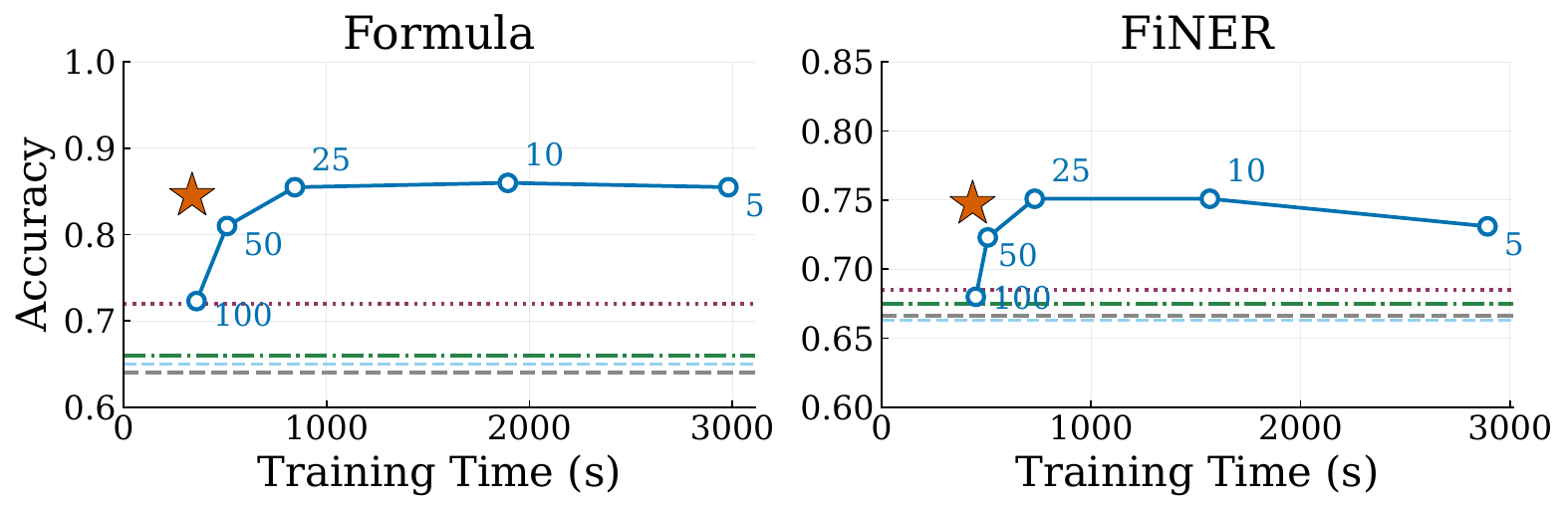}
    \caption{Combee achieves superior quality-delay trade off on GEPA for finance benchmarks.}
    \label{fig:gepa_fig}
\end{figure}

\begin{figure}[tbhp!]
    \centering
    % ---- legend rows ----
    \footnotesize
    \begin{minipage}{\linewidth}\centering
    \tikz[baseline=-0.6ex]{
        \draw[gray,dashed,thick] (0,0)--(1.6em,0);
    }\ Base LLM \quad
    \tikz[baseline=-0.6ex]{
        \draw[color={rgb,255:red,17;green,119;blue,51},thick,dash dot] (0,0)--(1.6em,0);
    }\ Summarization + ACE \quad
    \tikz[baseline=-0.6ex]{
        \draw[color={rgb,255:red,136;green,34;blue,85},thick,dotted] (0,0)--(1.6em,0);
    }\ TopK + ACE \\[3pt]
    \tikz[baseline=-0.6ex]{
        \draw[color={rgb,255:red,136;green,204;blue,238},thick,dashed] (0,0)--(1.6em,0);
    }\ Hierarchical Summarization + ACE \quad
    \tikz[baseline=-0.6ex]{
        \draw[color={rgb,255:red,0;green,114;blue,178},thick] (0,0)--(1.6em,0);
        \draw[color={rgb,255:red,0;green,114;blue,178},fill=white,thick]
            (0.8em,0) circle (0.18em);
    }\ ACE \quad
    \textcolor[HTML]{D55E00}{$\star$}\ Combee + ACE
    \end{minipage}

    \includegraphics[width=0.8\columnwidth]{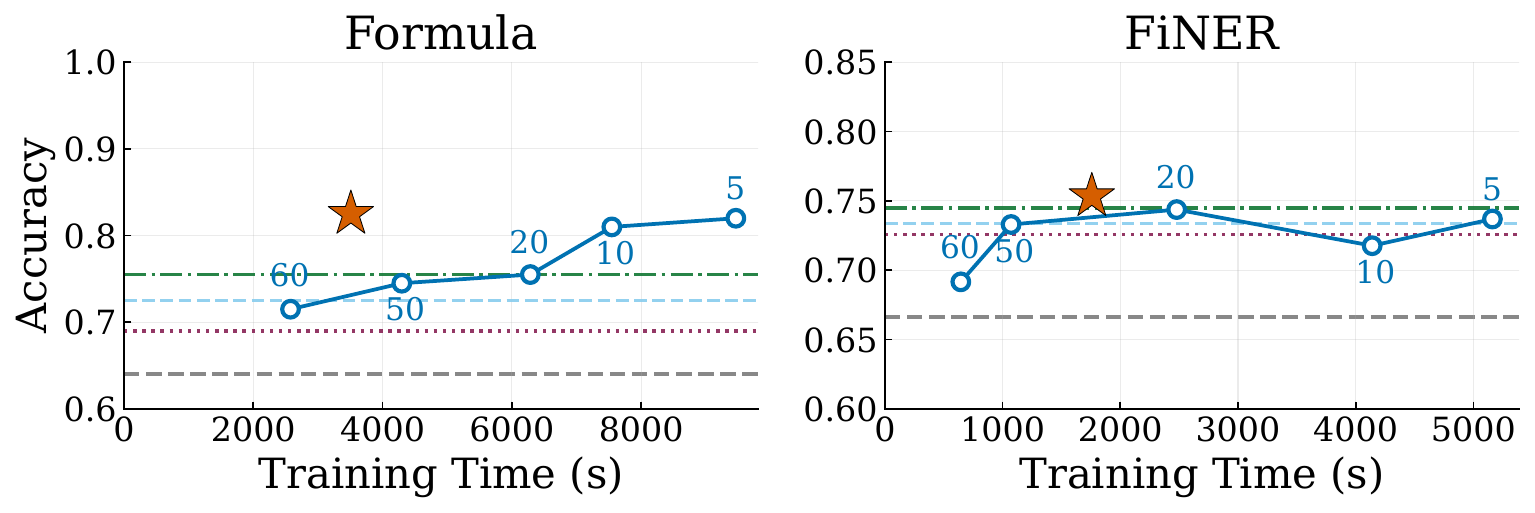}
    \caption{Combee achieves superior quality-delay trade off on ACE for finance benchmarks.}
    \label{fig:ace_fig}
    % \vspace{-5pt}
\end{figure}
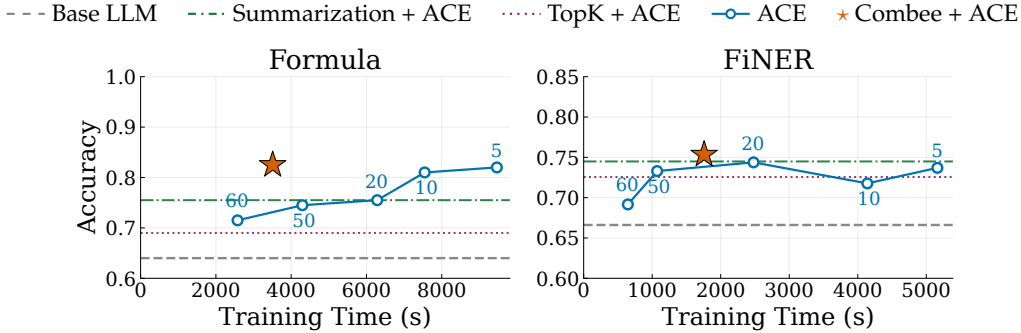

Figures~\ref{fig:gepa_fig} and~\ref{fig:ace_fig} show results on Formula and FiNER with GEPA and ACE.
Since there are many training samples in Formula (250) and FiNER (500), we use the dynamic batch size controller, which selects batch sizes of 84 and 94 respectively.
Summarization and Top-K baselines use batch size 50; hierarchical summarization uses Combee's batch size.

The same quality--speed trade-off from the agent benchmarks persists across both tasks and both frameworks: small batch sizes yield higher accuracy but require long training times, while large batch sizes are fast but suffer from context overload: on GEPA FiNER, batch~100 even falls below the base LLM.
Combee consistently reaches the Pareto frontier, matching or exceeding the best fixed-batch accuracy while training significantly faster than quality-matching setups.
With GEPA (Figure~\ref{fig:gepa_fig}), Combee matches the best fixed-batch accuracy on FiNER and achieves competitive accuracy on Formula with less than half of the time by fixed-batch baseline.
With ACE (Figure~\ref{fig:ace_fig}), Combee achieves the highest accuracy on Formula and FiNER, while training more than 2.4$\times$ faster than the quality-comparable baselines.
The Top K and Summarization baselines achieved much worse generation quality compared with Combee or naive ACE methods.
Appendix~\ref{app:full_stats} reports the exact values for all plotted points, together with multi-run statistics showing that these improvements are robust rather than within noise.
These results confirm that Combee's design is framework-agnostic and effective for domain-specific tasks.

\mypara{Hierarchical Summarization Baseline} We also compare Combee against the stronger two-level hierarchical summarization baseline at matched batch size; Combee outperforms it by wide margins on both datasets and both frameworks (Formula: 0.857 vs. 0.650 with GEPA and 0.855 vs. 0.725 with ACE; FiNER: 0.742 vs. 0.663 with GEPA and 0.770 vs. 0.734 with ACE).
This highlights that hierarchical structure alone is not sufficient: reflections are information-rich, and the core difficulty is distilling them into task-specific, concise updates rather than generic ones.
Compressing reflections into summaries before curation discards specificity at the first level, whereas Combee's parallel scan keeps every reflection intact while bounding the per-call load, and augmented shuffling protects co-dependent insights from unlucky partitions.

\tightsubsection{Extended Analysis}
We conduct ablation study and robustness analysis on the Formula dataset.
Combee is also robust across model families: evaluated on top of GPT-OSS 120B, it follows the same pattern of superior quality over fixed-batch baselines with much reduced training time (Appendix~\ref{app:gptoss}).

\begin{figure}[t]
    \centering
    \begin{minipage}{0.48\columnwidth}
        \centering
        \footnotesize
        \tikz[baseline=-0.6ex]{\draw[acegray,dashed,thick] (0,0)--(1.5em,0);}\ Baseline\quad
        \tikz[baseline=-0.6ex]{
            \draw[aceblue,thick] (0,0)--(1.5em,0);
            \draw[aceblue,fill=white,thick] (0.75em,0) circle (0.16em);
        }\ Normal parallel\quad
        \textcolor[HTML]{D55E00}{$\star$}\ Combee \quad
        \includegraphics[width=\linewidth]{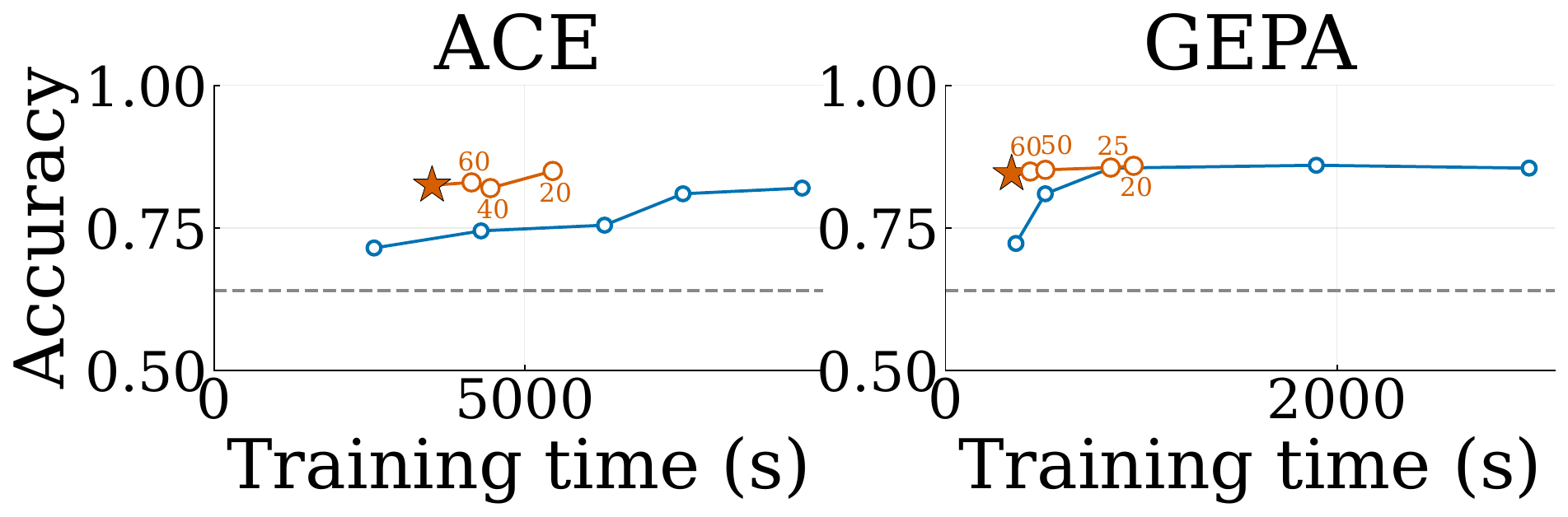}
        \caption{Combee's dynamic batch size adaptation saves delay besides maintaining high accuracy}
        \vspace{-10pt}
        \label{fig:bs_ablation}
    \end{minipage}\hfill
    \begin{minipage}{0.48\columnwidth}
        \centering
        {\footnotesize
        \tikz[baseline=-0.6ex]{
            \draw[aceorange,thick] (0,0)--(1.5em,0); 
            \draw[aceorange,fill=white,thick] (0.75em,0) circle (0.16em);
        }\ w/ augmented shuffling \\
        \tikz[baseline=-0.6ex]{
            \draw[aceblue,thick,dashed] (0,0)--(1.5em,0);
            \draw[aceblue,fill=white,thick] (0.7em,-0.1em) rectangle (0.7em+0.2em,0.1em);
        }\ w/o augmented shuffling
        \par\vspace{4pt}}
        \includegraphics[width=\linewidth]{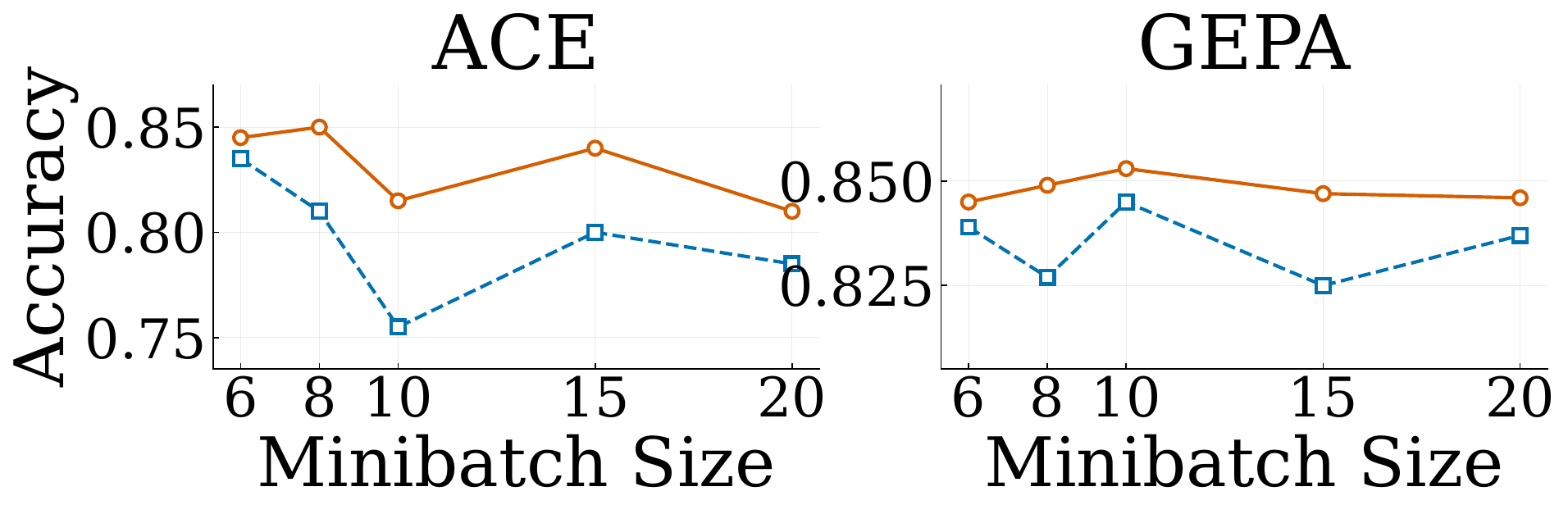}
        \caption{Combee's augmented shuffling improves learning robustness across mini-batch sizes.}
        \vspace{-10pt}
        \label{fig:worker_ablation}
    \end{minipage}
\end{figure}

Figure~\ref{fig:bs_ablation} ablates the dynamic batch size controller by comparing Combee against a variant that uses a fixed batch size throughout training on the Formula dataset.
Without a dynamic controller, Combee with a fixed batch size may choose a necessarily small batch size, causing a delay increase with little quality change. This demonstrates the effectiveness of the batch size controller of Combee.

Figure~\ref{fig:worker_ablation} demonstrates the effectiveness of the augmented shuffling for Combee. We compared Combee against the plain parallel scan variant across different group sizes. 
The batch size is set to be 50.
Without augmented shuffling, quality fluctuates and is significantly worse than Combee, confirming the necessity of our design. 
Moreover, when mini-batch size is around $\sqrt{bs}$, the quality is usually higher, which validates our design in~\S\ref{sec:scan}.
The empirically optimal mini-batch size is not always exactly $\sqrt{bs}$ (e.g., GEPA peaks at mini-batch size 10 rather than 7), but $\sqrt{bs}$ consistently lies close to the optimum, supporting its use as a principled default that avoids searching the mini-batch dimension.
The curve shapes also differ across frameworks, a property of the base method's update mechanism rather than of Combee's aggregation: ACE edits a single curated playbook at each step, producing a more erratic curve, whereas GEPA's selection over a population of candidate prompts averages out individual noisy updates, yielding a smoother trend.

\begin{table}[t!]
\centering
\small
\setlength{\tabcolsep}{6pt}
\begin{tabular}{c cc cc}
\toprule
& \multicolumn{2}{c}{\textbf{Formula}} & \multicolumn{2}{c}{\textbf{FiNER}} \\
\cmidrule(lr){2-3}\cmidrule(lr){4-5}
\textbf{Duplication Level $p$} & \textbf{GEPA} & \textbf{ACE} & \textbf{GEPA} & \textbf{ACE} \\
\midrule
1 & 0.803 (407s) & 0.825 (3856s) & 0.710 (487s) & 0.765 (3295s) \\
2 (default) & \textbf{0.857} (470s) & \textbf{0.855} (3895s) & 0.742 (503s) & \textbf{0.770} (3501s) \\
3 & 0.848 (445s) & 0.830 (4073s) & 0.740 (510s) & 0.756 (4514s) \\
4 & 0.834 (483s) & 0.820 (4057s) & \textbf{0.747} (517s) & 0.749 (4326s) \\
\bottomrule
\end{tabular}
\caption{\textbf{Duplication-level ablation for augmented shuffling.} Each cell reports accuracy (completion time). The batch size is fixed to the one selected by Combee in Figures~\ref{fig:gepa_fig} and~\ref{fig:ace_fig}.}
\label{tab:dup_ablation}
\end{table}

\mypara{Duplication Level} Table~\ref{tab:dup_ablation} ablates the duplication level $p$ of augmented shuffling across both datasets and both frameworks.
Across all four settings, accuracy improves substantially from $p=1$ to $p=2$ and then plateaus or degrades, with only marginal completion time increases.
We attribute this to two competing effects: a second copy gives each reflection another chance to be grouped with different neighbors, recovering co-dependent insights that a single unlucky partition would lose, while additional copies beyond that mostly fill mini-batches with redundant content that crowds out distinct reflections.
Thus $p=2$ captures the bulk of the benefit, supporting our default choice.

\mypara{Isolating Aggregation Quality from Playbook Size} Combee's larger playbook is by design: our diagnosis of naive batching is that context overload compresses away high-value insights, so retaining more specific content is the intended mechanism.
Still, to directly isolate aggregation quality from playbook size, we use an LLM judge to score the \emph{coverage} and \emph{specificity} of final FiNER playbooks (Appendix~\ref{app:playbook_judge}): Combee nearly matches sequential ACE on both dimensions while vastly outperforming naive large-batch ACE, confirming that the gains come from better aggregation rather than merely retaining more tokens.

%% file: sections/related.tex
\tightsection{Related Work}

\paragraph{Memory Mechanism for LLMs and Agents} 
Prompt learning has been extensively used to help language models and language agents improve over time by maintaining and updating an external non-parametric memory.
Dynamic Cheatsheet~\citep{dc,xu2025mem,reflexion} demonstrates that compact, evolving textual memory can help language agents adapt at inference time by accumulating reusable guidance from past experience.
ReasoningBank~\citep{reasoningbank} similarly investigates how agents can store and reuse distilled reasoning traces or experience to support future problem solving. 
Agentic Plan Caching~\citep{apc} extends this direction by caching reusable plans from prior executions to reduce repeated reasoning and improve efficiency. 
These works, along with ACE~\citep{ace}, GEPA~\citep{gepa}, ExpeL~\citep{zhao2024expel}, Voyager~\citep{voyager}, Agent-Pro~\citep{zhang2024agent}, and TextGrad~\citep{yuksekgonul2024textgrad}, primarily focus on \emph{what} information to store, retrieve, or reuse across tasks. 
In contrast, our work focuses on \emph{how to scale the context-learning process itself}: rather than proposing a new memory abstraction, we study how multiple workers can learn in parallel and how context updates can be aggregated effectively under high concurrency.
The parallel abstraction we proposed is expected to work along with existing memory frameworks.

\paragraph{Parallel Agents} 
Recent work has explored parallel agent systems, where multiple agents or workers collaborate to solve tasks concurrently~\citep{hong2023metagpt,qian2024chatdev}. 
Learning to Share (LTS)~\citep{fioresi2026learning} studies how agents can share useful intermediate information while avoiding redundant computation. 
In practice, modern agentic coding systems such as Claude Code~\citep{claudecode}, OpenHands~\citep{wang2024openhands}, and SWE-agent~\citep{yang2024swe} also increasingly rely on parallel task decomposition and concurrent execution to improve throughput on complex workloads~\citep{li2024more,qian2024scaling}. However, these systems are primarily concerned with parallelizing \emph{task solving}~\citep{zhang2024chain,zhou2024llm,wang2024mixture}. Our focus is orthogonal: we study how to parallelize \emph{learning from experience}, i.e., how multiple workers can independently produce local context updates and how those updates can be merged into a coherent global context. Thus, while prior work on parallel agents improves execution efficiency, our work addresses the systems challenges of scalable context adaptation.

% \paragraph{Distributed Training}
% This work is partially inspired by classic ideas from distributed model training in deep learning. 
% Hogwild!~\citep{hogwild} showed that, under suitable conditions, asynchronous updates from multiple workers can still yield effective learning despite stale or conflicting writes. 
% Parameter-server architectures~\citep{li2014communication} similarly separate local computation from global state aggregation, enabling scalable training across many workers~\citep{mcmahan2017communication,stich2018local,lin2018don}. 
% While these methods operate over model parameters rather than natural-language context, they provide a useful conceptual analogy for our setting: local workers generate candidate updates based on their own experience, and a central mechanism aggregates them into a shared global state. 
% Unlike distributed optimization, however, our setting involves structured and semantically entangled context updates rather than numeric gradients, which introduces new challenges such as redundancy, inconsistency, and context overload~\citep{yuksekgonul2024textgrad}.

%% file: sections/appendix.tex
\newpage

\section{Use of Large Language Models (LLMs)}

This work focuses on developing algorithms and system frameworks for effective context adaptation in large language models (LLMs). 
Accordingly, our experiments employ LLMs for the empirical evaluation of the proposed methods. 
For paper preparation, we used LLMs only to polish writing (e.g., correcting grammatical errors), and not to generate new text from scratch.
We also use Claude Code and Cursor in the empirical experiment development process.

\section{Limitations and Future Work}
\label{app:limitations}

While Combee demonstrates consistent improvements across our evaluation settings, several aspects remain open for future work. 
First, our experiments focus on two base prompt learning frameworks (ACE and GEPA). 
Although both follow the \emph{generate-reflect-update} paradigm and Combee's design is intended to be framework-agnostic, we plan to further validate integration with methods that maintain structurally different context artifacts (e.g., program libraries or retrieval-augmented skill stores). 
Second, the dynamic batch size controller relies on a power-law delay model with a fixed marginal-reduction threshold ($\tau$), which performed well in our settings but may require adjustment for workloads with substantially different latency profiles; a systematic sensitivity analysis over $\tau$ and the candidate batch-size grid is a useful addition we leave for future work. 
The controller also profiles each candidate batch size with a single trial iteration to preserve the training budget on our relatively small training sets; production deployments with larger data volumes could afford multiple profiling iterations for more stable estimates.
Finally, the current design assumes synchronous parallel execution within each iteration; exploring asynchronous or partially-synchronous variants, analogous to asynchronous SGD in distributed training, could further improve throughput in heterogeneous deployment environments and is an interesting direction we leave for future investigation.

\mypara{Failure Modes of Hierarchical Aggregation} 
Beyond the failure modes of naive scaling characterized in \S\ref{sec:failure}, hierarchical aggregation itself has conditions under which it may degrade. 
When useful reflections are \emph{rare}, they must survive multiple levels of aggregation, and although augmented shuffling gives each reflection multiple chances to be incorporated, a sufficiently rare signal can still be diluted at the reduce level. 
When reflections are \emph{contradictory}, the aggregator must reconcile them within a single mini-batch; if contradicting reflections land in different mini-batches, the conflict is deferred to the final aggregation call, which sees only the already-consolidated updates. 
Similarly, \emph{order-dependent} or strongly \emph{co-dependent} insights benefit from appearing in the same grouping, which shuffling encourages but cannot guarantee. 
Characterizing these regimes more precisely, and designing aggregation operators that are robust to them, is an important direction for future research.

\mypara{Scaling to Larger Trace Volumes} 
Our evaluation is bounded by the scale of available open-source agentic benchmarks (90 tasks on AppWorld, 89 on Terminal-Bench 2.0), which limits the number of optimization steps in the agentic settings; the finance benchmarks provide larger training volumes but are not agentic. 
As a result, the largest-scale parallel regime that motivates Combee, e.g., industry deployments producing thousands of agent traces per day, remains untested and is exactly the setting future work can push on as richer benchmarks emerge. 
Two aspects deserve attention at that scale. 
First, context growth: playbook size management is handled by the downstream prompt learning framework rather than by Combee; ACE, for instance, includes retrieval, semantic deduplication, and metadata-based pruning designed to prevent unbounded growth, and these mechanisms were simply not triggered at our training scales. 
Whether such mechanisms remain sufficient in long-horizon deployments, and whether accumulated information becomes redundant or contradictory, is an open question. 
Second, deeper hierarchies: the parallel scan in Combee is not fundamentally limited to two levels. 
We adopt a two-level design because, at the dataset scales available, a third aggregation level would add curator calls without meaningfully reducing per-call reflection load; as training sets grow into the thousands, multi-level aggregation becomes a natural extension. 
More broadly, we view Combee as establishing parallel prompt learning as a direction, within which follow-up questions, such as better aggregation strategies, alternative shuffling or partitioning schemes, and integration with more advanced prompt learning methods, can now be studied.

\section{Extended Problem Formulation}
\label{app:formulation}

We formalize the problem of \textit{Prompt Learning at Scale} here. 
For the previous single-threaded prompt learning pipeline, agents first execute the task, 
and then reflect upon their execution to update their context. 
Mathematically, let $C_t$ denote the agent context at iteration $t$, $E$ the environment, $\tau_t$ the interaction trajectory, and $F_t$ the feedback extracted from the trajectory. The process can be written as
\[
\tau_t \sim \mathrm{Exec}(A, E \mid C_t), \qquad
F_t = \mathrm{Reflect}(\tau_t, E \mid C_t), \qquad
C_{t+1} = \mathrm{Update}(C_t, F_t).
\]

For the new paradigm of prompt learning at scale, we adopt a Map--Reduce style approach. For each iteration, we spin up parallel agents $(A_1, A_2, \dots, A_{bs})$ to interact with the environment and collect feedback $(F_t^{(1)}, F_t^{(2)}, \dots, F_t^{(bs)})$. Each agent produces its own trajectory $\tau_t^{(i)}$ and feedback signal. We then aggregate the feedback through an aggregation function to update the global agent context. Mathematically, the pipeline can be represented as
\[
\tau_t^{(i)} \sim \mathrm{Exec}(A_i, E \mid C_t), \qquad
F_t^{(i)} = \mathrm{Reflect}(\tau_t^{(i)}, E \mid C_t), \quad i=1,\dots,bs,
\]
\[
\bar{F}_t = \mathrm{Agg}\!\left(F_t^{(1)}, F_t^{(2)}, \dots, F_t^{(bs)}\right), \qquad
C_{t+1} = \mathrm{Update}(C_t, \bar{F}_t).
\]

\section{Analogy to Distributed Training}
We motivate the research vision by drawing an analogy between parallel prompt learning and distributed training of machine learning models. 
In distributed training, learning is parallelized across multiple workers, each processing a shard of data and computing local gradients. 
These gradients are periodically aggregated, either synchronously or asynchronously, by a parameter server or through collective communication, yielding a globally improved model without requiring any single worker to observe the full dataset~\citep{hogwild, dean2012large, paramserver, tensorflow}.

\emph{Prompt Learning at Scale} follows a similar pattern, but replaces parameter updates with \emph{contextual adaptation}. 
Instead of updating shared weights, multiple agents or workers independently interact with tasks, environments, or documents, and learn from their local contexts during inference. 
Each worker acquires task-relevant knowledge, such as rules, heuristics, plans, or summaries, which can be represented as prompts, memories, files, or structured artifacts like playbooks~\citep{ace}. 
These context-level updates can then be optionally consolidated, accumulated, or shared across workers, enabling downstream agents to benefit from experience they did not directly observe.

Under this analogy, contexts play a role similar to gradients: they are locally generated learning signals that encode how an agent should behave on future inputs. 
Accumulating contexts across workers resembles gradient aggregation, while curating, compressing, or distilling these artifacts parallels techniques such as gradient averaging, compression, or delayed synchronization in distributed systems. 
Crucially, this process scales learning capacity without modifying model parameters, allowing systems to improve continuously under strict inference-time and deployment constraints.

This highlights context as a first-class medium for scalable learning, suggesting that many principles from distributed training, such as parallelism, aggregation strategies, communication efficiency, and consistency trade-offs, can inspire the design of large-scale prompt learning systems.

% \paragraph{Related Work on Distributed Training}
% This work is partially inspired by classic ideas from distributed model training in deep learning. 
% Hogwild!~\citep{hogwild} showed that, under suitable conditions, asynchronous updates from multiple workers can still yield effective learning despite stale or conflicting writes. 
% Parameter-server architectures~\citep{li2014communication} similarly separate local computation from global state aggregation, enabling scalable training across many workers~\citep{mcmahan2017communication,stich2018local,lin2018don}. 
% While these methods operate over model parameters rather than natural-language context, they provide a useful conceptual analogy for our setting: local workers generate candidate updates based on their own experience, and a central mechanism aggregates them into a shared global state. 
% Unlike distributed optimization, however, our setting involves structured and semantically entangled context updates rather than numeric gradients, which introduces new challenges such as redundancy, inconsistency, and context overload~\citep{yuksekgonul2024textgrad}.

\section{Full Statistics with Standard Deviations}
\label{app:full_stats}

The full evaluation suite is computationally and financially expensive to repeat at scale. 
To provide a representative uncertainty estimate for the finance-benchmark results (Figures~\ref{fig:gepa_fig} and~\ref{fig:ace_fig}), we re-ran Combee+GEPA and the GEPA batch-size configurations with comparable delay three times each, and report the mean and standard deviation of both score and epoch time in Tables~\ref{tab:std_formula} and~\ref{tab:std_finer}. 
On both datasets, Combee matches the score of the slower GEPA configurations (batch size 5 and 10) within one standard deviation while running 3.4--7.4$\times$ faster, and substantially outperforms the large-batch GEPA configurations (batch size 50 and 80) at comparable epoch time. 
The variance across runs is relatively small, indicating that these improvements are robust rather than within noise.
In addition, Tables~\ref{tab:exact_gepa_fig}--\ref{tab:exact_gptoss} report the exact values plotted in Figures~\ref{fig:gepa_fig},~\ref{fig:ace_fig},~\ref{fig:bs_ablation},~\ref{fig:worker_ablation}, and~\ref{fig:robustness_gpt}.

\begin{table}[htb!]
\centering
\small
\begin{tabular}{l cc}
\toprule
\textbf{Method} & \textbf{Score (mean $\pm$ std)} & \textbf{Epoch time (mean $\pm$ std, s)} \\
\midrule
Combee + GEPA & 0.857 $\pm$ 0.005 & 459.3 $\pm$ 33.3 \\
GEPA (bs 80) & 0.771 $\pm$ 0.005 & 408.0 $\pm$ 14.0 \\
GEPA (bs 50) & 0.807 $\pm$ 0.013 & 461.7 $\pm$ 43.2 \\
GEPA (bs 25) & 0.855 $\pm$ 0.004 & 862.0 $\pm$ 39.2 \\
GEPA (bs 10) & 0.857 $\pm$ 0.007 & 2071.3 $\pm$ 284.9 \\
GEPA (bs 5) & 0.857 $\pm$ 0.005 & 3380.3 $\pm$ 404.6 \\
\bottomrule
\end{tabular}
\caption{\textbf{Formula dataset:} mean $\pm$ standard deviation over three runs.}
\label{tab:std_formula}
\end{table}

\begin{table}[htb!]
\centering
\small
\begin{tabular}{l cc}
\toprule
\textbf{Method} & \textbf{Score (mean $\pm$ std)} & \textbf{Epoch time (mean $\pm$ std, s)} \\
\midrule
Combee + GEPA & 0.748 $\pm$ 0.006 & 503.3 $\pm$ 14.5 \\
GEPA (bs 80) & 0.684 $\pm$ 0.011 & 429.5 $\pm$ 24.7 \\
GEPA (bs 50) & 0.714 $\pm$ 0.008 & 472.7 $\pm$ 32.3 \\
GEPA (bs 25) & 0.757 $\pm$ 0.007 & 733.0 $\pm$ 32.0 \\
GEPA (bs 10) & 0.755 $\pm$ 0.005 & 1719.0 $\pm$ 327.6 \\
GEPA (bs 5) & 0.744 $\pm$ 0.014 & 3325.7 $\pm$ 518.5 \\
\bottomrule
\end{tabular}
\caption{\textbf{FiNER dataset:} mean $\pm$ standard deviation over three runs.}
\label{tab:std_finer}
\end{table}

% TODO(authors): the Summarization and TopK values in the two tables below were recovered
% from the plotted figures via pixel calibration (the plotting scripts hardcode only the
% GEPA/ACE curves and Combee stars). Please replace them with the exact logged values.
\begin{table}[htb!]
\centering
\small
\setlength{\tabcolsep}{5pt}
\begin{tabular}{l cc cc}
\toprule
& \multicolumn{2}{c}{\textbf{Formula}} & \multicolumn{2}{c}{\textbf{FiNER}} \\
\cmidrule(lr){2-3}\cmidrule(lr){4-5}
\textbf{Method} & \textbf{Accuracy} & \textbf{Time (s)} & \textbf{Accuracy} & \textbf{Time (s)} \\
\midrule
Base LLM & 0.640 & -- & 0.666 & -- \\
Summarization + GEPA (bs 50) & 0.660 & -- & 0.675 & -- \\
TopK + GEPA (bs 50) & 0.720 & -- & 0.685 & -- \\
Hierarchical Summarization + GEPA (bs 50) & 0.650 & -- & 0.663 & -- \\
GEPA (bs 5) & 0.855 & 2981 & 0.731 & 2892 \\
GEPA (bs 10) & 0.860 & 1894 & 0.751 & 1567 \\
GEPA (bs 25) & 0.855 & 844 & 0.751 & 731 \\
GEPA (bs 50) & 0.810 & 510 & 0.723 & 507 \\
GEPA (bs 100) & 0.723 & 360 & 0.680 & 450 \\
Combee + GEPA & 0.845 (bs 84) & 338 & 0.747 (bs 94) & 435 \\
\bottomrule
\end{tabular}
\caption{\textbf{Exact values plotted in Figure~\ref{fig:gepa_fig}} (GEPA on the finance benchmarks). Each entry is the single representative run shown in the figure; Tables~\ref{tab:std_formula} and~\ref{tab:std_finer} report means and standard deviations over three runs for the corresponding configurations.}
\label{tab:exact_gepa_fig}
\end{table}

\begin{table}[htb!]
\centering
\small
\setlength{\tabcolsep}{5pt}
\begin{tabular}{l cc cc}
\toprule
& \multicolumn{2}{c}{\textbf{Formula}} & \multicolumn{2}{c}{\textbf{FiNER}} \\
\cmidrule(lr){2-3}\cmidrule(lr){4-5}
\textbf{Method} & \textbf{Accuracy} & \textbf{Time (s)} & \textbf{Accuracy} & \textbf{Time (s)} \\
\midrule
Base LLM & 0.640 & -- & 0.666 & -- \\
Summarization + ACE (bs 50) & 0.755 & -- & 0.744 & -- \\
TopK + ACE (bs 50) & 0.690 & -- & 0.725 & -- \\
Hierarchical Summarization + ACE (bs 60) & 0.725 & -- & 0.734 & -- \\
ACE (bs 5) & 0.820 & 9464 & 0.737 & 5162 \\
ACE (bs 10) & 0.810 & 7546 & 0.718 & 4140 \\
ACE (bs 20) & 0.755 & 6285 & 0.744 & 2479 \\
ACE (bs 50) & 0.745 & 4297 & 0.733 & 1072 \\
ACE (bs 60) & 0.715 & 2576 & 0.692 & 645 \\
Combee + ACE & 0.825 (bs 80) & 3510 & 0.753 (bs 64) & 1758 \\
\bottomrule
\end{tabular}
\caption{\textbf{Exact values plotted in Figure~\ref{fig:ace_fig}} (ACE on the finance benchmarks).}
\label{tab:exact_ace_fig}
\end{table}

\begin{table}[htb!]
\centering
\small
\setlength{\tabcolsep}{5pt}
\begin{tabular}{l l cc}
\toprule
\textbf{Framework} & \textbf{Configuration} & \textbf{Accuracy} & \textbf{Time (s)} \\
\midrule
\multirow{10}{*}{ACE}
& Normal parallel (bs 5) & 0.820 & 9464 \\
& Normal parallel (bs 10) & 0.810 & 7546 \\
& Normal parallel (bs 20) & 0.755 & 6285 \\
& Normal parallel (bs 50) & 0.745 & 4297 \\
& Normal parallel (bs 60) & 0.715 & 2576 \\
& Fixed bs + shuffle/scan (bs 20) & 0.850 & 5450 \\
& Fixed bs + shuffle/scan (bs 40) & 0.830 & 4145 \\
& Fixed bs + shuffle/scan (bs 60) & 0.820 & 4448 \\
& Fixed bs + shuffle/scan (bs 80) & 0.825 & 3510 \\
& Combee (dynamic, selected bs 80) & 0.825 & 3510 \\
\midrule
\multirow{10}{*}{GEPA}
& Normal parallel (bs 5) & 0.855 & 2981 \\
& Normal parallel (bs 10) & 0.860 & 1894 \\
& Normal parallel (bs 25) & 0.855 & 844 \\
& Normal parallel (bs 50) & 0.810 & 510 \\
& Normal parallel (bs 100) & 0.723 & 360 \\
& Fixed bs + shuffle/scan (bs 20) & 0.859 & 961 \\
& Fixed bs + shuffle/scan (bs 25) & 0.856 & 844 \\
& Fixed bs + shuffle/scan (bs 50) & 0.852 & 510 \\
& Fixed bs + shuffle/scan (bs 60) & 0.849 & 433 \\
& Combee (dynamic, selected bs 84) & 0.845 & 338 \\
\bottomrule
\end{tabular}
\caption{\textbf{Exact values plotted in Figure~\ref{fig:bs_ablation}} (dynamic batch size controller ablation on Formula). ``Fixed bs + shuffle/scan'' runs use augmented shuffling ($p=2$) and parallel scan with mini-batch size $\sqrt{bs}$, but a fixed batch size instead of the dynamic controller.}
\label{tab:exact_bs_ablation}
\end{table}

\begin{table}[htb!]
\centering
\small
\setlength{\tabcolsep}{5pt}
\begin{tabular}{l l ccccc}
\toprule
& & \multicolumn{5}{c}{\textbf{Mini-batch size}} \\
\cmidrule(lr){3-7}
\textbf{Framework} & \textbf{Variant} & \textbf{6} & \textbf{8} & \textbf{10} & \textbf{15} & \textbf{20} \\
\midrule
\multirow{2}{*}{ACE}
& w/o augmented shuffling & 0.835 & 0.810 & 0.755 & 0.800 & 0.785 \\
& w/ augmented shuffling & 0.845 & 0.850 & 0.815 & 0.840 & 0.810 \\
\midrule
\multirow{2}{*}{GEPA}
& w/o augmented shuffling & 0.839 & 0.827 & 0.845 & 0.825 & 0.837 \\
& w/ augmented shuffling & 0.845 & 0.849 & 0.853 & 0.847 & 0.846 \\
\bottomrule
\end{tabular}
\caption{\textbf{Exact values plotted in Figure~\ref{fig:worker_ablation}} (augmented shuffling ablation on Formula, batch size 60).}
\label{tab:exact_worker_ablation}
\end{table}

\begin{table}[htb!]
\centering
\small
\setlength{\tabcolsep}{5pt}
\begin{tabular}{l l cc}
\toprule
\textbf{Framework} & \textbf{Configuration} & \textbf{Accuracy} & \textbf{Time (s)} \\
\midrule
\multirow{7}{*}{ACE}
& Seed baseline & 0.700 & -- \\
& Normal parallel (bs 5) & 0.860 & 27906 \\
& Normal parallel (bs 10) & 0.870 & 8704 \\
& Normal parallel (bs 20) & 0.805 & 5955 \\
& Normal parallel (bs 50) & 0.730 & 3141 \\
& Normal parallel (bs 100) & 0.740 & 1955 \\
& Combee (dynamic, selected bs 75) & 0.840 & 4535 \\
\midrule
\multirow{7}{*}{GEPA}
& Seed baseline & 0.700 & -- \\
& Normal parallel (bs 5) & 0.855 & 6512 \\
& Normal parallel (bs 10) & 0.880 & 2295 \\
& Normal parallel (bs 20) & 0.870 & 942 \\
& Normal parallel (bs 50) & 0.850 & 682 \\
& Normal parallel (bs 100) & 0.825 & 532 \\
& Combee (dynamic, selected bs 88) & 0.874 & 545 \\
\bottomrule
\end{tabular}
\caption{\textbf{Exact values plotted in Figure~\ref{fig:robustness_gpt}} (GPT-OSS 120B on Formula).}
\label{tab:exact_gptoss}
\end{table}

\section{Robustness Across Model Families (GPT-OSS 120B)}
\label{app:gptoss}

\begin{figure}[htb!]
    \centering
    {\footnotesize
    \tikz[baseline=-0.6ex]{\draw[acegray,dashed,thick] (0,0)--(1.6em,0);}\ Baseline\quad
    \tikz[baseline=-0.6ex]{
        \draw[aceblue,thick] (0,0)--(1.6em,0);
        \draw[aceblue,fill=white,thick] (0.8em,0) circle (0.18em);
    }\ Normal parallel\quad
    \textcolor[HTML]{D55E00}{$\star$}\ Combee
    \par\vspace{4pt}}
    \includegraphics[width=0.6\columnwidth]{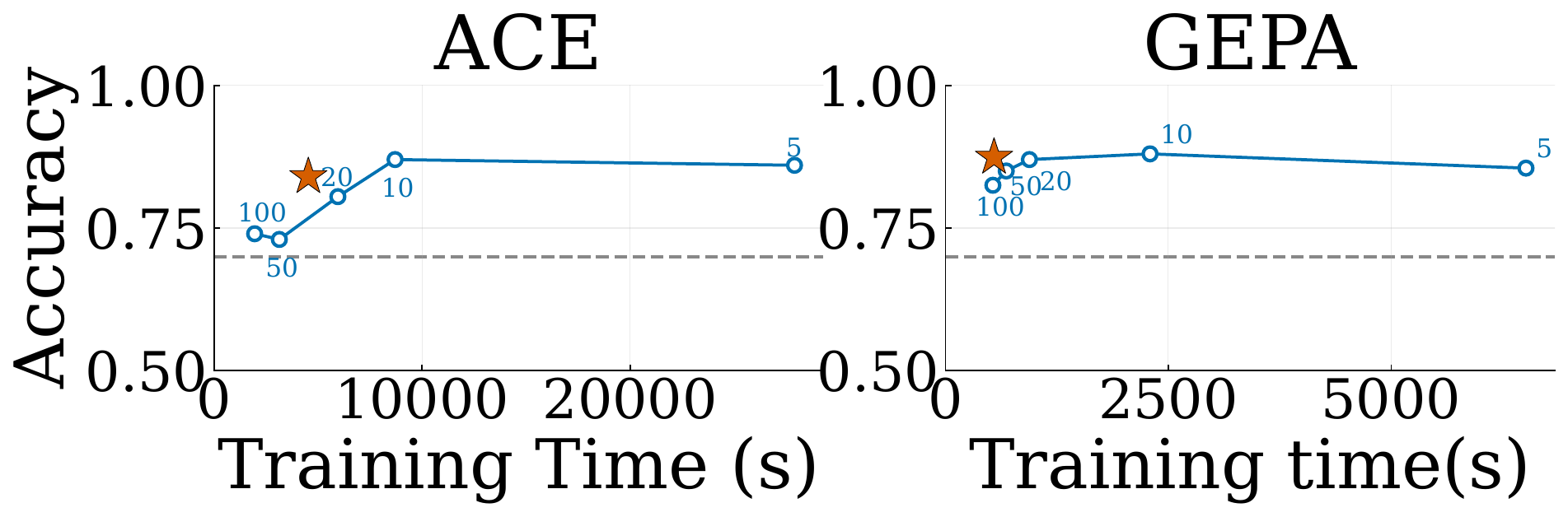}
    \caption{Combee achieves similar improvement with GPT-OSS 120B.}
    \label{fig:robustness_gpt}
\end{figure}

Figure~\ref{fig:robustness_gpt} evaluates Combee on top of GPT-OSS 120B on the same Formula dataset.
The batch size controller and parallel scan aggregator transfer seamlessly across model families, 
and Combee with GPT-OSS follows the same pattern: superior quality over fixed-batch baselines with much reduced training time.

\section{LLM-as-Judge Evaluation of Playbook Quality}
\label{app:playbook_judge}

To provide a more direct measure of retained useful information than playbook size alone, we run an LLM judge (Gemini-3.5-Flash) on the final playbooks generated by ACE and Combee on the FiNER dataset. 
The judge scores each playbook's \emph{coverage} and \emph{specificity}, and we repeat the evaluation five times to reduce randomness in LLM-as-judge scoring. 
As shown in Table~\ref{tab:playbook_judge}, Combee nearly matches sequential ACE on specificity, comes close on coverage, and vastly outperforms naive large-batch ACE (batch size 60 and 125), whose specificity collapses at batch size 125.

\begin{table}[htb!]
\centering
\small
\setlength{\tabcolsep}{8pt}
\begin{tabular}{l cc}
\toprule
\textbf{Method} & \textbf{Coverage} & \textbf{Specificity} \\
\midrule
ACE (batch 1, sequential) & 90.4 $\pm$ 3.6 & 96.0 $\pm$ 1.2 \\
ACE (batch 60) & 59.6 $\pm$ 13.0 & 90.6 $\pm$ 3.1 \\
ACE (batch 125) & 71.4 $\pm$ 2.2 & 73.8 $\pm$ 1.8 \\
Combee + ACE & 83.6 $\pm$ 5.0 & 93.8 $\pm$ 1.6 \\
\bottomrule
\end{tabular}
\caption{\textbf{LLM-as-judge evaluation of playbook quality on FiNER} (mean $\pm$ std over 5 judging runs with Gemini-3.5-Flash).}
\label{tab:playbook_judge}
\end{table}

\section{Cost and Latency Breakdown}
\label{app:cost_breakdown}

Table~\ref{tab:cost_breakdown} provides a detailed delay and cost breakdown for Combee on top of ACE on Terminal-Bench 2.0, based on a re-run of the training. 
The table reports the number of LLM calls, wall-clock time, and dollar cost separately for the reflection and curation stages, along with input and output token counts. 
Because this task reuses agent traces from an offline pool, the generation stage is not invoked during prompt learning and is therefore omitted from the breakdown; Combee shares the same parallel generation pattern as the baselines, so generation cost would be equivalent across methods and depends only on the batch size.
Combee achieves over 17$\times$ lower wall-clock time than sequential ACE at essentially identical reflection cost, while adding only negligible overhead over naive batching (12 extra LLM calls and roughly \$0.04 in curation cost due to the extended input size).

\begin{table}[htb!]
\centering
\setlength{\tabcolsep}{4pt}
\resizebox{\columnwidth}{!}{%
\begin{tabular}{l c cc cc cc cc}
\toprule
& & \multicolumn{2}{c}{\textbf{Time (s)}} & \multicolumn{2}{c}{\textbf{Cost (\$)}} & \multicolumn{2}{c}{\textbf{Reflect Tokens}} & \multicolumn{2}{c}{\textbf{Curate Tokens}} \\
\cmidrule(lr){3-4}\cmidrule(lr){5-6}\cmidrule(lr){7-8}\cmidrule(lr){9-10}
\textbf{Method (Batch Size)} & \textbf{LLM Calls} & \textbf{Reflect} & \textbf{Curate} & \textbf{Reflect} & \textbf{Curate} & \textbf{Input} & \textbf{Output} & \textbf{Input} & \textbf{Output} \\
\midrule
Combee (bs 30) & 84 & 65.68 & 96.12 & 0.1086 & 0.0659 & 266,314 & 33,365 & 158,954 & 20,906 \\
ACE (bs 1) & 120 & 1465.52 & 1580.42 & 0.1087 & 0.1405 & 266,332 & 33,470 & 366,772 & 37,692 \\
ACE (bs 30) & 72 & 64.66 & 87.65 & 0.1087 & 0.0279 & 266,338 & 33,411 & 61,542 & 10,228 \\
\bottomrule
\end{tabular}%
}
\caption{\textbf{Per-stage cost and latency breakdown on Terminal-Bench 2.0} (ACE as the base prompt learning method).}
\label{tab:cost_breakdown}
\end{table}

\section{Aggregator Prompt}
\label{app:aggregator_prompt}

We report the full aggregator (curator) prompt used in our ACE experiments below.
Placeholders in curly braces (e.g., \texttt{\{recent\_reflection\}}) are filled in at runtime.
A second variant of the prompt is used for tasks without ground-truth answers; it is identical except that the second context bullet reads ``The reflection is generated using environment feedback that will NOT be available when the playbook is being used.''
Notably, the prompt already instructs the LLM to focus on specific, high-value insights rather than generic ones (``Focus on quality over quantity,'' ``Be concise and specific---each addition should be actionable'').
Despite these explicit instructions, we still observe the lossy compression behavior described in \S\ref{sec:failure} when the batch size is large: the aggregator tends to keep generic playbook entries. 
This suggests that prompt-level guidance alone is insufficient to overcome context overload, motivating Combee's structural approach of reducing the per-call reflection load via parallel scan aggregation.

\begin{playbookbox}{Aggregator (Curator) Prompt}
\small
You are a master curator of knowledge. Your job is to identify what new insights should be added to an existing playbook based on a reflection from a previous attempt.

\medskip
\textbf{Context:}
\begin{itemize}[leftmargin=1.2em, itemsep=1pt, parsep=0pt, topsep=2pt]
\item The playbook you created will be used to help answering similar questions.
\item The reflection is generated using ground truth answers that will NOT be available when the playbook is being used. So you need to come up with content that can aid the playbook user to create predictions that likely align with ground truth.
\end{itemize}

\medskip
\textbf{CRITICAL: You MUST respond with valid JSON only. Do not use markdown formatting or code blocks.}

\medskip
\textbf{Instructions:}
\begin{itemize}[leftmargin=1.2em, itemsep=1pt, parsep=0pt, topsep=2pt]
\item Review the existing playbook and the reflection from the previous attempt
\item Identify ONLY the NEW insights, strategies, or mistakes that are MISSING from the current playbook
\item Avoid redundancy---if similar advice already exists, only add new content that is a perfect complement to the existing playbook
\item Do NOT regenerate the entire playbook---only provide the additions needed
\item Focus on quality over quantity---a focused, well-organized playbook is better than an exhaustive one
\item Format your response as a PURE JSON object with specific sections
\item For any operation if no new content to add, return an empty list for the operations field
\item Be concise and specific---each addition should be actionable
\end{itemize}

\medskip
\textbf{Training Context:}
\begin{itemize}[leftmargin=1.2em, itemsep=1pt, parsep=0pt, topsep=2pt]
\item Total token budget: \texttt{\{token\_budget\}} tokens
\item Training progress: Sample \texttt{\{current\_step\}} out of \texttt{\{total\_samples\}}
\end{itemize}

\medskip
\textbf{Current Playbook Stats:} \texttt{\{playbook\_stats\}}

\textbf{Recent Reflection:} \texttt{\{recent\_reflection\}}

\textbf{Current Playbook:} \texttt{\{current\_playbook\}}

\textbf{Question Context:} \texttt{\{question\_context\}}

\medskip
\textbf{Your Task:}
Output ONLY a valid JSON object with these exact fields:
\begin{itemize}[leftmargin=1.2em, itemsep=1pt, parsep=0pt, topsep=2pt]
\item reasoning: your chain of thought / reasoning / thinking process, detailed analysis and calculations
\item operations: a list of operations to be performed on the playbook
\begin{itemize}[leftmargin=1.2em, itemsep=1pt, parsep=0pt, topsep=1pt]
\item type: the type of operation to be performed
\item section: the section to add the bullet to
\item content: the new content of the bullet
\end{itemize}
\end{itemize}

\medskip
\textbf{Available Operations:}
\begin{enumerate}[leftmargin=1.4em, itemsep=1pt, parsep=0pt, topsep=2pt]
\item ADD: Create new bullet points with fresh IDs
\begin{itemize}[leftmargin=1.2em, itemsep=1pt, parsep=0pt, topsep=1pt]
\item section: the section to add the new bullet to
\item content: the new content of the bullet. Note: no need to include the bullet\_id in the content like `[ctx-00263] helpful=1 harmful=0 ::', the bullet\_id will be added by the system.
\end{itemize}
\end{enumerate}

\medskip
\textbf{RESPONSE FORMAT---Output ONLY this JSON structure (no markdown, no code blocks):}

\medskip
{\ttfamily
\{\\
\hspace*{1em}"reasoning": "[Your chain of thought / reasoning / thinking process, detailed analysis and calculations here]",\\
\hspace*{1em}"operations": [\\
\hspace*{2em}\{\\
\hspace*{3em}"type": "ADD",\\
\hspace*{3em}"section": "formulas\_and\_calculations",\\
\hspace*{3em}"content": "[New calculation method...]"\\
\hspace*{2em}\}\\
\hspace*{1em}]\\
\}
}
\end{playbookbox}

\section{Qualitative Examples of Context Overload}
\label{app:overload_examples}

% We provide concrete playbook snapshots to illustrate how context overload degrades the quality of learned artifacts.
% Figures~\ref{fig:playbook_formula_bs1}--\ref{fig:playbook_finer_bs125} show final playbooks produced by ACE on the Formula and FiNER benchmarks under sequential learning (batch size~1) and high-parallelism naive scaling (batch size~100 and 125, respectively).
% Under sequential learning, the aggregator produces detailed, actionable entries that capture nuanced, task-specific strategies.
% As the batch size increases under naive scaling, the aggregator is overwhelmed by the volume of reflections, and the resulting playbook collapses to far fewer and more generic entries---directly mirroring the quantitative trends in Figure~\ref{fig:naive_failing}.

\input{sections/playbook_exs}

%% file: sections/playbook_exs.tex
\begin{playbookbox}{ACE Final Playbook --- Formula, Batch Size = 100}
\vspace{2pt}
\begin{center}
{\small Total context entries: \textbf{21} \quad|\quad Test accuracy: \textbf{72.5\%}}
\end{center}
\vspace{2pt}
\pbsection{FORMULAS \& CALCULATIONS}
\begin{itemize}[leftmargin=1.2em, itemsep=1pt, parsep=0pt, topsep=2pt]
\item {\ttfamily\bfseries\color{bulletblue}[calc-00009]} {\scriptsize\color{countgray}h=0 r=2} For relative purchasing power parity (PPP) with changing exchange rates: $P_{1}$ = $P_{0}$ * ($S_{1}$/$S_{0}$), where $P_{1}$ is the new price level in the quote country, $P_{0}$ is the initial price level in the base country, ...
\item {\ttfamily\bfseries\color{bulletblue}[calc-00013]} {\scriptsize\color{countgray}h=0 r=0} For put-call parity with discrete compounding, use P = C - S + X / (1 + r)\^{}t to find put price, where C is call price, S is spot price, X is strike price, r is risk-free rate, and t is time. This ...
\item {\ttfamily\bfseries\color{bulletblue}[calc-00014]} {\scriptsize\color{countgray}h=2 r=0} For annuity due (cash flows at beginning of periods), use PV = C * [1 - (1 + r)\^{}(-n)] / r * (1 + r). For ordinary annuity (end of periods), use PV = C * [1 - (1 + r)\^{}(-n)] / r. Verify timing from ...
\item {\ttfamily\bfseries\color{bulletblue}[calc-00015]} {\scriptsize\color{countgray}h=0 r=0} For Interest Rate Parity (IRP), the forward rate formula is F = S * (1 + r\_base) / (1 + r\_quote), where base is the numerator currency and quote is the denominator currency in the exchange rate ...
\item {\ttfamily\bfseries\color{bulletblue}[calc-00020]} {\scriptsize\color{countgray}h=0 r=0} For annuity due calculations where the initial investment is not explicitly given (e.g., rental properties), the present value of cash inflows alone may be the expected answer. Use PV = C * [1 - ...
\end{itemize}
\pbsection{COMMON MISTAKES TO AVOID}
\begin{itemize}[leftmargin=1.2em, itemsep=1pt, parsep=0pt, topsep=2pt]
\item {\ttfamily\bfseries\color{bulletblue}[err-00001]} {\scriptsize\color{countgray}h=1 r=2} Avoid automatically converting decimal results to percentages. When the question asks for a 'plain floating point number', output the decimal equivalent (e.g., 0.05 for 5\%) rather than the ...
\item {\ttfamily\bfseries\color{bulletblue}[err-00002]} {\scriptsize\color{countgray}h=0 r=2} For annuity calculations, carefully determine the timing of cash flows. If payments occur at the beginning of periods (e.g., grants, rents, or insurance premiums), use the annuity due formula: PV ...
\item {\ttfamily\bfseries\color{bulletblue}[err-00005]} {\scriptsize\color{countgray}h=0 r=0} For financial ratios like ROA, ROE, profit margins, and other performance metrics, output the percentage value (e.g., 5.0) rather than the decimal equivalent (0.05) when the context indicates ...
\item {\ttfamily\bfseries\color{bulletblue}[err-00006]} {\scriptsize\color{countgray}h=0 r=0} When calculating option price differences using put-call parity, remember to multiply by the standard contract multiplier of 100 shares for equity options. The formula gives per-share values, but ...
\item {\ttfamily\bfseries\color{bulletblue}[err-00007]} {\scriptsize\color{countgray}h=0 r=2} Verify the compounding assumption in option pricing formulas. Put-call parity typically uses continuous compounding (e\^{}(-r*t)), but some contexts may expect discrete compounding ((1+r)\^{}(-t)). ...
\item {\ttfamily\bfseries\color{bulletblue}[err-00008]} {\scriptsize\color{countgray}h=0 r=0} For ROI calculations on investments with ongoing benefits, clarify whether to use incremental profit or (new profit - cost) in the numerator. When in doubt, use incremental profit (increase from ...
\item {\ttfamily\bfseries\color{bulletblue}[err-00010]} {\scriptsize\color{countgray}h=0 r=0} For financial ratios and rates (e.g., ROA, ROE, profit margins, inflation rate, ROI), output the percentage value as a floating point number (e.g., 5.0) rather than the decimal (0.05), as they are ...
\item {\ttfamily\bfseries\color{bulletblue}[err-00011]} {\scriptsize\color{countgray}h=0 r=2} Maintain high precision in intermediate calculations, especially for exponentiation (e.g., (1+r)\^{}n), division, and financial models like Black-Scholes. Avoid rounding until the final result to ...
\item {\ttfamily\bfseries\color{bulletblue}[err-00012]} {\scriptsize\color{countgray}h=0 r=0} In ROI calculations, clearly distinguish whether 'gain' refers to incremental income increase or total new income. When uncertain, use (incremental income - cost) / cost, but verify against ...
\item {\ttfamily\bfseries\color{bulletblue}[err-00016]} {\scriptsize\color{countgray}h=0 r=0} For financial ratios and rates, output percentages (e.g., 5.0) for ROA, ROE, profit margins, inflation rate, and ROI, as they are conventionally reported as percentages. Output decimals (e.g., ...
\item {\ttfamily\bfseries\color{bulletblue}[err-00017]} {\scriptsize\color{countgray}h=0 r=0} When the question explicitly specifies rounding precision (e.g., 'round to nearest hundredth'), always apply the rounding directive to the final result, regardless of financial conventions. For ...
\item {\ttfamily\bfseries\color{bulletblue}[err-00018]} {\scriptsize\color{countgray}h=0 r=0} For financial ratios like profit margins, ROE, ROA, and efficiency ratios, output the percentage value (e.g., 15.0) rather than the decimal (0.15) unless explicitly instructed otherwise. These ...
\item {\ttfamily\bfseries\color{bulletblue}[err-00019]} {\scriptsize\color{countgray}h=0 r=0} When asked how much a value 'changes' or the 'change' in a ratio, determine if the context requires the absolute magnitude of change (e.g., decreased by 1.5) or the directional change with sign ...
\end{itemize}
\pbsection{OTHERS}
\begin{itemize}[leftmargin=1.2em, itemsep=1pt, parsep=0pt, topsep=2pt]
\item {\ttfamily\bfseries\color{bulletblue}[misc-00003]} {\scriptsize\color{countgray}h=1 r=0} When calculating financial ratios or rates, first compute the decimal result, then check the required output format. Only convert to percentage if explicitly requested or if the context clearly ...
\item {\ttfamily\bfseries\color{bulletblue}[misc-00004]} {\scriptsize\color{countgray}h=1 r=0} The phrase 'plain floating point number' in questions typically indicates that results should be in decimal form (e.g., 0.05) rather than percentage form (5.0). Examples like '5 million should be ...
\item {\ttfamily\bfseries\color{bulletblue}[misc-00021]} {\scriptsize\color{countgray}h=0 r=0} When time values are calculated and the result seems unusually small or large, consider whether unit conversion is expected (e.g., years to days by multiplying by 365). Check ground truth patterns ...
\end{itemize}
\end{playbookbox}
\vspace{8pt}
\phantomsection\label{fig:playbook_finer_bs125}
\begin{playbookbox}{ACE Final Playbook --- FiNER, Batch Size = 125}
\vspace{2pt}
\begin{center}
{\small Total context entries: \textbf{11} \quad|\quad Test accuracy: \textbf{70.6\%}}
\end{center}
\vspace{2pt}
\pbsection{STRATEGIES \& INSIGHTS}
\begin{itemize}[leftmargin=1.2em, itemsep=1pt, parsep=0pt, topsep=2pt]
\item {\ttfamily\bfseries\color{bulletblue}[sai-00010]} {\scriptsize\color{countgray}h=0 r=0} When tagging line of credit facilities, distinguish between current borrowing capacity (for existing facilities - use LineOfCreditFacilityCurrentBorrowingCapacity) and maximum borrowing capacity ...
\item {\ttfamily\bfseries\color{bulletblue}[sai-00011]} {\scriptsize\color{countgray}h=0 r=0} For business acquisitions, use BusinessAcquisitionEquityInterestsIssuedOrIssuableNumberOfSharesIssued for equity instruments issued as consideration, in addition to distinguishing between total ...
\end{itemize}
\pbsection{COMMON MISTAKES TO AVOID}
\begin{itemize}[leftmargin=1.2em, itemsep=1pt, parsep=0pt, topsep=2pt]
\item {\ttfamily\bfseries\color{bulletblue}[err-00003]} {\scriptsize\color{countgray}h=0 r=0} Avoid confusing share-based compensation expense recognition with grant measurement. Use AllocatedShareBasedCompensationExpense for recognized expenses during a period, and reserve grant-related ...
\item {\ttfamily\bfseries\color{bulletblue}[err-00007]} {\scriptsize\color{countgray}h=0 r=0} Avoid confusing debt instrument components: use DebtInstrumentFaceAmount for principal/face value, DebtInstrumentCarryingAmount for net book value after discounts/premiums, and ...
\item {\ttfamily\bfseries\color{bulletblue}[err-00008]} {\scriptsize\color{countgray}h=0 r=0} Avoid over-specifying share issuance tags: use StockIssuedDuringPeriodSharesNewIssues for general period-level reporting of shares issued, and reserve SaleOfStockNumberOfSharesIssuedInTransaction ...
\end{itemize}
\pbsection{CONTEXT CLUES \& INDICATORS}
\begin{itemize}[leftmargin=1.2em, itemsep=1pt, parsep=0pt, topsep=2pt]
\item {\ttfamily\bfseries\color{bulletblue}[ctx-00009]} {\scriptsize\color{countgray}h=0 r=0} Key phrases indicating specific tag requirements: 'borrowings' or 'outstanding borrowings' $\rightarrow$ DebtInstrumentCarryingAmount, 'available' or 'unused capacity' $\rightarrow$ ...
\end{itemize}
\pbsection{OTHERS}
\begin{itemize}[leftmargin=1.2em, itemsep=1pt, parsep=0pt, topsep=2pt]
\item {\ttfamily\bfseries\color{bulletblue}[misc-00001]} {\scriptsize\color{countgray}h=1 r=0} When tagging business acquisition transactions, distinguish between total consideration (BusinessCombinationConsiderationTransferred1) and cash components (PaymentsToAcquireBusinessesGross). Use ...
\item {\ttfamily\bfseries\color{bulletblue}[misc-00002]} {\scriptsize\color{countgray}h=1 r=0} For debt instruments, prefer specific tags over general ones when context supports it. Use LongTermDebtFairValue for long-term debt fair values rather than the generic DebtInstrumentFairValue, and ...
\item {\ttfamily\bfseries\color{bulletblue}[misc-00004]} {\scriptsize\color{countgray}h=0 r=0} Key phrases indicating specific tag requirements: 'undrawn amounts' $\rightarrow$ LineOfCreditFacilityUnusedCapacityCommitmentFeePercentage, 'net of actual and estimated forfeitures' $\rightarrow$ ...
\item {\ttfamily\bfseries\color{bulletblue}[misc-00005]} {\scriptsize\color{countgray}h=1 r=0} When multiple tags could apply, always choose the most specific tag that matches the exact context. Verify tag names precisely against available options (e.g., OperatingLeasesRentExpenseNet vs ...
\item {\ttfamily\bfseries\color{bulletblue}[misc-00006]} {\scriptsize\color{countgray}h=1 r=0} For ownership percentages, use MinorityInterestOwnershipPercentageByParent for parent's ownership in consolidated subsidiaries, and EquityMethodInvestmentOwnershipPercentage for investments with ...
\end{itemize}
\end{playbookbox}
\vspace{8pt}
\phantomsection\label{fig:playbook_formula_bs1}
\begin{playbookbox}{ACE Final Playbook --- Formula, Batch Size = 1}
\vspace{2pt}
\begin{center}
{\small Total context entries: \textbf{264} \quad|\quad Test accuracy: \textbf{87.0\%}}
\end{center}
\vspace{2pt}
\pbsection{FORMULAS \& CALCULATIONS}
\begin{itemize}[leftmargin=1.2em, itemsep=1pt, parsep=0pt, topsep=2pt]
\item {\ttfamily\bfseries\color{bulletblue}[calc-00001]} {\scriptsize\color{countgray}h=6 r=0} Current Ratio = Current Assets / Current Liabilities. Always verify both values are in the same units before calculating. Round the result to two decimal places unless otherwise specified.
\item {\ttfamily\bfseries\color{bulletblue}[calc-00002]} {\scriptsize\color{countgray}h=1 r=1} ROI = (Net Profit / Total Investment) * 100. Convert percentage to decimal by dividing by 100. Note: 0.20 and 0.2 are numerically equivalent; trailing zeros after decimal don't change value.
\item {\ttfamily\bfseries\color{bulletblue}[calc-00003]} {\scriptsize\color{countgray}h=8 r=1} NPV = $\Sigma$ [CF\_t / (1 + r)\^{}t]. Carefully check cash flow timing: if described as 'for the next year' or 'starting now', the first cash flow may be at t=0 (immediate, no discounting). Subsequent cash ...
\item {\ttfamily\bfseries\color{bulletblue}[calc-00004]} {\scriptsize\color{countgray}h=2 r=0} Sharpe Ratio = (Portfolio Return - Risk-Free Rate) / Portfolio Standard Deviation. Convert all percentage inputs to decimal form before calculation (e.g., 7\% = 0.07). Round the final result to two ...
\item {\ttfamily\bfseries\color{bulletblue}[calc-00005]} {\scriptsize\color{countgray}h=1 r=0} CAPM: Expected Return = Risk-Free Rate + Beta * (Market Return - Risk-Free Rate). All inputs and outputs should be in decimal form (not percentages). Round the final result to the specified ...
\item {\ttfamily\bfseries\color{bulletblue}[calc-00006]} {\scriptsize\color{countgray}h=2 r=0} DSO (Days Sales Outstanding) = (Accounts Receivable / Credit Sales) * Number of Days in Period. For a quarter, assume 90 days unless specified otherwise. Ensure both Accounts Receivable and Credit ...
\item {\ttfamily\bfseries\color{bulletblue}[calc-00007]} {\scriptsize\color{countgray}h=0 r=0} Sortino Ratio = (Portfolio Return - Risk-Free Rate) / Downward Volatility. Convert all percentage inputs to decimal form before calculation (e.g., 10\% = 0.10). Round the final result to two ...
\item {\ttfamily\bfseries\color{bulletblue}[calc-00008]} {\scriptsize\color{countgray}h=0 r=0} Accounts Receivable Turnover = Net Credit Sales / Average Accounts Receivable. Ensure both values are in the same currency units. The result is typically expressed as a number (not percentage). ...
\item {\ttfamily\bfseries\color{bulletblue}[calc-00009]} {\scriptsize\color{countgray}h=7 r=4} For NPV of annuity cash flows: if cash flows occur at period end (ordinary annuity), use NPV = CF * [1 - (1 + r)\^{}(-n)] / r; if at period beginning (annuity due), use NPV = CF * [1 - (1 + r)\^{}(-n)] ...
\item {\ttfamily\bfseries\color{bulletblue}[calc-00010]} {\scriptsize\color{countgray}h=4 r=0} Modigliani-Miller Proposition I with taxes: V\_L = V\_U + (T\_c * D), where V\_L is levered firm value, V\_U is unlevered firm value, T\_c is corporate tax rate (as decimal), and D is debt value. The ...
\item {\ttfamily\bfseries\color{bulletblue}[calc-00011]} {\scriptsize\color{countgray}h=2 r=0} Gross Profit Margin = (Total Sales - Cost of Goods Sold) / Total Sales * 100. Ensure both sales and COGS values are in the same currency units and from the same period. The result should be ...
\item {\ttfamily\bfseries\color{bulletblue}[calc-00012]} {\scriptsize\color{countgray}h=0 r=0} WACC = (E/V) * Re + (D/V) * Rd * (1 - Tc). Always use decimal form for all inputs (Re, Rd, Tc) and output. When the question specifies 'a plain floating point number', it means the decimal form ...
\item {\ttfamily\bfseries\color{bulletblue}[calc-00013]} {\scriptsize\color{countgray}h=1 r=0} Inventory Turnover = Cost of Goods Sold (COGS) / Average Inventory. Ensure both values are in the same currency units. The result is a ratio (number of times), not a percentage. Round to the ...
\item {\ttfamily\bfseries\color{bulletblue}[calc-00014]} {\scriptsize\color{countgray}h=6 r=2} For NPV of project cash flows described as 'annual cash inflows' without explicit timing, assume annuity due (cash flows at beginning of period) as projects often start immediately. Use ...
\item {\ttfamily\bfseries\color{bulletblue}[calc-00015]} {\scriptsize\color{countgray}h=0 r=0} Beta = Covariance(Stock, Market) / Variance(Market). Ensure both covariance and variance are calculated from the same data period and in consistent units. Round the final result to the required ...
\item {\ttfamily\bfseries\color{bulletblue}[calc-00016]} {\scriptsize\color{countgray}h=0 r=0} For continuous compounding FV = P * e\^{}(r*t), use e\^{}(r*t) rounded to 6 decimal places by default (e.g., e\^{}(0.21) $\approx$ 1.233678) unless higher precision is specified. Financial contexts may expect this ...
\item {\ttfamily\bfseries\color{bulletblue}[calc-00017]} {\scriptsize\color{countgray}h=3 r=3} Put-Call Parity: S = C - P + X * (1 + r)\^{}\{-t\} for discrete compounding (annual) or S = C - P + X * e\^{}\{-r*t\} for continuous compounding. Default to discrete compounding ((1+r)\^{}\{-t\}) unless the ...
\item {\ttfamily\bfseries\color{bulletblue}[calc-00018]} {\scriptsize\color{countgray}h=0 r=0} For put-call parity problems, if the standard formula (C - P = S - X * (1 + r)\^{}\{-t\}) yields a result significantly different from expected values, verify the arrangement. In some contexts, the ...
\item {\ttfamily\bfseries\color{bulletblue}[calc-00020]} {\scriptsize\color{countgray}h=1 r=0} For relative Purchasing Power Parity (PPP) problems where the exchange rate is quoted as EUR/CHF (domestic/foreign), the formula for the final domestic price level (P\_1\_EUR) is: P\_1\_EUR = P\_0\_CHF ...
\item {\ttfamily\bfseries\color{bulletblue}[calc-00021]} {\scriptsize\color{countgray}h=1 r=1} For APR calculations: APR = (Total Finance Charge / Principal) * (365 / Term in Days). Output must be in decimal form (e.g., 1.22 for 122\%) unless specified otherwise. Always adhere to 'plain ...
\item {\ttfamily\bfseries\color{bulletblue}[calc-00022]} {\scriptsize\color{countgray}h=3 r=0} Gordon Growth Model (Constant Growth DDM): P0 = D1 / (r - g). D1 is the expected dividend next period, r is the required rate of return (cost of equity), and g is the constant growth rate. Convert ...
\item {\ttfamily\bfseries\color{bulletblue}[calc-00023]} {\scriptsize\color{countgray}h=3 r=0} Black-Scholes Call Option Price: C = S\_0*N(d1) - X*e\^{}(-r*t)*N(d2). Always convert percentage inputs (r, $\sigma$) to decimal form. Calculate d1 = [ln(S\_0/X) + (r + $\sigma^2$/2)*t] / ($\sigma$*$\sqrt{}$t) and d2 = d1 - $\sigma$*$\sqrt{}$t ...
\item {\ttfamily\bfseries\color{bulletblue}[calc-00024]} {\scriptsize\color{countgray}h=0 r=0} P/B Ratio = Market Price per Share / Book Value per Share. Ensure both values are in the same currency units (e.g., dollars). The result is a ratio and should be output as a plain floating point ...
\item {\ttfamily\bfseries\color{bulletblue}[calc-00025]} {\scriptsize\color{countgray}h=1 r=0} Treynor Ratio = (Portfolio Return - Risk-Free Rate) / Beta. Convert all percentage inputs to decimal form before calculation (e.g., 6\% = 0.06). Round the final result to two decimal places unless ...
\item {\ttfamily\bfseries\color{bulletblue}[calc-00026]} {\scriptsize\color{countgray}h=0 r=0} DSCR = Net Operating Income / Annual Debt Service. Use the values as provided without unit conversion or unnecessary adjustments. The result is typically a plain floating point number; round only ...
\item {\ttfamily\bfseries\color{bulletblue}[calc-00027]} {\scriptsize\color{countgray}h=4 r=0} Free Cash Flow (FCF) = Operational Cash Flow - Capital Expenditures (CAPEX). CAPEX includes expenditures that maintain or enhance long-term assets (e.g., machinery maintenance). Ensure both values ...
\item {\ttfamily\bfseries\color{bulletblue}[calc-00028]} {\scriptsize\color{countgray}h=2 r=0} Net Profit Margin = (Net Profit / Total Revenue) * 100. When the question asks for a 'plain floating point number', output the percentage value (e.g., 15.0 for 15\%) rather than the decimal ratio ...
\item {\ttfamily\bfseries\color{bulletblue}[calc-00029]} {\scriptsize\color{countgray}h=0 r=0} Interest Coverage Ratio = EBIT / Interest Expense. Ensure both EBIT and Interest Expense are in the same currency units. The result is a plain floating point number; no rounding is necessary if ...
\item {\ttfamily\bfseries\color{bulletblue}[calc-00030]} {\scriptsize\color{countgray}h=0 r=0} Simple Interest = Principal $\times$ Rate (as decimal). When compounding is not specified (e.g., 'annual interest rate' without mention of compounding periods), assume simple interest. Convert the ...
\item {\ttfamily\bfseries\color{bulletblue}[calc-00031]} {\scriptsize\color{countgray}h=2 r=0} Operating Margin = (Operating Income / Sales) * 100. Ensure both values are in the same currency units. When output is requested as a 'plain floating point number', provide the percentage value ...
\item {\ttfamily\bfseries\color{bulletblue}[calc-00032]} {\scriptsize\color{countgray}h=1 r=0} Debt-to-Equity Ratio = Total Liabilities / Shareholder Equity. Ensure both values are in the same currency units. The result is a plain floating point number; no rounding is needed for integer ...
\item {\ttfamily\bfseries\color{bulletblue}[calc-00033]} {\scriptsize\color{countgray}h=1 r=0} Unemployment Rate = (Number of Unemployed / Labor Force) * 100. When the problem states the number of unemployed 'rises by' a specific amount and the labor force is stagnant, and no initial ...
\item {\ttfamily\bfseries\color{bulletblue}[calc-00034]} {\scriptsize\color{countgray}h=0 r=0} P/E Ratio = Market Price per Share / Earnings per Share. Ensure both values are in the same currency units. The result is a plain floating point number; no unit conversion or percentage formatting ...
\item {\ttfamily\bfseries\color{bulletblue}[calc-00035]} {\scriptsize\color{countgray}h=0 r=0} Return on Assets (ROA) = Net Income / Total Assets. The result is typically reported as a percentage. When the question asks for a 'plain floating point number', output the percentage value (e.g., ...
\item {\ttfamily\bfseries\color{bulletblue}[calc-00036]} {\scriptsize\color{countgray}h=3 r=0} For Black-Scholes calculations, compute cumulative normal distribution values N(d1) and N(d2) with high precision (at least 4-5 decimal places). Use precise methods like Excel's NORM.S.DIST, ...
\item {\ttfamily\bfseries\color{bulletblue}[calc-00037]} {\scriptsize\color{countgray}h=1 r=0} Return on Equity (ROE) = Net Income / Shareholder's Equity. The result is typically expressed as a percentage in financial contexts. After calculating the decimal value, multiply by 100 to get the ...
\item {\ttfamily\bfseries\color{bulletblue}[calc-00038]} {\scriptsize\color{countgray}h=2 r=0} Market Capitalization = Shares Outstanding $\times$ Price per Share. Ensure both inputs are in consistent units (e.g., shares and dollars per share). The result is a currency value; output as a plain ...
\item {\ttfamily\bfseries\color{bulletblue}[calc-00039]} {\scriptsize\color{countgray}h=0 r=0} Gordon Growth Model for cost of equity: r = (D1 / P0) + g. Use this rearrangement when solving for required return rather than stock price. Convert percentage growth rates (g) to decimal form ...
\item {\ttfamily\bfseries\color{bulletblue}[calc-00040]} {\scriptsize\color{countgray}h=2 r=0} Working Capital = Current Assets - Current Liabilities. Ensure both values are in the same currency units. The result is a currency value; output as a plain floating point number (e.g., 70000.0) ...
\item {\ttfamily\bfseries\color{bulletblue}[calc-00041]} {\scriptsize\color{countgray}h=0 r=0} EPS = (Net Income - Preferred Dividends) / Weighted Average Shares Outstanding. Always subtract preferred dividends from net income first to get earnings available to common shareholders. Round ...
\item {\ttfamily\bfseries\color{bulletblue}[calc-00042]} {\scriptsize\color{countgray}h=1 r=0} Trade Balance = Exports - Imports. Convert all values to the same units (e.g., millions to actual numbers: \$120 million = 120,000,000) before subtracting. Output the result as a plain floating ...
\item {\ttfamily\bfseries\color{bulletblue}[calc-00043]} {\scriptsize\color{countgray}h=3 r=0} For present value calculations using PV = FV / (1 + r)\^{}t, ensure high numerical precision in both exponentiation and division. Avoid manual multiplication steps for (1+r)\^{}t as they can introduce ...
\item {\ttfamily\bfseries\color{bulletblue}[calc-00044]} {\scriptsize\color{countgray}h=1 r=0} Discrete Compound Interest: FV = PV * (1 + r)\^{}n where r is the periodic interest rate (convert percentage to decimal by dividing by 100) and n is the number of compounding periods. Ensure n ...
\item {\ttfamily\bfseries\color{bulletblue}[calc-00045]} {\scriptsize\color{countgray}h=1 r=0} Enterprise Value (EV) = Market Capitalization + Total Debt - Cash \& Cash Equivalents. Ensure all values are in consistent units (e.g., convert millions to actual numbers: 70 million = 70,000,000). ...
\item {\ttfamily\bfseries\color{bulletblue}[calc-00047]} {\scriptsize\color{countgray}h=0 r=0} For quarterly compounding interest: FV = PV * (1 + r/4)\^{}(4*t). Always convert the annual rate (r) to a quarterly rate by dividing by 4. Calculate the total number of quarters (4*t) for the ...
\item {\ttfamily\bfseries\color{bulletblue}[calc-00048]} {\scriptsize\color{countgray}h=1 r=0} Dividend Payout Ratio = Dividends per Share / Earnings per Share. Output as a plain floating point number (decimal ratio, not percentage). Both 0.2 and 0.20 are numerically equivalent and ...
\item {\ttfamily\bfseries\color{bulletblue}[calc-00049]} {\scriptsize\color{countgray}h=1 r=0} Dividend Yield = Annual Dividend per Share / Share Price. When the question asks for a 'plain floating point number', output the decimal ratio (e.g., 0.02) without multiplying by 100. This aligns ...
\item {\ttfamily\bfseries\color{bulletblue}[calc-00050]} {\scriptsize\color{countgray}h=1 r=0} Inflation Rate = ((Current CPI - Previous CPI) / Previous CPI) * 100. Ensure both CPI values are from consecutive periods (e.g., previous year vs. current year). Convert the result to a percentage ...
\item {\ttfamily\bfseries\color{bulletblue}[calc-00051]} {\scriptsize\color{countgray}h=0 r=0} For high-precision compound interest calculations, especially with small interest rates over many periods, use FV = PV * e\^{}(n * ln(1 + r)) to minimize numerical errors. This method is more stable ...
\item {\ttfamily\bfseries\color{bulletblue}[calc-00052]} {\scriptsize\color{countgray}h=1 r=0} Jensen's Alpha = Actual Portfolio Return - [Risk-Free Rate + Beta * (Market Return - Risk-Free Rate)]. Convert all percentage inputs to decimal form before calculation. First calculate the CAPM ...
\item {\ttfamily\bfseries\color{bulletblue}[calc-00053]} {\scriptsize\color{countgray}h=2 r=0} For Treynor Ratio and similar financial ratios, note that numerical outputs like 0.10 and 0.1 are equivalent; trailing zeros after the decimal do not change the value. This is particularly ...
\item {\ttfamily\bfseries\color{bulletblue}[calc-00054]} {\scriptsize\color{countgray}h=0 r=0} For semi-annual compounding interest: FV = PV * (1 + r/2)\^{}(2*t). Always convert the annual rate (r) to a semi-annual rate by dividing by 2. Calculate the total number of semi-annual periods (2*t) ...
\item {\ttfamily\bfseries\color{bulletblue}[calc-00055]} {\scriptsize\color{countgray}h=5 r=1} EAR (Effective Annual Rate) = (1 + r/n)\^{}n - 1, where r is the nominal annual rate (as decimal) and n is the number of compounding periods per year. Output the result as a plain floating point ...
\item {\ttfamily\bfseries\color{bulletblue}[calc-00056]} {\scriptsize\color{countgray}h=8 r=1} For financial ratios that are conventionally reported as percentages (e.g., ROE, ROA, Net Profit Margin, Operating Margin), output the percentage value (e.g., 20.0) when the question requests a ...
\item {\ttfamily\bfseries\color{bulletblue}[calc-00057]} {\scriptsize\color{countgray}h=12 r=0} For ratios that yield exact, integer-like results (e.g., Current Ratio = 150000 / 75000 = 2.0), output the result without applying additional rounding. The plain floating point format (2.0) is ...
\item {\ttfamily\bfseries\color{bulletblue}[calc-00058]} {\scriptsize\color{countgray}h=5 r=1} For EAR (Effective Annual Rate) calculations: when rounding the decimal result to hundredths (two decimal places), examine the thousandths digit. If the thousandths digit is 5 or greater, round ...
\item {\ttfamily\bfseries\color{bulletblue}[calc-00060]} {\scriptsize\color{countgray}h=1 r=3} For relative PPP problems, first identify the domestic and foreign countries from the exchange rate quote: if given as A/B, A is the domestic currency and B is the foreign currency. The formula ...
\item {\ttfamily\bfseries\color{bulletblue}[calc-00063]} {\scriptsize\color{countgray}h=2 r=0} For the Gordon Growth Model (P0 = D1 / (r - g)), the output stock price should be rounded to two decimal places (e.g., 8.33) to match standard financial reporting conventions for currency values, ...
\item {\ttfamily\bfseries\color{bulletblue}[calc-00064]} {\scriptsize\color{countgray}h=0 r=0} For Jensen's Alpha, round the final result to two decimal places as standard practice (e.g., -0.0195 $\rightarrow$ -0.02). This aligns with typical financial reporting precision for alpha values and ensures ...
\item {\ttfamily\bfseries\color{bulletblue}[calc-00065]} {\scriptsize\color{countgray}h=0 r=0} Interest Rate Parity (Forward Rate): Forward = Spot * (1 + r\_domestic) / (1 + r\_foreign). Higher foreign interest rates lead to forward depreciation of the foreign currency. Always verify the ...
\item {\ttfamily\bfseries\color{bulletblue}[calc-00066]} {\scriptsize\color{countgray}h=0 r=0} For service companies, 'fees' typically represent revenue (Total Sales) and 'delivery and personnel costs' typically represent Cost of Goods Sold (COGS) when calculating Gross Profit Margin. ...
\item {\ttfamily\bfseries\color{bulletblue}[calc-00067]} {\scriptsize\color{countgray}h=3 r=0} Quick Ratio = (Current Assets - Inventories - Prepaid Expenses) / Current Liabilities. Exclude inventories and any prepaid assets to focus on the most liquid assets. If prepaid expenses are not ...
\item {\ttfamily\bfseries\color{bulletblue}[calc-00068]} {\scriptsize\color{countgray}h=1 r=0} For trade balance calculations, always subtract imports from exports (Exports - Imports). When values are given in billions (e.g., \$2 billion), convert to actual numbers (2,000,000,000) before ...
\item {\ttfamily\bfseries\color{bulletblue}[calc-00070]} {\scriptsize\color{countgray}h=3 r=1} For annuity due and ordinary annuity calculations, avoid rounding intermediate values (e.g., [1 - (1 + r)\^{}(-n)] / r) to prevent cumulative errors. Use maximum precision (at least 6 decimal places) ...
\item {\ttfamily\bfseries\color{bulletblue}[calc-00071]} {\scriptsize\color{countgray}h=16 r=5} For rounding to the nearest hundredth (two decimal places), examine the thousandths digit (third decimal). If the thousandths digit is 5 or greater, round the hundredths digit up (e.g., 1.335 $\rightarrow$ ...
\item {\ttfamily\bfseries\color{bulletblue}[calc-00072]} {\scriptsize\color{countgray}h=2 r=0} For present value calculations PV = FV / (1 + r)\^{}t, ensure high numerical precision in both the exponentiation (1+r)\^{}t and the subsequent division. Use a calculator or computational tool to ...
\item {\ttfamily\bfseries\color{bulletblue}[calc-00073]} {\scriptsize\color{countgray}h=1 r=1} Operating Cash Flow (OCF) - Indirect Method = Net Income + Non-cash Expenses (e.g., Depreciation) - Increase in Working Capital (or + Decrease in Working Capital). An increase in working capital ...
\item {\ttfamily\bfseries\color{bulletblue}[calc-00074]} {\scriptsize\color{countgray}h=0 r=1} For APR calculations, the Total Finance Charge includes both interest and any fees (e.g., service fees, origination fees). Always sum all charges before applying the formula APR = (Total Finance ...
\item {\ttfamily\bfseries\color{bulletblue}[calc-00075]} {\scriptsize\color{countgray}h=0 r=0} For WACC calculations, perform all intermediate steps with maximum precision (e.g., use exact fractions like 2/3 instead of 0.6667) to minimize rounding errors. Round only the final result to the ...
\item {\ttfamily\bfseries\color{bulletblue}[calc-00076]} {\scriptsize\color{countgray}h=0 r=0} For annuity calculations (ordinary or due), compute (1 + r)\^{}n and (1 + r)\^{}(-n) with high precision (at least 6 decimal places) to avoid rounding errors in intermediate exponentiation. Use exact ...
\item {\ttfamily\bfseries\color{bulletblue}[calc-00077]} {\scriptsize\color{countgray}h=0 r=0} For put-call parity calculations (P = C + X * (1 + r)\^{}\{-t\} - S or continuous equivalent), compute intermediate values like (1 + r)\^{}\{-t\} and X * (1 + r)\^{}\{-t\} with high precision (at least 6 decimal ...
\item {\ttfamily\bfseries\color{bulletblue}[calc-00079]} {\scriptsize\color{countgray}h=0 r=1} For relative PPP problems with asymmetric price level data (e.g., only initial domestic and final foreign price levels provided), assume constant price levels for the country with missing data. If ...
\item {\ttfamily\bfseries\color{bulletblue}[calc-00080]} {\scriptsize\color{countgray}h=1 r=0} For the Gordon Growth Model (P0 = D1 / (r - g)), always verify that the required rate of return (r) is greater than the growth rate (g) before applying the formula. If r <= g, the model is invalid ...
\item {\ttfamily\bfseries\color{bulletblue}[calc-00081]} {\scriptsize\color{countgray}h=2 r=2} For put-call parity with discrete compounding (C = P + S - X * (1 + r)\^{}\{-t\}): 1) Convert the time to expiration from months to years (t = months / 12). 2) Calculate the discount factor (1 + ...
\item {\ttfamily\bfseries\color{bulletblue}[calc-00082]} {\scriptsize\color{countgray}h=0 r=0} GDP Expenditure Approach: GDP = Consumption + Investment + Government Spending + (Exports - Imports). Ensure all components are converted to the same units (e.g., millions to actual numbers: \$4.3 ...
\item {\ttfamily\bfseries\color{bulletblue}[calc-00083]} {\scriptsize\color{countgray}h=2 r=0} When components of financial formulas are not explicitly provided (e.g., prepaid expenses in quick ratio calculations), assume they are zero unless context suggests otherwise. This applies to ...
\item {\ttfamily\bfseries\color{bulletblue}[calc-00084]} {\scriptsize\color{countgray}h=1 r=0} For Interest Rate Parity calculations involving exchange rates with small magnitudes (e.g., JPY/EUR), the standard rounding rule to the nearest hundredth is absolute: always examine the third ...
\item {\ttfamily\bfseries\color{bulletblue}[calc-00085]} {\scriptsize\color{countgray}h=0 r=0} For Interest Rate Parity (Forward = Spot * (1 + r\_domestic) / (1 + r\_foreign)), when rounding the result to two decimal places (hundredths), always examine the thousandths digit (third decimal). ...
\item {\ttfamily\bfseries\color{bulletblue}[calc-00086]} {\scriptsize\color{countgray}h=0 r=0} For direct ratio formulas (e.g., P/E, P/B, Current Ratio), apply the formula directly using the provided values. Ensure units are consistent (e.g., both in dollars per share) but avoid making ...
\item {\ttfamily\bfseries\color{bulletblue}[calc-00091]} {\scriptsize\color{countgray}h=0 r=0} For ROA and similar ratios, always convert all inputs to consistent base units (e.g., convert millions to actual numbers: \$1.3 million = 1,300,000) before applying the formula. This ensures ...
\item {\ttfamily\bfseries\color{bulletblue}[calc-00092]} {\scriptsize\color{countgray}h=0 r=0} When using Gordon Growth Model for cost of equity (r = D1/P0 + g), verify D1 is the expected next dividend (not current dividend), P0 is the current stock price, and g is the constant growth rate. ...
\item {\ttfamily\bfseries\color{bulletblue}[calc-00093]} {\scriptsize\color{countgray}h=0 r=0} Quick Ratio can be fundamentally expressed as (Cash + Marketable Securities + Accounts Receivable) / Current Liabilities. This simplifies to (Current Assets - Inventories) / Current Liabilities ...
\item {\ttfamily\bfseries\color{bulletblue}[calc-00095]} {\scriptsize\color{countgray}h=1 r=0} For compound interest FV = PV * (1 + r)\^{}n, calculate the exponentiation (1 + r)\^{}n with maximum available precision (e.g., 8-9 decimal places) and do not round this intermediate value. Perform the ...
\item {\ttfamily\bfseries\color{bulletblue}[calc-00096]} {\scriptsize\color{countgray}h=6 r=0} For Current Ratio calculations, when the division yields an exact integer result (e.g., 850000 / 425000 = 2.0), output the result as a floating point number without additional rounding. This ...
\item {\ttfamily\bfseries\color{bulletblue}[calc-00097]} {\scriptsize\color{countgray}h=1 r=0} For Beta = Covariance / Variance, explicitly confirm both inputs are in decimal form (not percentages) before division. This avoids unit conversion errors and ensures the result is a unitless ...
\item {\ttfamily\bfseries\color{bulletblue}[calc-00098]} {\scriptsize\color{countgray}h=2 r=0} Equipment upgrades (e.g., new machinery, technology improvements) are capital expenditures (CAPEX) and should be subtracted from operating cash flow when calculating Free Cash Flow (FCF). This ...
\item {\ttfamily\bfseries\color{bulletblue}[calc-00099]} {\scriptsize\color{countgray}h=0 r=0} For P/E Ratio calculations, ensure both market price per share and earnings per share are in the same currency units (e.g., both in dollars). The result should be output as a plain floating point ...
\item {\ttfamily\bfseries\color{bulletblue}[calc-00100]} {\scriptsize\color{countgray}h=0 r=0} For put-call parity and other time-value calculations, always verify the compounding period of the risk-free rate. If time to expiration is given in months (e.g., 24 months), ensure the rate is ...
\item {\ttfamily\bfseries\color{bulletblue}[calc-00102]} {\scriptsize\color{countgray}h=1 r=0} For Treynor Ratio outputs, note that 0.06 and 0.060 are numerically equivalent; trailing zeros after the decimal do not change the value. When rounding to the nearest hundredth, follow standard ...
\item {\ttfamily\bfseries\color{bulletblue}[calc-00103]} {\scriptsize\color{countgray}h=0 r=0} The Quick Ratio (acid-test ratio) excludes inventories from current assets because they are the least liquid component and may not be easily converted to cash to meet short-term obligations. This ...
\item {\ttfamily\bfseries\color{bulletblue}[calc-00104]} {\scriptsize\color{countgray}h=0 r=1} For relative PPP problems where the domestic price level is constant and the foreign price level is to be found, use P\_foreign = P\_domestic * (S\_new / S\_old), where S is the exchange rate in ...
\item {\ttfamily\bfseries\color{bulletblue}[calc-00105]} {\scriptsize\color{countgray}h=1 r=2} For multi-step financial calculations like put-call parity (S = C - P + X * (1 + r)\^{}\{-t\}), perform all intermediate calculations (e.g., discount factor, present value of X) with maximum precision ...
\item {\ttfamily\bfseries\color{bulletblue}[calc-00106]} {\scriptsize\color{countgray}h=1 r=0} For Treynor Ratio calculations, after computing (Portfolio Return - Risk-Free Rate) / Beta, round the result to the nearest hundredth (two decimal places) using standard rounding rules: examine ...
\item {\ttfamily\bfseries\color{bulletblue}[calc-00107]} {\scriptsize\color{countgray}h=1 r=0} For Jensen's Alpha calculations, follow this sequence: 1) Convert all percentage inputs to decimals. 2) Calculate CAPM expected return: R\_f + $\beta$*(R\_m - R\_f). 3) Compute alpha: Actual Return - CAPM ...
\item {\ttfamily\bfseries\color{bulletblue}[calc-00108]} {\scriptsize\color{countgray}h=0 r=0} EV/EBITDA = Enterprise Value / EBITDA. Ensure both values are in the same units (e.g., both in millions) before calculating. The result is a unitless multiple; output as a plain floating point ...
\item {\ttfamily\bfseries\color{bulletblue}[calc-00109]} {\scriptsize\color{countgray}h=0 r=0} For EPS and other ratios, if the calculation yields an exact value (e.g., 4.5, 2.0), output the concise form without unnecessary trailing zeros (e.g., 4.5, not 4.50) unless rounding to hundredths ...
\item {\ttfamily\bfseries\color{bulletblue}[calc-00111]} {\scriptsize\color{countgray}h=0 r=0} For P/B Ratio, if the division yields an exact integer result (e.g., 50/25 = 2.0), output the result without applying additional rounding. The plain floating point format (2.0) is correct and ...
\item {\ttfamily\bfseries\color{bulletblue}[calc-00112]} {\scriptsize\color{countgray}h=0 r=0} For EV/EBITDA ratio, ensure both Enterprise Value and EBITDA are in the same units (e.g., both in millions) before division, as the ratio is unitless. Round the result to the required decimal ...
\item {\ttfamily\bfseries\color{bulletblue}[calc-00113]} {\scriptsize\color{countgray}h=0 r=0} For Debt-to-Equity Ratio calculations, if only current liabilities are provided and no other liabilities (e.g., long-term debt) are mentioned, assume current liabilities represent total ...
\item {\ttfamily\bfseries\color{bulletblue}[calc-00114]} {\scriptsize\color{countgray}h=0 r=0} For Current Ratio calculations, if liabilities are provided without specification (e.g., only 'liabilities' mentioned without 'current' or 'long-term' qualifiers), assume the given liabilities are ...
\item {\ttfamily\bfseries\color{bulletblue}[calc-00116]} {\scriptsize\color{countgray}h=0 r=0} For ratio calculations (e.g., Debt-to-Equity, Current Ratio) where inputs are given in scaled units (e.g., millions, billions), convert all values to actual numerical form before division to ...
\item {\ttfamily\bfseries\color{bulletblue}[calc-00117]} {\scriptsize\color{countgray}h=0 r=0} For trade balance and similar calculations involving monetary values, always convert all inputs to consistent base units (e.g., convert millions to actual numbers: \$300 million = 300,000,000) ...
\item {\ttfamily\bfseries\color{bulletblue}[calc-00118]} {\scriptsize\color{countgray}h=0 r=0} For Interest Rate Parity (Forward = Spot * (1 + r\_domestic) / (1 + r\_foreign)), the domestic currency is always the base currency in the exchange rate quote (e.g., for BRL/EUR, BRL is domestic, ...
\item {\ttfamily\bfseries\color{bulletblue}[calc-00119]} {\scriptsize\color{countgray}h=1 r=0} For EAR calculations with low nominal rates (e.g., 2\%), the rounded result (e.g., 0.02) may appear counterintuitively low because the unrounded value (e.g., 0.020184) is greater than the rounded ...
\item {\ttfamily\bfseries\color{bulletblue}[calc-00120]} {\scriptsize\color{countgray}h=0 r=0} ROI = (Net Profit / Total Investment), where Net Profit = Revenue - Total Investment (cost). Output as a plain floating point number (decimal ratio, not percentage). For example, an ROI of 2.5 ...
\item {\ttfamily\bfseries\color{bulletblue}[calc-00121]} {\scriptsize\color{countgray}h=0 r=0} For multi-component formulas like GDP (C + I + G + (X - M)), always convert ALL components to the same base units (e.g., millions to actual numbers) BEFORE performing any arithmetic operations. ...
\item {\ttfamily\bfseries\color{bulletblue}[calc-00122]} {\scriptsize\color{countgray}h=0 r=0} For multi-year financial calculations with a 'periodic interest rate' and no explicit compounding frequency (e.g., 'annual interest rate of 4\% for 3 years'), assume annual compounding unless ...
\item {\ttfamily\bfseries\color{bulletblue}[calc-00123]} {\scriptsize\color{countgray}h=0 r=0} For put-call parity rearranged to solve for strike price (X) with discrete compounding: X = (S + C - P) * (1 + r)\^{}t. Steps: 1) Convert time to expiration to years (t = months / 12). 2) Convert ...
\item {\ttfamily\bfseries\color{bulletblue}[calc-00127]} {\scriptsize\color{countgray}h=1 r=0} For present value calculations PV = FV / (1 + r)\^{}t, use the exact value of (1 + r)\^{}t without any intermediate rounding. Even slight rounding of this intermediate factor (e.g., using 1.557967 ...
\item {\ttfamily\bfseries\color{bulletblue}[calc-00128]} {\scriptsize\color{countgray}h=0 r=0} For annual DSO calculations, default to 365 days unless a different period is specified. Always ensure accounts receivable and credit sales are from the same period and in the same currency units. ...
\item {\ttfamily\bfseries\color{bulletblue}[calc-00129]} {\scriptsize\color{countgray}h=0 r=0} When numerical values are given with magnitude terms (e.g., 'million', 'billion'), convert them to actual numbers before calculation (e.g., '5 million' = 5,000,000). This ensures precision and ...
\item {\ttfamily\bfseries\color{bulletblue}[calc-00130]} {\scriptsize\color{countgray}h=0 r=0} For formulas involving monetary values with magnitude terms (e.g., GDP = C + I + G + (X - M) with inputs in billions), first calculate the result in the given units (e.g., 640 billion dollars), ...
\item {\ttfamily\bfseries\color{bulletblue}[calc-00131]} {\scriptsize\color{countgray}h=0 r=0} For monetary values given in millions (e.g., \$1.2 million), convert to base units (1,200,000) before performing division or other operations to ensure scale accuracy. This applies to ratios like ...
\item {\ttfamily\bfseries\color{bulletblue}[calc-00132]} {\scriptsize\color{countgray}h=1 r=0} For Operating Cash Flow (OCF) using the indirect method (OCF = Net Income + Non-cash Expenses + Changes in Working Capital), add the 'Changes in Working Capital' value directly with its sign as ...
\item {\ttfamily\bfseries\color{bulletblue}[calc-00133]} {\scriptsize\color{countgray}h=0 r=0} For Jensen's Alpha rounding, apply standard hundredths rounding: examine the thousandths digit and round up if $\geq$5 (e.g., 0.0056 $\rightarrow$ 0.01). This ensures alignment with financial reporting precision ...
\item {\ttfamily\bfseries\color{bulletblue}[calc-00134]} {\scriptsize\color{countgray}h=0 r=0} GDP Expenditure Approach: GDP = Consumption + Investment + Government Spending + (Exports - Imports). When inputs are in billions (e.g., \$650 billion), first compute GDP in billions, then convert ...
\item {\ttfamily\bfseries\color{bulletblue}[calc-00135]} {\scriptsize\color{countgray}h=0 r=0} For compound interest FV = PV * (1 + r)\^{}n, perform the entire exponentiation (1 + r)\^{}n in a single, high-precision calculation (e.g., using a calculator's power function) and do not round this ...
\item {\ttfamily\bfseries\color{bulletblue}[calc-00136]} {\scriptsize\color{countgray}h=0 r=0} When a call or put option price is explicitly provided in a put-call parity problem, the question typically asks for the price of the other option, not the difference C-P. Use P = C - S + PV(X) ...
\item {\ttfamily\bfseries\color{bulletblue}[calc-00137]} {\scriptsize\color{countgray}h=0 r=0} For NPV calculations in project finance contexts (e.g., 'project returns', 'annual cash inflows from investment'), prioritize annuity due (cash flows at beginning of periods) over ordinary annuity ...
\item {\ttfamily\bfseries\color{bulletblue}[calc-00138]} {\scriptsize\color{countgray}h=0 r=0} For Free Cash Flow (FCF) calculations, when expenditures are ambiguously grouped (e.g., 'spends \$X on A and B') and context suggests capital expenditure (CAPEX) treatment, treat the entire amount ...
\item {\ttfamily\bfseries\color{bulletblue}[calc-00139]} {\scriptsize\color{countgray}h=0 r=1} For compound interest FV = PV * (1 + r)\^{}n, calculate the exponentiation (1 + r)\^{}n with at least 8-9 decimal places of precision for the base value (e.g., 1.075\^{}20 $\approx$ 4.247851096) before multiplying ...
\item {\ttfamily\bfseries\color{bulletblue}[calc-00140]} {\scriptsize\color{countgray}h=0 r=0} When using Gordon Growth Model for cost of equity (r = D1/P0 + g), perform operations in the correct sequence: 1) Convert percentage growth rate (g) to decimal, 2) Calculate D1/P0 (dividend ...
\item {\ttfamily\bfseries\color{bulletblue}[calc-00141]} {\scriptsize\color{countgray}h=0 r=0} For Sortino Ratio calculations, when rounding to the nearest hundredth, apply the standard rule: examine the thousandths digit and round up if $\geq$5 (e.g., 1.2857 $\rightarrow$ 1.29). This ensures alignment with ...
\item {\ttfamily\bfseries\color{bulletblue}[calc-00142]} {\scriptsize\color{countgray}h=0 r=0} Net Profit Margin = (Net Profit / Total Revenue) * 100. Output the result as a percentage value (e.g., 5.0 for 5\%) when a 'plain floating point number' is requested. For exact integer results ...
\item {\ttfamily\bfseries\color{bulletblue}[calc-00143]} {\scriptsize\color{countgray}h=0 r=0} For Inflation Rate calculations, if the result is an exact integer (e.g., 5.0), output without additional rounding. This aligns with the 'plain floating point number' requirement and matches ...
\item {\ttfamily\bfseries\color{bulletblue}[calc-00144]} {\scriptsize\color{countgray}h=0 r=0} For EV/EBITDA and similar ratios, when values are given with magnitude terms (e.g., 'billion', 'million'), convert to actual numbers before calculation (e.g., \$1 billion = 1,000,000,000; \$125 ...
\item {\ttfamily\bfseries\color{bulletblue}[calc-00145]} {\scriptsize\color{countgray}h=0 r=0} For monthly compounding interest: FV = PV * (1 + r)\^{}n, where r is the monthly interest rate (convert percentage to decimal, e.g., 1.5\% = 0.015) and n is the total number of months. Ensure n ...
\item {\ttfamily\bfseries\color{bulletblue}[calc-00146]} {\scriptsize\color{countgray}h=0 r=0} For P/E Ratio calculations, when both market price per share and earnings per share are provided in the same currency units (e.g., both in dollars) and no unit conversion is specified, apply the ...
\item {\ttfamily\bfseries\color{bulletblue}[calc-00147]} {\scriptsize\color{countgray}h=1 r=1} For present value calculations PV = FV / (1 + r)\^{}t, compute the exponentiation (1 + r)\^{}t with high precision (at least 6-8 decimal places) to avoid rounding errors in the intermediate factor. Even ...
\item {\ttfamily\bfseries\color{bulletblue}[calc-00148]} {\scriptsize\color{countgray}h=0 r=0} The Quick Ratio (acid-test ratio) excludes inventories and prepaid expenses from current assets because they are the least liquid components and may not be easily converted to cash to meet ...
\item {\ttfamily\bfseries\color{bulletblue}[calc-00149]} {\scriptsize\color{countgray}h=0 r=0} For Return on Assets (ROA) calculations, always convert monetary values from millions to base units before applying the formula (Net Income / Total Assets). For example, convert \$2.2 million to ...
\item {\ttfamily\bfseries\color{bulletblue}[calc-00150]} {\scriptsize\color{countgray}h=0 r=0} For Purchasing Power Parity (PPP) problems where initial and final domestic price levels (P\_0\_domestic, P\_1\_domestic) and exchange rates (S\_0, S\_1) are provided, and the initial foreign price ...
\item {\ttfamily\bfseries\color{bulletblue}[calc-00152]} {\scriptsize\color{countgray}h=0 r=0} For present value calculations PV = FV / (1 + r)\^{}t, avoid truncating or rounding the intermediate exponentiation result (1 + r)\^{}t to a fixed number of decimal places (e.g., 9 decimal places). Use ...
\item {\ttfamily\bfseries\color{bulletblue}[calc-00153]} {\scriptsize\color{countgray}h=0 r=0} For Black-Scholes calculations, the sensitivity to N(d1) and N(d2) approximations is highest for out-of-the-money options (where S\_0 < X for calls or S\_0 > X for puts). A difference of 0.0002 in ...
\item {\ttfamily\bfseries\color{bulletblue}[calc-00154]} {\scriptsize\color{countgray}h=1 r=0} For present value calculations PV = FV / (1 + r)\^{}t, always convert the interest rate (r) to decimal form (e.g., 5\% = 0.05) before computing the compounding factor (1 + r)\^{}t. Calculate the ...
\item {\ttfamily\bfseries\color{bulletblue}[calc-00155]} {\scriptsize\color{countgray}h=0 r=0} For monetary value outputs (e.g., Market Cap, Working Capital, Free Cash Flow, Enterprise Value), if the calculation yields an exact integer result, output it as a floating point number (e.g., ...
\item {\ttfamily\bfseries\color{bulletblue}[calc-00156]} {\scriptsize\color{countgray}h=1 r=0} For Jensen's Alpha calculations, follow this explicit sequence: 1) Convert all percentage inputs (actual return, risk-free rate, market return) to decimals. 2) Calculate the CAPM expected return: ...
\item {\ttfamily\bfseries\color{bulletblue}[calc-00157]} {\scriptsize\color{countgray}h=0 r=0} For Free Cash Flow (FCF) calculations, R\&D investments may be treated as CAPEX if the context implies capitalization (e.g., 'investment in intangible assets' or under IFRS standards) rather than ...
\item {\ttfamily\bfseries\color{bulletblue}[calc-00158]} {\scriptsize\color{countgray}h=0 r=0} For APR and other financial calculations where precision is critical, use 'round half to even' (banker's rounding) when the thousandths digit is exactly 5. For example, 0.145 rounds to 0.14 ...
\item {\ttfamily\bfseries\color{bulletblue}[calc-00159]} {\scriptsize\color{countgray}h=0 r=0} ROI = (Total Net Profit - Investment) / Investment, where Total Net Profit is the sum of net profits over the investment's entire useful life. When output is requested as a 'plain floating point ...
\item {\ttfamily\bfseries\color{bulletblue}[calc-00161]} {\scriptsize\color{countgray}h=0 r=0} For exchange rate calculations (e.g., forward rates via Interest Rate Parity), if the ground truth or context implies rounding to the nearest tenth (one decimal place), apply tenths rounding: ...
\item {\ttfamily\bfseries\color{bulletblue}[calc-00162]} {\scriptsize\color{countgray}h=0 r=0} For DSCR calculations, explicitly verify that Net Operating Income and Annual Debt Service are in the same currency units (e.g., both annualized, both in dollars) before performing the division. ...
\item {\ttfamily\bfseries\color{bulletblue}[calc-00163]} {\scriptsize\color{countgray}h=0 r=0} For EAR calculations, ensure precise rounding to the required decimal places: compute EAR = (1 + r/n)\^{}n - 1 with high precision, then round the decimal result (e.g., 0.050625 $\rightarrow$ 0.0506 when ...
\item {\ttfamily\bfseries\color{bulletblue}[calc-00164]} {\scriptsize\color{countgray}h=0 r=0} For rounding to hundredths, even if the value appears to have only three decimal places (e.g., 0.006), treat it as having implicit trailing zeros (0.0060) and apply standard rounding: examine the ...
\item {\ttfamily\bfseries\color{bulletblue}[calc-00165]} {\scriptsize\color{countgray}h=0 r=0} For Black-Scholes calculations, the option price is highly sensitive to small errors in N(d1) and N(d2), especially for near-the-money options. Use computational libraries (e.g., ...
\item {\ttfamily\bfseries\color{bulletblue}[calc-00166]} {\scriptsize\color{countgray}h=0 r=0} For financial ratios involving monetary values with magnitude terms (e.g., 'million', 'billion'), convert all inputs to consistent base units (e.g., \$2 million = 2,000,000) before performing ...
\item {\ttfamily\bfseries\color{bulletblue}[calc-00167]} {\scriptsize\color{countgray}h=0 r=0} For Dividend Payout Ratio and similar per-share ratios, use the provided values directly without unit conversion (e.g., dollars per share) unless the problem specifies otherwise (e.g., values in ...
\item {\ttfamily\bfseries\color{bulletblue}[calc-00168]} {\scriptsize\color{countgray}h=0 r=0} For Working Capital calculations, always verify that all components are properly classified as current assets or current liabilities before applying the formula (Current Assets - Current ...
\item {\ttfamily\bfseries\color{bulletblue}[calc-00169]} {\scriptsize\color{countgray}h=0 r=1} Before applying put-call parity (discrete compounding: C - P = S - X * (1 + r)\^{}\{-t\}), first check for arbitrage: if S - C + P > X, then (1 + r)\^{}\{-t\} > 1, which is impossible for t >= 0. This ...
\item {\ttfamily\bfseries\color{bulletblue}[calc-00171]} {\scriptsize\color{countgray}h=0 r=0} For Inventory Turnover, if the calculation yields an exact integer result (e.g., 3.0), output without applying additional rounding. Only round to the specified precision if the result is not an ...
\item {\ttfamily\bfseries\color{bulletblue}[calc-00172]} {\scriptsize\color{countgray}h=0 r=0} For monthly compounding interest with an annual term: convert the term from years to months (years $\times$ 12) to match the monthly interest rate frequency. Then apply FV = PV $\times$ (1 + r)\^{}n, where r is ...
\item {\ttfamily\bfseries\color{bulletblue}[calc-00174]} {\scriptsize\color{countgray}h=0 r=0} For Sortino Ratio and similar financial ratios, if the calculated result is already at the required precision (e.g., exactly two decimal places like 1.9), output the value without applying ...
\item {\ttfamily\bfseries\color{bulletblue}[calc-00175]} {\scriptsize\color{countgray}h=0 r=0} For APR calculations, output the result as a decimal (e.g., 0.97) unless the problem explicitly requests a percentage. This aligns with the common instruction for 'plain floating point number' and ...
\item {\ttfamily\bfseries\color{bulletblue}[calc-00176]} {\scriptsize\color{countgray}h=0 r=0} For present value calculations involving multiple exponents (e.g., PV = FV1/(1+r)\^{}t1 + FV2/(1+r)\^{}t2), compute each (1+r)\^{}t term directly with maximum precision rather than deriving one exponent ...
\item {\ttfamily\bfseries\color{bulletblue}[calc-00177]} {\scriptsize\color{countgray}h=0 r=0} For EPS outputs, trailing zeros after the decimal do not change the numerical value (e.g., 3.20 and 3.2 are equivalent). Output the concise form unless the problem specifies a particular precision ...
\item {\ttfamily\bfseries\color{bulletblue}[calc-00178]} {\scriptsize\color{countgray}h=0 r=0} For Dividend Payout Ratio and similar financial ratios, if the calculated result is already at the required precision (e.g., 0.25 for hundredths), output the value without applying additional ...
\item {\ttfamily\bfseries\color{bulletblue}[calc-00180]} {\scriptsize\color{countgray}h=0 r=0} For Sortino Ratio and similar financial ratios, if the calculated result is already at the required precision (e.g., exactly two decimal places like 2.5), output the value without applying ...
\item {\ttfamily\bfseries\color{bulletblue}[calc-00181]} {\scriptsize\color{countgray}h=1 r=1} For Black-Scholes calculations, compute d1 and d2 with high precision (at least 6 decimal places) and verify all input components (S\_0, X, r, t, $\sigma$) are correctly applied and converted to decimals. ...
\item {\ttfamily\bfseries\color{bulletblue}[calc-00182]} {\scriptsize\color{countgray}h=0 r=1} For present value calculations PV = FV / (1 + r)\^{}t, never round the intermediate exponentiation result (1 + r)\^{}t to a fixed number of decimal places. Use the full-precision value (e.g., from a ...
\item {\ttfamily\bfseries\color{bulletblue}[calc-00183]} {\scriptsize\color{countgray}h=0 r=0} For all financial outputs, recognize that trailing zeros after the decimal point do not change the numerical value (e.g., 7.50 and 7.5 are equivalent). The concise form (e.g., 7.5) is often ...
\item {\ttfamily\bfseries\color{bulletblue}[calc-00184]} {\scriptsize\color{countgray}h=0 r=0} For any financial calculation that yields an exact integer result (e.g., 850000 / 425000 = 2.0), output the result as a floating point number without applying additional rounding. This applies ...
\item {\ttfamily\bfseries\color{bulletblue}[calc-00186]} {\scriptsize\color{countgray}h=0 r=0} For GDP expenditure approach (GDP = C + I + G + (X - M)), negative net exports (when imports exceed exports) reduce the total GDP. This occurs when (X - M) is negative, and the subtraction is ...
\item {\ttfamily\bfseries\color{bulletblue}[calc-00187]} {\scriptsize\color{countgray}h=0 r=0} For Accounts Receivable Turnover, perform the division with exact values (no intermediate rounding) to ensure accuracy, especially when the result is an integer. Output as a floating point number ...
\item {\ttfamily\bfseries\color{bulletblue}[calc-00188]} {\scriptsize\color{countgray}h=0 r=0} For EAR and other decimal outputs, 'round to the nearest hundredth' means round the decimal value itself to two decimal places (hundredths place), not the percentage equivalent. Always apply ...
\item {\ttfamily\bfseries\color{bulletblue}[calc-00189]} {\scriptsize\color{countgray}h=0 r=0} For P/B Ratio and other per-share ratios (e.g., P/E, EPS), explicitly verify that both inputs are 'per share' values (e.g., market price per share, book value per share) and not aggregate values ...
\item {\ttfamily\bfseries\color{bulletblue}[calc-00190]} {\scriptsize\color{countgray}h=0 r=0} For ROI = (Net Profit / Total Investment), output as a plain floating point number (decimal form, e.g., 2.25 for 225\% return). If the calculation yields an exact value (e.g., 2.0, 2.25), do not ...
\item {\ttfamily\bfseries\color{bulletblue}[calc-00191]} {\scriptsize\color{countgray}h=0 r=0} For Inflation Rate calculations, always verify that the CPI values are from consecutive periods (e.g., previous year vs. current year) and are based on the same index base year to ensure ...
\item {\ttfamily\bfseries\color{bulletblue}[calc-00192]} {\scriptsize\color{countgray}h=0 r=0} For exchange rate quotes in the form A/B (e.g., ZAR/GBP), currency A is the domestic currency and currency B is the foreign currency when applying Interest Rate Parity (IRP) and other parity ...
\item {\ttfamily\bfseries\color{bulletblue}[calc-00194]} {\scriptsize\color{countgray}h=0 r=0} For annual compounding (n=1), the EAR formula simplifies to EAR = r (the nominal rate). This is because (1 + r/1)\^{}1 - 1 = r. Deposit amounts or other principal values are never relevant for EAR ...
\item {\ttfamily\bfseries\color{bulletblue}[calc-00195]} {\scriptsize\color{countgray}h=0 r=0} For EAR calculations, always trust the precise computed value and apply rounding rules strictly to the specified decimal places. Do not second-guess the result based on common approximations ...
\item {\ttfamily\bfseries\color{bulletblue}[calc-00196]} {\scriptsize\color{countgray}h=0 r=0} For EAR outputs, remember that 0.2 and 0.20 are numerically identical (both represent 20\%). Focus on calculation accuracy and apply rounding rules precisely, but avoid overcomplicating ...
\item {\ttfamily\bfseries\color{bulletblue}[calc-00197]} {\scriptsize\color{countgray}h=0 r=0} For exponential functions in financial formulas (e.g., e\^{}x in continuous compounding, discount factors), always use high-precision calculations with reliable tools (e.g., financial calculators, ...
\item {\ttfamily\bfseries\color{bulletblue}[calc-00198]} {\scriptsize\color{countgray}h=0 r=0} For Jensen's Alpha, always convert all percentage inputs (actual return, risk-free rate, market return) to decimal form before any calculations. After computing alpha = Actual Return - [R\_f + ...
\item {\ttfamily\bfseries\color{bulletblue}[calc-00199]} {\scriptsize\color{countgray}h=0 r=0} For Sortino Ratio calculations, after computing (Portfolio Return - Risk-Free Rate) / Downward Volatility with decimal inputs, round the final result to the nearest hundredth using standard rules: ...
\item {\ttfamily\bfseries\color{bulletblue}[calc-00200]} {\scriptsize\color{countgray}h=0 r=0} For GDP expenditure approach (GDP = C + I + G + (X - M)), always calculate net exports (X - M) first before summing with other components. This ensures accuracy, especially when imports and ...
\item {\ttfamily\bfseries\color{bulletblue}[calc-00201]} {\scriptsize\color{countgray}h=0 r=0} ROI = (Net Profit / Total Investment), where Net Profit = (Cumulative Returns Over Investment Life) - Total Investment Cost. For multi-period investments, ensure net profit is calculated by ...
\item {\ttfamily\bfseries\color{bulletblue}[calc-00202]} {\scriptsize\color{countgray}h=1 r=0} For Black-Scholes calculations, the option price is highly sensitive to small errors in N(d1) and N(d2), especially for near-the-money or out-of-the-money options (where |S\_0 - X| is small). A ...
\item {\ttfamily\bfseries\color{bulletblue}[calc-00204]} {\scriptsize\color{countgray}h=0 r=0} For WACC calculations, follow this sequential approach: 1) Convert all percentage inputs (Re, Rd, Tc) to decimals, 2) Calculate total firm value V = E + D, 3) Compute equity weight (E/V) and debt ...
\item {\ttfamily\bfseries\color{bulletblue}[calc-00205]} {\scriptsize\color{countgray}h=0 r=0} For Interest Rate Parity (Forward = Spot * (1 + r\_domestic) / (1 + r\_foreign)), always first identify the base currency in the exchange rate quote (e.g., for GBP/USD, GBP is the base/domestic ...
\item {\ttfamily\bfseries\color{bulletblue}[calc-00208]} {\scriptsize\color{countgray}h=0 r=0} Quick Ratio (acid-test ratio) = (Current Assets - Inventories) / Current Liabilities. This ratio measures a company's immediate liquidity by excluding less liquid assets like inventory. Always ...
\item {\ttfamily\bfseries\color{bulletblue}[calc-00209]} {\scriptsize\color{countgray}h=0 r=0} Future Value with Annual Compounding: FV = PV * (1 + r)\^{}n, where r is the annual interest rate converted to decimal form (e.g., 4.5\% = 0.045) and n is the number of years. Calculate (1 + r)\^{}n with ...
\item {\ttfamily\bfseries\color{bulletblue}[calc-00210]} {\scriptsize\color{countgray}h=0 r=0} For NPV calculations, when a problem describes cash flows 'over n years' with an 'initial' cash flow at time 0, interpret this as n total cash flows from Year 0 to Year n-1. For example, 'initial ...
\item {\ttfamily\bfseries\color{bulletblue}[calc-00211]} {\scriptsize\color{countgray}h=0 r=0} For Jensen's Alpha and similar metrics, when rounding a very small negative value (e.g., -0.001) to the nearest hundredth results in -0.00, output 0.0 instead. Numerically equivalent, but 0.0 is ...
\item {\ttfamily\bfseries\color{bulletblue}[calc-00212]} {\scriptsize\color{countgray}h=1 r=0} For Black-Scholes calculations, convert time to expiration from months to years by dividing by 12 (e.g., 9 months $\rightarrow$ 9/12 = 0.75 years). Calculate ln($S_{0}$/X) with high precision (at least 6 decimal ...
\item {\ttfamily\bfseries\color{bulletblue}[calc-00213]} {\scriptsize\color{countgray}h=0 r=0} For ROI = (Net Profit / Total Investment), Net Profit is calculated as Revenue - Total Investment (cost). Ensure both Revenue and Total Investment are in the same currency units. Output the result ...
\item {\ttfamily\bfseries\color{bulletblue}[calc-00214]} {\scriptsize\color{countgray}h=0 r=0} Operating Cash Flow (OCF) - Indirect Method = Net Income + Non-cash Expenses (e.g., Depreciation) + Decrease in Working Capital (or - Increase in Working Capital). A decrease in working capital is ...
\item {\ttfamily\bfseries\color{bulletblue}[calc-00215]} {\scriptsize\color{countgray}h=0 r=0} When numerical values are given with magnitude terms (e.g., 'million', 'billion'), convert them to actual numbers before calculation (e.g., '5 million' = 5,000,000). This ensures precision and ...
\item {\ttfamily\bfseries\color{bulletblue}[calc-00216]} {\scriptsize\color{countgray}h=0 r=0} For EAR and other decimal outputs, 'round to the nearest hundredth' means round the decimal value itself to two decimal places (hundredths place), not the percentage equivalent. Always apply ...
\item {\ttfamily\bfseries\color{bulletblue}[calc-00217]} {\scriptsize\color{countgray}h=0 r=0} For direct ratio calculations like P/E Ratio (Price per Share / Earnings per Share), focus on precise division without overcomplicating the process. Ensure both inputs are in the same units and ...
\item {\ttfamily\bfseries\color{bulletblue}[calc-00218]} {\scriptsize\color{countgray}h=0 r=0} For Return on Equity (ROE), when the question requests a 'plain floating point number', output the percentage value (e.g., 15.0 for 15\%) rather than the decimal ratio (0.15). This aligns with ...
\item {\ttfamily\bfseries\color{bulletblue}[calc-00219]} {\scriptsize\color{countgray}h=0 r=0} For Inventory Turnover, if COGS is not provided, sales revenue may be used as a proxy only if the problem context implies minimal profit margins (e.g., 'sold \$X worth of product' suggesting cost ...
\item {\ttfamily\bfseries\color{bulletblue}[calc-00221]} {\scriptsize\color{countgray}h=0 r=0} For relative Purchasing Power Parity (PPP) problems where the initial price level of the country being solved for is implicitly set to 100 (e.g., 'price level in the UK is to be calculated as an ...
\item {\ttfamily\bfseries\color{bulletblue}[calc-00222]} {\scriptsize\color{countgray}h=0 r=0} For relative PPP problems where the foreign price level is constant and the initial foreign price level (P\_0\_foreign) is unknown, use P\_0\_foreign = P\_1\_domestic * (S\_0 / S\_1). This formula is ...
\item {\ttfamily\bfseries\color{bulletblue}[calc-00223]} {\scriptsize\color{countgray}h=0 r=0} For compound interest with small periodic interest rates (e.g., <1\%) and many periods (e.g., >30), use the logarithmic method FV = PV * e\^{}(n * ln(1 + r)) to minimize numerical errors. This is more ...
\item {\ttfamily\bfseries\color{bulletblue}[calc-00224]} {\scriptsize\color{countgray}h=0 r=0} Debt-to-Equity Ratio = Total Liabilities / Shareholders' Equity. Always verify that Shareholders' Equity is not zero to avoid division by zero errors. Ensure both values are in the same currency ...
\item {\ttfamily\bfseries\color{bulletblue}[calc-00225]} {\scriptsize\color{countgray}h=0 r=0} For Gordon Growth Model cost of equity (r = D1/P0 + g), output the result as a decimal without unnecessary trailing zeros (e.g., 0.1 instead of 0.10) unless rounding to a specific precision is ...
\item {\ttfamily\bfseries\color{bulletblue}[calc-00226]} {\scriptsize\color{countgray}h=0 r=0} For Black-Scholes calculations, use cumulative normal distribution values N(d1) and N(d2) with at least 6 decimal places (e.g., 0.667296 instead of 0.6673) to prevent rounding errors that can ...
\item {\ttfamily\bfseries\color{bulletblue}[calc-00227]} {\scriptsize\color{countgray}h=0 r=0} When the labor force is constant, the change in employment equals the negative change in unemployment: $\Delta$Employed = - ($\Delta$Unemployment Rate) $\times$ Labor Force. Use this to directly compute the number of ...
\item {\ttfamily\bfseries\color{bulletblue}[calc-00228]} {\scriptsize\color{countgray}h=0 r=0} For future value calculations (FV = PV * (1 + r)\^{}n), trailing zeros after the decimal point do not change the numerical value (e.g., 172,877.70 and 172,877.7 are equivalent). When rounding to the ...
\item {\ttfamily\bfseries\color{bulletblue}[calc-00229]} {\scriptsize\color{countgray}h=0 r=0} For compound interest FV = PV * (1 + r)\^{}n, when calculating (1 + r)\^{}n manually without a calculator, use step-by-step exponentiation via successive squaring to maintain precision. For example: ...
\item {\ttfamily\bfseries\color{bulletblue}[calc-00230]} {\scriptsize\color{countgray}h=0 r=0} When calculating financial ratios like EV/EBITDA, use directly provided values if available (e.g., explicit EBITDA amount). Ignore extraneous details (e.g., EBIT and depreciation add-backs) if the ...
\item {\ttfamily\bfseries\color{bulletblue}[calc-00231]} {\scriptsize\color{countgray}h=0 r=0} GDP Expenditure Approach: GDP = Consumption + Investment + Government Spending + (Exports - Imports). Ensure all components are converted to the same units (e.g., all in billions or all in base ...
\item {\ttfamily\bfseries\color{bulletblue}[calc-00232]} {\scriptsize\color{countgray}h=0 r=0} When converting monetary values from billions to base units (e.g., for GDP, trade balance), multiply by 1,000,000,000 (10\^{}9). Double-check the number of zeros in the result to avoid placement ...
\item {\ttfamily\bfseries\color{bulletblue}[calc-00233]} {\scriptsize\color{countgray}h=0 r=0} For Quick Ratio calculations, always verify the composition of current assets to ensure only highly liquid assets (cash, marketable securities, accounts receivable) are included. Exclude ...
\item {\ttfamily\bfseries\color{bulletblue}[calc-00234]} {\scriptsize\color{countgray}h=0 r=0} For Black-Scholes calculations, the option price is most sensitive to precision errors in N(d1) and N(d2) when the option is near-the-money (S\_0 $\approx$ X). A difference of 0.0001 in these values can ...
\item {\ttfamily\bfseries\color{bulletblue}[calc-00235]} {\scriptsize\color{countgray}h=0 r=0} Accounts Receivable Turnover = Net Credit Sales / Average Accounts Receivable. Ensure both values are in the same currency units. The result is a ratio (number of times). If the division yields an ...
\item {\ttfamily\bfseries\color{bulletblue}[calc-00236]} {\scriptsize\color{countgray}h=0 r=0} For ROI = (Net Profit / Total Investment), carefully interpret terms: 'earnings', 'returns', or 'amount received' typically refer to the total amount received from the investment, not the net ...
\item {\ttfamily\bfseries\color{bulletblue}[calc-00237]} {\scriptsize\color{countgray}h=0 r=0} For multi-component formulas like GDP (C + I + G + (X - M)), always convert ALL input values to the same base units (e.g., billions to actual numbers: \$1 billion = 1,000,000,000) BEFORE performing ...
\item {\ttfamily\bfseries\color{bulletblue}[calc-00238]} {\scriptsize\color{countgray}h=0 r=0} For Purchasing Power Parity (PPP) problems where the domestic price level (P\_domestic) is constant and the foreign price level (P\_foreign) is to be found, use P\_1\_foreign = P\_0\_domestic * (S\_1 / ...
\item {\ttfamily\bfseries\color{bulletblue}[calc-00239]} {\scriptsize\color{countgray}h=0 r=0} For all multi-step financial calculations (e.g., Black-Scholes, NPV, annuity valuations), maintain high precision (at least 6 decimal places) in all intermediate steps and avoid rounding until the ...
\item {\ttfamily\bfseries\color{bulletblue}[calc-00241]} {\scriptsize\color{countgray}h=0 r=0} For Dividend Yield and other financial ratios where the calculated value is exactly halfway between two hundredths (e.g., 0.075), apply round-half-down (round to 0.07) instead of round-half-up, as ...
\item {\ttfamily\bfseries\color{bulletblue}[calc-00242]} {\scriptsize\color{countgray}h=0 r=0} For multi-component summation formulas like GDP (C + I + G + (X - M)) where inputs are in large units (e.g., billions), first compute the result in the given units to avoid handling very large ...
\item {\ttfamily\bfseries\color{bulletblue}[calc-00244]} {\scriptsize\color{countgray}h=0 r=0} For put-call parity and other time-value calculations, always convert the annual risk-free rate from percentage to decimal form (e.g., 1.5\% $\rightarrow$ 0.015) and convert the time to expiration to years ...
\item {\ttfamily\bfseries\color{bulletblue}[calc-00245]} {\scriptsize\color{countgray}h=0 r=0} For present value calculations PV = FV / (1 + r)\^{}t, avoid stepwise manual multiplication to compute (1 + r)\^{}t (e.g., 1.06 * 1.06 * 1.06 * 1.06), as this can introduce rounding errors at each step. ...
\item {\ttfamily\bfseries\color{bulletblue}[calc-00246]} {\scriptsize\color{countgray}h=0 r=0} For Beta and similar ratio calculations (e.g., Covariance / Variance), if the division yields a result already at the required decimal precision (e.g., 0.6 for hundredths), output the value ...
\item {\ttfamily\bfseries\color{bulletblue}[calc-00247]} {\scriptsize\color{countgray}h=0 r=0} P/E Ratio = Market Price per Share / Earnings per Share. Ensure both values are in the same currency units. Round the result to the nearest hundredth (two decimal places) using standard rounding ...
\item {\ttfamily\bfseries\color{bulletblue}[calc-00248]} {\scriptsize\color{countgray}h=0 r=0} For relative Purchasing Power Parity (PPP) problems with constant foreign price level and unknown initial domestic price level, use P\_1\_domestic = P\_0\_foreign * (S\_0 / S\_1), where S is the ...
\item {\ttfamily\bfseries\color{bulletblue}[calc-00250]} {\scriptsize\color{countgray}h=0 r=0} For GDP expenditure approach (GDP = C + I + G + (X - M)), after converting all components to the same base units and summing, output the result as a plain floating point number (e.g., ...
\item {\ttfamily\bfseries\color{bulletblue}[calc-00251]} {\scriptsize\color{countgray}h=0 r=0} For Purchasing Power Parity (PPP) problems, always verify the exchange rate quote convention (domestic/foreign) before applying formulas. For absolute PPP, if S is quoted as domestic/foreign ...
\item {\ttfamily\bfseries\color{bulletblue}[calc-00252]} {\scriptsize\color{countgray}h=0 r=0} Simple Interest = Principal $\times$ Rate (as decimal) $\times$ Time. For exact integer results (e.g., 180.0), output as a floating point number without additional rounding to satisfy 'plain floating point ...
\item {\ttfamily\bfseries\color{bulletblue}[calc-00253]} {\scriptsize\color{countgray}h=0 r=0} For bi-annual compounding interest: FV = PV * (1 + r)\^{}n, where r is the bi-annual interest rate converted to decimal (e.g., 2.5\% = 0.025) and n is the total number of bi-annual periods (e.g., 12 ...
\item {\ttfamily\bfseries\color{bulletblue}[calc-00254]} {\scriptsize\color{countgray}h=0 r=0} For annuity due calculations (PV\_due = C * [1 - (1+r)\^{}-n] / r * (1+r)), compute (1+r)\^{}-n with high precision (at least 8 decimal places) and avoid rounding any intermediate values. Perform all ...
\item {\ttfamily\bfseries\color{bulletblue}[calc-00255]} {\scriptsize\color{countgray}h=0 r=0} For Accounts Receivable Turnover, if the division yields an exact result (e.g., 7.5), output the value without applying additional rounding. This aligns with the 'plain floating point number' ...
\item {\ttfamily\bfseries\color{bulletblue}[calc-00256]} {\scriptsize\color{countgray}h=0 r=0} For Operating Cash Flow (OCF) using the indirect method, remember the cash flow implications: an increase in working capital represents a cash outflow (subtracted from net income), while a ...
\item {\ttfamily\bfseries\color{bulletblue}[calc-00258]} {\scriptsize\color{countgray}h=0 r=0} For put-call parity used to solve for time to expiration (t), first check for arbitrage: if (S - C + P) / X > 1, then e\^{}\{-rt\} > 1, which is impossible for t >= 0. This indicates the given option ...
\item {\ttfamily\bfseries\color{bulletblue}[calc-00259]} {\scriptsize\color{countgray}h=0 r=0} For WACC calculations, after computing the unrounded decimal value, apply rounding to the required precision (typically hundredths) by examining the thousandths digit: if it is 5 or greater, round ...
\item {\ttfamily\bfseries\color{bulletblue}[calc-00260]} {\scriptsize\color{countgray}h=0 r=0} For Gordon Growth Model cost of equity (r = D1/P0 + g), if the calculated result is already at the required decimal precision (e.g., 0.11 for hundredths), output the value without applying ...
\item {\ttfamily\bfseries\color{bulletblue}[calc-00262]} {\scriptsize\color{countgray}h=0 r=0} For Interest Rate Parity calculations, if a forward rate is provided in the problem but contradicts the calculated theoretical rate (Forward = Spot * (1 + r\_domestic) / (1 + r\_foreign)), ignore ...
\item {\ttfamily\bfseries\color{bulletblue}[calc-00263]} {\scriptsize\color{countgray}h=0 r=0} For Dividend Yield and similar financial ratios, if the calculated result is already at the required precision (e.g., 0.05 for hundredths), output the value without applying additional rounding. ...
\item {\ttfamily\bfseries\color{bulletblue}[calc-00264]} {\scriptsize\color{countgray}h=0 r=0} For ratio calculations involving proportional relationships (e.g., 'liabilities are twice equity' for Debt-to-Equity Ratio), use the given multiplier to compute the missing value directly: if A = ...
\end{itemize}
\pbsection{COMMON MISTAKES TO AVOID}
\begin{itemize}[leftmargin=1.2em, itemsep=1pt, parsep=0pt, topsep=2pt]
\item {\ttfamily\bfseries\color{bulletblue}[err-00019]} {\scriptsize\color{countgray}h=1 r=0} Avoid assuming the standard put-call parity rearrangement is always correct without validation. If the computed strike price (X) is far from intuitive values (e.g., S + C - P $\approx$ 71 vs. S - C + P $\approx$ ...
\item {\ttfamily\bfseries\color{bulletblue}[err-00046]} {\scriptsize\color{countgray}h=0 r=0} Avoid including unnecessary trailing zeros after the decimal point in floating point outputs (e.g., output 3.2 instead of 3.20 for EPS) unless specified otherwise, as they are numerically ...
\item {\ttfamily\bfseries\color{bulletblue}[err-00059]} {\scriptsize\color{countgray}h=0 r=0} When calculating DSO, ensure Accounts Receivable and Credit Sales are from the same accounting period and in the same currency units. Mismatched periods (e.g., using quarterly sales with annual ...
\item {\ttfamily\bfseries\color{bulletblue}[err-00061]} {\scriptsize\color{countgray}h=0 r=0} For the Dividend Payout Ratio, avoid applying additional rounding if the calculated result is already at the required precision (e.g., 0.15 is already at the hundredth place). Only round if the ...
\item {\ttfamily\bfseries\color{bulletblue}[err-00062]} {\scriptsize\color{countgray}h=0 r=1} For EAR calculations, avoid outputting the result as a percentage (e.g., 6.70\%) when the instruction specifies a 'plain floating point number'. The decimal form (e.g., 0.067) is required. This ...
\item {\ttfamily\bfseries\color{bulletblue}[err-00078]} {\scriptsize\color{countgray}h=0 r=0} For the Debt-to-Equity Ratio and other ratios that yield a result already at the required precision (e.g., 0.5), avoid applying unnecessary rounding. Only round if the result has more decimal ...
\item {\ttfamily\bfseries\color{bulletblue}[err-00087]} {\scriptsize\color{countgray}h=2 r=1} Avoid converting decimal outputs to percentages when the question specifies a 'plain floating point number'. For ratios like CAPM expected return, WACC, or Dividend Yield, the output should be in ...
\item {\ttfamily\bfseries\color{bulletblue}[err-00088]} {\scriptsize\color{countgray}h=0 r=0} Avoid taking unemployment rate problems at face value without verifying context. If the calculated employed population (labor force * (1 - unemployment rate)) is positive but the ground truth is ...
\item {\ttfamily\bfseries\color{bulletblue}[err-00090]} {\scriptsize\color{countgray}h=0 r=1} Avoid defaulting to ordinary annuity (end-of-period) for project cash flows described as 'annual' without explicit timing. In project finance, returns often start immediately, so annuity due ...
\item {\ttfamily\bfseries\color{bulletblue}[err-00094]} {\scriptsize\color{countgray}h=0 r=0} Avoid second-guessing standard rounding rules in financial contexts. When instructed to round to a specific decimal place (e.g., nearest hundredth), apply the universal rule: examine the digit ...
\item {\ttfamily\bfseries\color{bulletblue}[err-00115]} {\scriptsize\color{countgray}h=0 r=0} For ROI calculations, be aware that some contexts may incorrectly use (Total Profit After Investment - Investment Cost) / Investment Cost instead of the standard (Incremental Profit Gain / ...
\item {\ttfamily\bfseries\color{bulletblue}[err-00125]} {\scriptsize\color{countgray}h=0 r=0} Avoid automatically applying annuity formulas for multi-period cash flows without first verifying the number of payments intended. If the computed result (e.g., using ordinary or due annuity) ...
\item {\ttfamily\bfseries\color{bulletblue}[err-00160]} {\scriptsize\color{countgray}h=0 r=0} Avoid outputting ROI as a percentage when the instruction specifies a 'plain floating point number'. The decimal form (e.g., 2.857 for 285.7\% return) is required unless percentage output is ...
\item {\ttfamily\bfseries\color{bulletblue}[err-00170]} {\scriptsize\color{countgray}h=0 r=0} Avoid strictly adhering to 'end of each year' wording for annuity timing if the calculated PV (ordinary annuity) significantly diverges from the expected answer. In such cases, verify if the ...
\item {\ttfamily\bfseries\color{bulletblue}[err-00179]} {\scriptsize\color{countgray}h=0 r=0} Avoid using imprecise exponentiation methods for compound interest calculations, especially with small interest rates over many periods. Even slight rounding in the base (1 + r)\^{}n can lead to ...
\item {\ttfamily\bfseries\color{bulletblue}[err-00185]} {\scriptsize\color{countgray}h=0 r=0} Avoid overcomplicating straightforward rounding tasks by introducing extraneous financial context (e.g., percentage interpretation) when the instruction explicitly requests a plain floating point ...
\item {\ttfamily\bfseries\color{bulletblue}[err-00203]} {\scriptsize\color{countgray}h=0 r=0} Avoid altering numerically equivalent floating point outputs (e.g., changing 0.10 to 0.1) for financial calculations when the output format is specified as a 'plain floating point number'. ...
\item {\ttfamily\bfseries\color{bulletblue}[err-00207]} {\scriptsize\color{countgray}h=0 r=0} When using the Gordon Growth Model to calculate cost of equity (r = D1/P0 + g), use only the current stock price (P0) as the denominator. Do not use historical or previous stock prices, as they ...
\item {\ttfamily\bfseries\color{bulletblue}[err-00240]} {\scriptsize\color{countgray}h=0 r=0} Always remember the fundamental working capital formula: Working Capital = Current Assets - Current Liabilities. Ensure both values are in the same currency units before subtracting, and follow ...
\item {\ttfamily\bfseries\color{bulletblue}[err-00243]} {\scriptsize\color{countgray}h=0 r=0} For ROI calculations, avoid using revenue directly without subtracting the investment cost to compute net profit first. ROI = (Revenue - Cost) / Cost, not Revenue / Cost. This ensures the net gain ...
\item {\ttfamily\bfseries\color{bulletblue}[err-00249]} {\scriptsize\color{countgray}h=0 r=1} Avoid using P\_1\_domestic = P\_0\_foreign * (S\_1 / S\_0) for relative PPP problems with constant foreign price level. This incorrect formula assumes domestic prices change proportionally with the ...
\item {\ttfamily\bfseries\color{bulletblue}[err-00261]} {\scriptsize\color{countgray}h=0 r=0} For ratio change questions (e.g., P/B, P/E), carefully distinguish between calculating the new ratio value versus the change in the ratio. The question often asks for the new value after a ...
\end{itemize}
\pbsection{CONTEXT CLUES \& INDICATORS}
\begin{itemize}[leftmargin=1.2em, itemsep=1pt, parsep=0pt, topsep=2pt]
\item {\ttfamily\bfseries\color{bulletblue}[ctx-00069]} {\scriptsize\color{countgray}h=0 r=0} For unemployment rate problems: when the question states unemployed 'rises by' a specific number and the labor force is stagnant with no initial unemployed value provided, assume initial ...
\item {\ttfamily\bfseries\color{bulletblue}[ctx-00089]} {\scriptsize\color{countgray}h=0 r=0} For unemployment rate problems, a ground truth of 0.0 when the calculated employed is positive may indicate a misstated problem or trick question. Look for keywords like 'all' or 'entire' labor ...
\item {\ttfamily\bfseries\color{bulletblue}[ctx-00110]} {\scriptsize\color{countgray}h=0 r=0} For DSO calculations, carefully extract the period length from the problem context (e.g., '45 days' explicitly stated). If not specified, default to standard periods (e.g., 90 days for a quarter, ...
\item {\ttfamily\bfseries\color{bulletblue}[ctx-00126]} {\scriptsize\color{countgray}h=0 r=0} Phrases like 'due to receive' often indicate that a payment is immediate or very near-term, potentially implying only one cash flow should be considered for present value calculations, even if ...
\item {\ttfamily\bfseries\color{bulletblue}[ctx-00206]} {\scriptsize\color{countgray}h=0 r=0} For lottery or prize contexts, phrases like 'pay you \$X annually for Y years' may be misleading. If the computed present value using annuity formulas is significantly higher than the prize amount ...
\item {\ttfamily\bfseries\color{bulletblue}[ctx-00220]} {\scriptsize\color{countgray}h=0 r=0} When questions ask for 'impact on [metric]' due to a parameter change, carefully determine whether they require the resulting value or the change/difference. If the calculated change is small ...
\end{itemize}
\pbsection{OTHERS}
\begin{itemize}[leftmargin=1.2em, itemsep=1pt, parsep=0pt, topsep=2pt]
\item {\ttfamily\bfseries\color{bulletblue}[misc-00101]} {\scriptsize\color{countgray}h=0 r=0} When encountering phrases like 'decreased market cap' vs 'decrease in market cap', interpret carefully: 'decreased market cap' typically refers to the new market value after the decrease, while ...
\item {\ttfamily\bfseries\color{bulletblue}[misc-00124]} {\scriptsize\color{countgray}h=0 r=0} When both revenue and net profit are provided in a question asking for 'Net Profit from selling products' or similar phrasing, interpret this as a request for the net profit margin percentage (Net ...
\item {\ttfamily\bfseries\color{bulletblue}[misc-00151]} {\scriptsize\color{countgray}h=1 r=0} For financial ratios output as plain floating point numbers, recognize that trailing zeros after the decimal point do not change the numerical value (e.g., 0.2, 0.20, and 0.200 are all ...
\item {\ttfamily\bfseries\color{bulletblue}[misc-00173]} {\scriptsize\color{countgray}h=0 r=0} Investments in software development tools, technology upgrades, or intangible assets that enhance long-term operational capacity are typically classified as capital expenditures (CAPEX) and should ...
\item {\ttfamily\bfseries\color{bulletblue}[misc-00193]} {\scriptsize\color{countgray}h=0 r=0} For DSO calculations, the period length (number of days) must be inferred from context: 'half-year' or 'semi-annual' typically implies 180 days, 'quarter' implies 90 days, and 'annual' implies 365 ...
\item {\ttfamily\bfseries\color{bulletblue}[misc-00257]} {\scriptsize\color{countgray}h=0 r=0} For Free Cash Flow calculations, classify expenditures as CAPEX if they involve acquiring, upgrading, or maintaining long-term assets (e.g., machinery, buildings, software) that provide benefits ...
\end{itemize}
\end{playbookbox}
\vspace{8pt}
\begin{playbookbox}{ACE Final Playbook --- FiNER, Batch Size = 1}
\vspace{2pt}
\begin{center}
{\small Total context entries: \textbf{246} \quad|\quad Test accuracy: \textbf{76.0\%}}
\end{center}
\vspace{2pt}
\pbsection{STRATEGIES \& INSIGHTS}
\begin{itemize}[leftmargin=1.2em, itemsep=1pt, parsep=0pt, topsep=2pt]
\item {\ttfamily\bfseries\color{bulletblue}[sai-00006]} {\scriptsize\color{countgray}h=21 r=0} When identifying US GAAP tags for numerical entities, focus on the contextual meaning: credit facility amounts relate to borrowing capacity tags (e.g., ...
\item {\ttfamily\bfseries\color{bulletblue}[sai-00007]} {\scriptsize\color{countgray}h=1 r=0} When the same numerical value appears in multiple questions about the same financial facility, it typically represents the same fundamental metric. Phrases like 'may be less than \$X' indicate that ...
\item {\ttfamily\bfseries\color{bulletblue}[sai-00008]} {\scriptsize\color{countgray}h=3 r=4} For interest expense amounts specifically related to debt instruments, use InterestExpenseDebt. This tag is appropriate for both current period expense amounts and comparative period amounts when ...
\item {\ttfamily\bfseries\color{bulletblue}[sai-00009]} {\scriptsize\color{countgray}h=3 r=0} For concentration risk percentages, particularly those related to revenue from major customers, use ConcentrationRiskPercentage1. This tag applies to percentages that quantify the proportion of ...
\item {\ttfamily\bfseries\color{bulletblue}[sai-00010]} {\scriptsize\color{countgray}h=22 r=0} When identifying stated interest rates for debt instruments (expressed as percentages), use DebtInstrumentInterestRateStatedPercentage. This tag is appropriate for fixed rates or the stated ...
\item {\ttfamily\bfseries\color{bulletblue}[sai-00011]} {\scriptsize\color{countgray}h=11 r=0} For share-based compensation and common stock metrics, carefully distinguish between: par value per share (CommonStockParOrStatedValuePerShare), shares outstanding (CommonStockSharesOutstanding), ...
\item {\ttfamily\bfseries\color{bulletblue}[sai-00012]} {\scriptsize\color{countgray}h=14 r=3} For credit facility amounts, carefully distinguish between current and maximum borrowing capacity: 'borrowing base' refers to the LineOfCreditFacilityCurrentBorrowingCapacity (the currently ...
\item {\ttfamily\bfseries\color{bulletblue}[sai-00013]} {\scriptsize\color{countgray}h=19 r=7} For debt instruments, distinguish between principal/face amounts and fair value measurements: use DebtInstrumentFaceAmount for the original principal value (e.g., 'principal amount of notes') and ...
\item {\ttfamily\bfseries\color{bulletblue}[sai-00014]} {\scriptsize\color{countgray}h=1 r=1} For insurance company disclosures about 'development' in prior year claims reserves (adverse or favorable), use the specialized tag ...
\item {\ttfamily\bfseries\color{bulletblue}[sai-00015]} {\scriptsize\color{countgray}h=11 r=0} For share-based compensation events, precisely match the context to the GAAP tag: options exercised (intrinsic value) use ...
\item {\ttfamily\bfseries\color{bulletblue}[sai-00016]} {\scriptsize\color{countgray}h=3 r=0} For debt-related fair value measurements, carefully distinguish context: use LongTermDebtFairValue when the context refers to 'debt obligations' at a balance sheet date or specifically mentions ...
\item {\ttfamily\bfseries\color{bulletblue}[sai-00017]} {\scriptsize\color{countgray}h=1 r=2} For lease and rental expenses, use LeaseAndRentalExpense as the default tag when the context mentions 'rent expense' without specifying lease type (operating vs capital). Reserve ...
\item {\ttfamily\bfseries\color{bulletblue}[sai-00018]} {\scriptsize\color{countgray}h=5 r=0} For statutory tax rates mentioned in effective tax rate reconciliation disclosures, use EffectiveIncomeTaxRateReconciliationAtFederalStatutoryIncomeTaxRate regardless of whether it represents a ...
\item {\ttfamily\bfseries\color{bulletblue}[sai-00019]} {\scriptsize\color{countgray}h=8 r=0} The ConcentrationRiskPercentage1 tag applies broadly to various concentration risk disclosures, including revenue concentration (percentage of total revenue from major customers), accounts ...
\item {\ttfamily\bfseries\color{bulletblue}[sai-00020]} {\scriptsize\color{countgray}h=1 r=0} For US GAAP tag identification, first analyze the sentence context to determine the exact nature of the financial metric (expense type, liability characteristic, or measurement basis), then select ...
\item {\ttfamily\bfseries\color{bulletblue}[sai-00021]} {\scriptsize\color{countgray}h=15 r=1} For share-based compensation disclosures, carefully distinguish between recognized expense and other metrics: use AllocatedShareBasedCompensationExpense for compensation 'recognized' during a ...
\item {\ttfamily\bfseries\color{bulletblue}[sai-00022]} {\scriptsize\color{countgray}h=1 r=0} For treasury stock transactions, carefully distinguish between per-share and aggregate measurements: TreasuryStockAcquiredAverageCostPerShare for average price per share of repurchased shares, and ...
\item {\ttfamily\bfseries\color{bulletblue}[sai-00023]} {\scriptsize\color{countgray}h=0 r=0} For revolving credit facilities described at inception (e.g., 'entered into a \$X million facility'), the stated amount represents the current borrowing capacity available at that time before any ...
\item {\ttfamily\bfseries\color{bulletblue}[sai-00024]} {\scriptsize\color{countgray}h=6 r=2} For letters of credit under credit facilities, use LettersOfCreditOutstandingAmount for outstanding balances, regardless of currency denomination or sub-limit context. This tag applies to the ...
\item {\ttfamily\bfseries\color{bulletblue}[sai-00025]} {\scriptsize\color{countgray}h=9 r=0} For share-based compensation disclosures, precisely distinguish between the quantity of awards granted (use GrantsInPeriod tags like ...
\item {\ttfamily\bfseries\color{bulletblue}[sai-00026]} {\scriptsize\color{countgray}h=1 r=0} When classifying debt-related amounts, distinguish between instrument characteristics and balance sheet presentation: use DebtInstrumentFaceAmount for the original principal value of specific debt ...
\item {\ttfamily\bfseries\color{bulletblue}[sai-00027]} {\scriptsize\color{countgray}h=0 r=0} When encountering incomplete financial context with multiple unlabeled amounts (e.g., 'Includes of \$X million, and of \$Y million'), analyze the relative magnitudes of the numbers. Large amounts ...
\item {\ttfamily\bfseries\color{bulletblue}[sai-00028]} {\scriptsize\color{countgray}h=2 r=0} For tax-related amounts, distinguish between current period income tax expenses (use IncomeTaxExpenseBenefit) and uncertain tax positions (use UnrecognizedTaxBenefits for gross amounts and ...
\item {\ttfamily\bfseries\color{bulletblue}[sai-00029]} {\scriptsize\color{countgray}h=1 r=0} For utility companies filing rate change applications with regulatory commissions, use PublicUtilitiesRequestedRateIncreaseDecreaseAmount for the dollar amount of the requested change. This tag ...
\item {\ttfamily\bfseries\color{bulletblue}[sai-00030]} {\scriptsize\color{countgray}h=0 r=0} For lease accounting under ASC 842, when a numerical range is provided for 'lease assets and lease liabilities' together, carefully analyze the context to determine which component is being ...
\item {\ttfamily\bfseries\color{bulletblue}[sai-00031]} {\scriptsize\color{countgray}h=7 r=0} For share-based compensation plan limits, precisely distinguish between three key metrics: 1) Total authorized shares for the plan ...
\item {\ttfamily\bfseries\color{bulletblue}[sai-00032]} {\scriptsize\color{countgray}h=0 r=0} For XBRL tag selection, pay meticulous attention to capitalization patterns as they are case-sensitive and must exactly match the standard taxonomy. Common patterns include camelCase for ...
\item {\ttfamily\bfseries\color{bulletblue}[sai-00033]} {\scriptsize\color{countgray}h=0 r=0} Unit appreciation rights (UARs) and stock appreciation rights (SARs) are classified as 'equity instruments other than options' under US GAAP, not as options. Use ...
\item {\ttfamily\bfseries\color{bulletblue}[sai-00034]} {\scriptsize\color{countgray}h=1 r=4} For equity share reservations, distinguish between share-based compensation authorizations and general future issuances: Use ...
\item {\ttfamily\bfseries\color{bulletblue}[sai-00035]} {\scriptsize\color{countgray}h=12 r=0} For stock option disclosures, carefully distinguish between vesting period (time until options become exercisable) and expiration period (total time options remain valid). Use ...
\item {\ttfamily\bfseries\color{bulletblue}[sai-00036]} {\scriptsize\color{countgray}h=2 r=0} For amortization of intangible assets (e.g., patents, trademarks), use AmortizationOfIntangibleAssets. For amortization of deferred contract costs under ASC 340-40 (e.g., capitalized incremental ...
\item {\ttfamily\bfseries\color{bulletblue}[sai-00037]} {\scriptsize\color{countgray}h=0 r=1} For counts of primary business lines or segments that meet the definition of operating segments under ASC 280, use NumberOfOperatingSegments. This tag applies to the quantitative disclosure of how ...
\item {\ttfamily\bfseries\color{bulletblue}[sai-00038]} {\scriptsize\color{countgray}h=1 r=0} When selecting US GAAP tags, systematically evaluate three dimensions: 1) Financial instrument type (debt, equity, derivative, etc.), 2) Measurement context (face amount, fair value, outstanding ...
\item {\ttfamily\bfseries\color{bulletblue}[sai-00039]} {\scriptsize\color{countgray}h=1 r=0} For debt-related payments, carefully distinguish between transaction types: use RepaymentsOfDebt for installment payments or scheduled reductions of debt principal, and ...
\item {\ttfamily\bfseries\color{bulletblue}[sai-00040]} {\scriptsize\color{countgray}h=4 r=0} For intangible assets, distinguish between amortization expense and useful life: use AmortizationOfIntangibleAssets for the periodic expense amount, and FiniteLivedIntangibleAssetUsefulLife for ...
\item {\ttfamily\bfseries\color{bulletblue}[sai-00041]} {\scriptsize\color{countgray}h=3 r=1} For property, plant and equipment (PP\&E), distinguish between depreciation expense and useful life: use Depreciation for the periodic expense amount, and PropertyPlantAndEquipmentUsefulLife for ...
\item {\ttfamily\bfseries\color{bulletblue}[sai-00042]} {\scriptsize\color{countgray}h=0 r=4} For segment reporting under ASC 280, carefully distinguish between operating segments and reportable segments: use NumberOfOperatingSegments for counts of segments identified by management before ...
\item {\ttfamily\bfseries\color{bulletblue}[sai-00043]} {\scriptsize\color{countgray}h=0 r=0} For Dividend Reinvestment Plans (DRIPs), use ShareBasedCompensationArrangementByShareBasedPaymentAwardNumberOfSharesAuthorized for shares authorized for issuance under such plans. DRIPs are ...
\item {\ttfamily\bfseries\color{bulletblue}[sai-00044]} {\scriptsize\color{countgray}h=5 r=1} For effective tax rate disclosures, use EffectiveIncomeTaxRateContinuingOperations when the context mentions 'effective tax rate' or 'effective income tax rate' for continuing operations. This tag ...
\item {\ttfamily\bfseries\color{bulletblue}[sai-00045]} {\scriptsize\color{countgray}h=6 r=0} For antidilutive securities excluded from EPS calculations, use AntidilutiveSecuritiesExcludedFromComputationOfEarningsPerShareAmount when the context specifies securities 'excluded from ...
\item {\ttfamily\bfseries\color{bulletblue}[sai-00046]} {\scriptsize\color{countgray}h=8 r=0} When distinguishing between common and preferred stock metrics, carefully analyze the context to identify the specific stock type: CommonStockParOrStatedValuePerShare for common stock par value, ...
\item {\ttfamily\bfseries\color{bulletblue}[sai-00047]} {\scriptsize\color{countgray}h=0 r=5} For share-based compensation, carefully distinguish between total recognized expense and allocated amounts: use ShareBasedCompensation for the total compensation expense recognized during a period ...
\item {\ttfamily\bfseries\color{bulletblue}[sai-00049]} {\scriptsize\color{countgray}h=0 r=0} When analyzing credit facility structures with multiple tranches, carefully distinguish between overall facility borrowing capacity (LineOfCreditFacilityMaximumBorrowingCapacity) and specific ...
\item {\ttfamily\bfseries\color{bulletblue}[sai-00050]} {\scriptsize\color{countgray}h=0 r=0} For lease-related charges, carefully distinguish between ongoing operational expenses and restructuring activities: use LeaseAndRentalExpense for normal recurring lease/rent expenses, but use ...
\item {\ttfamily\bfseries\color{bulletblue}[sai-00051]} {\scriptsize\color{countgray}h=0 r=0} For share issuance disclosures indicating no new issuances (e.g., 'no common units issued'), use StockIssuedDuringPeriodSharesNewIssues with a value of zero. This tag applies to both positive ...
\item {\ttfamily\bfseries\color{bulletblue}[sai-00052]} {\scriptsize\color{countgray}h=0 r=0} For employer contributions to defined benefit pension plans, use DefinedBenefitPlanContributionsByEmployer. This tag applies to cash funding amounts made by the employer to pension plans, distinct ...
\item {\ttfamily\bfseries\color{bulletblue}[sai-00053]} {\scriptsize\color{countgray}h=0 r=0} When tagging numerical values in financial contexts, prioritize the economic substance over surface features. For debt instruments, treat units (e.g., 'years') as integral to the measurement ...
\item {\ttfamily\bfseries\color{bulletblue}[sai-00054]} {\scriptsize\color{countgray}h=0 r=3} For operating lease rent expenses, always prefer OperatingLeasesRentExpenseNet over the more general OperatingLeaseExpense when the context specifically mentions 'rent expense for operating ...
\item {\ttfamily\bfseries\color{bulletblue}[sai-00056]} {\scriptsize\color{countgray}h=1 r=0} For unrecognized tax benefits, carefully analyze contextual phrases that indicate impact on the effective tax rate. Phrases like 'will reduce our effective tax rate', 'would impact the effective ...
\item {\ttfamily\bfseries\color{bulletblue}[sai-00060]} {\scriptsize\color{countgray}h=0 r=1} When derivative instruments (e.g., bond hedges, warrants) have terms explicitly defined by reference to a primary financial instrument's characteristic (e.g., conversion price of convertible ...
\item {\ttfamily\bfseries\color{bulletblue}[sai-00062]} {\scriptsize\color{countgray}h=0 r=0} When encountering benchmark rates (e.g., LIBOR, SOFR) in debt instrument contexts, carefully analyze whether they are presented as standalone reference rates or as components of basis spread ...
\item {\ttfamily\bfseries\color{bulletblue}[sai-00063]} {\scriptsize\color{countgray}h=2 r=0} For stock issuances during a period, default to 'StockIssuedDuringPeriodSharesNewIssues' for both common and preferred shares unless specific context requires more granular tags. Reserve ...
\item {\ttfamily\bfseries\color{bulletblue}[sai-00064]} {\scriptsize\color{countgray}h=0 r=0} For credit facility disclosures, carefully distinguish between actual debt drawn and available capacity: 'borrowings under [facility]' indicates actual debt incurred (use LongTermDebt or ...
\item {\ttfamily\bfseries\color{bulletblue}[sai-00065]} {\scriptsize\color{countgray}h=3 r=0} For share-based compensation, distinguish between the quantity of awards granted (ShareBasedCompensationArrangementByShareBasedPaymentAwardEquityInstrumentsOtherThanOptionsGrantsInPeriod) and ...
\item {\ttfamily\bfseries\color{bulletblue}[sai-00066]} {\scriptsize\color{countgray}h=0 r=0} When analyzing business transaction consideration, carefully determine the directional context: if the company is paying consideration (e.g., 'we paid', 'purchase price'), use acquisition tags ...
\item {\ttfamily\bfseries\color{bulletblue}[sai-00067]} {\scriptsize\color{countgray}h=1 r=0} When identifying transaction types, carefully analyze contextual clues for related party relationships. Phrases like 'with related parties', 'transactions with affiliates', or specific entity ...
\item {\ttfamily\bfseries\color{bulletblue}[sai-00068]} {\scriptsize\color{countgray}h=0 r=1} For debt instruments appearing in balance sheet contexts (e.g., 'notes payable', 'outstanding debt'), prefer LongTermDebt or ShortTermDebt over DebtInstrumentFaceAmount. Use ...
\item {\ttfamily\bfseries\color{bulletblue}[sai-00069]} {\scriptsize\color{countgray}h=0 r=0} When both 'operating' and 'reportable' are used together to describe segments (e.g., 'operating and reportable segments'), prefer NumberOfOperatingSegments as it captures the fundamental segments ...
\item {\ttfamily\bfseries\color{bulletblue}[sai-00070]} {\scriptsize\color{countgray}h=3 r=1} For loss contingency disclosures, carefully distinguish between recognized amounts and estimated ranges: use LossContingencyAccrualAtCarryingValue for liabilities already recognized on the balance ...
\item {\ttfamily\bfseries\color{bulletblue}[sai-00071]} {\scriptsize\color{countgray}h=0 r=2} For business acquisition disclosures involving ownership percentages, use BusinessAcquisitionPercentageOfVotingInterestsAcquired when the context describes the percentage of voting interests ...
\item {\ttfamily\bfseries\color{bulletblue}[sai-00072]} {\scriptsize\color{countgray}h=0 r=0} For amortization expenses specifically related to intangible assets from lease acquisitions (e.g., in-place leases), use 'AmortizationOfIntangibleAssets'. This tag applies to lease-related ...
\item {\ttfamily\bfseries\color{bulletblue}[sai-00073]} {\scriptsize\color{countgray}h=1 r=0} For weighted average interest rates on debt obligations, use 'DebtWeightedAverageInterestRate'. This tag specifically applies to the calculated average rate across multiple debt instruments, ...
\item {\ttfamily\bfseries\color{bulletblue}[sai-00074]} {\scriptsize\color{countgray}h=0 r=0} For share-based compensation expenses, carefully analyze contextual modifiers that indicate allocation or adjustment: use AllocatedShareBasedCompensationExpense when phrases like 'net of ...
\item {\ttfamily\bfseries\color{bulletblue}[sai-00075]} {\scriptsize\color{countgray}h=1 r=0} For revenue transactions involving related parties (e.g., equity method investees, affiliates), prioritize the transaction nature over the relationship when selecting US GAAP tags. Use Revenues ...
\item {\ttfamily\bfseries\color{bulletblue}[sai-00076]} {\scriptsize\color{countgray}h=1 r=0} When tagging measurement units (e.g., 'year', 'month') in share-based compensation contexts, treat them as integral to the measurement they describe rather than separate entities. For example, ...
\item {\ttfamily\bfseries\color{bulletblue}[sai-00077]} {\scriptsize\color{countgray}h=1 r=0} In share-based compensation disclosures, the phrase 'from the grant date' specifically indicates expiration period (total term of the award) rather than vesting period. Use ...
\item {\ttfamily\bfseries\color{bulletblue}[sai-00078]} {\scriptsize\color{countgray}h=0 r=1} When encountering fair value hierarchy disclosures (Level 1, 2, or 3) for debt instruments, carefully distinguish between the measurement basis disclosed (fair value) and the actual balance sheet ...
\item {\ttfamily\bfseries\color{bulletblue}[sai-00081]} {\scriptsize\color{countgray}h=2 r=0} For warrant exercise price disclosures, use ClassOfWarrantOrRightExercisePriceOfWarrantsOrRights1 when the context describes the price at which warrants can be exercised to purchase underlying ...
\item {\ttfamily\bfseries\color{bulletblue}[sai-00082]} {\scriptsize\color{countgray}h=0 r=0} When identifying useful life measurements for property, plant and equipment, use PropertyPlantAndEquipmentUsefulLife for the estimated service life period (e.g., '20 years' for buildings). This ...
\item {\ttfamily\bfseries\color{bulletblue}[sai-00083]} {\scriptsize\color{countgray}h=0 r=0} For revenue amounts specifically described in context (e.g., 'reimbursable revenues', 'contract revenue', 'customer revenue'), prefer the specific tag ...
\item {\ttfamily\bfseries\color{bulletblue}[sai-00086]} {\scriptsize\color{countgray}h=1 r=0} For restructuring costs, carefully distinguish temporal context: use RestructuringCharges for expenses that have been recognized and incurred (e.g., 'expense totaled', 'charges recognized'), but ...
\item {\ttfamily\bfseries\color{bulletblue}[sai-00088]} {\scriptsize\color{countgray}h=1 r=2} For debt facilities, critically distinguish between term loans and revolving credit lines: term loans represent fixed principal amounts requiring DebtInstrumentCarryingAmount (e.g., '\$20.0 million ...
\item {\ttfamily\bfseries\color{bulletblue}[sai-00089]} {\scriptsize\color{countgray}h=2 r=0} When analyzing numerical values, carefully distinguish between percentage-based metrics and dollar amount metrics. Percentage values (e.g., '0.5\% commitment fee') require percentage tags like ...
\item {\ttfamily\bfseries\color{bulletblue}[sai-00090]} {\scriptsize\color{countgray}h=0 r=0} For insurance risk concentration percentages (e.g., ceded risk percentages, maximum exposure percentages, risk allocation percentages), use ConcentrationRiskPercentage1. This tag applies to ...
\item {\ttfamily\bfseries\color{bulletblue}[sai-00091]} {\scriptsize\color{countgray}h=3 r=0} For stock repurchase program disclosures, when both the authorized amount and actual repurchased amount reference the same authorization program, use StockRepurchaseProgramAuthorizedAmount1 for ...
\item {\ttfamily\bfseries\color{bulletblue}[sai-00092]} {\scriptsize\color{countgray}h=6 r=0} For share-based compensation disclosures, carefully distinguish between different award phases: use grants tags (e.g., ...
\item {\ttfamily\bfseries\color{bulletblue}[sai-00093]} {\scriptsize\color{countgray}h=0 r=0} For lease and rental contexts, critically distinguish between cash payment language ('paid in rent', 'rental payments') and expense recognition language ('rent expense', 'lease expense'). Use ...
\item {\ttfamily\bfseries\color{bulletblue}[sai-00095]} {\scriptsize\color{countgray}h=0 r=0} For business acquisition disclosures, carefully distinguish between cash flow perspective and accounting perspective: use PaymentsToAcquireBusinessesNetOfCashAcquired for net cash outflow amounts ...
\item {\ttfamily\bfseries\color{bulletblue}[sai-00096]} {\scriptsize\color{countgray}h=0 r=0} For segment reporting under ASC 280, use NumberOfOperatingSegments when the context describes operational structure or management approach (e.g., 'conduct operations in/through segments', 'managed ...
\item {\ttfamily\bfseries\color{bulletblue}[sai-00097]} {\scriptsize\color{countgray}h=6 r=1} For fair value measurements of debt instruments, always analyze the maturity date to determine debt classification: use LongTermDebtFairValue for obligations due beyond one year (e.g., 'Senior ...
\item {\ttfamily\bfseries\color{bulletblue}[sai-00098]} {\scriptsize\color{countgray}h=0 r=0} For rent expense contexts that mention contingent rentals or multiple lease components, prefer LeaseAndRentalExpense over OperatingLeasesRentExpenseNet. The broader tag better captures ...
\item {\ttfamily\bfseries\color{bulletblue}[sai-00099]} {\scriptsize\color{countgray}h=0 r=0} For business acquisition payments, carefully distinguish between cash consideration and total consideration: use PaymentsToAcquireBusinessesGross for specific cash payments made to acquire ...
\item {\ttfamily\bfseries\color{bulletblue}[sai-00100]} {\scriptsize\color{countgray}h=1 r=2} For stock repurchase transactions, carefully distinguish between different measurement contexts: use StockRepurchasedDuringPeriodShares for total shares repurchased (e.g., 'repurchased X shares'), ...
\item {\ttfamily\bfseries\color{bulletblue}[sai-00101]} {\scriptsize\color{countgray}h=0 r=0} For XBRL tagging of time-based measurements in share-based compensation (e.g., vesting periods, expiration periods), treat the measurement unit (e.g., 'years', 'months') as an integral part of the ...
\item {\ttfamily\bfseries\color{bulletblue}[sai-00102]} {\scriptsize\color{countgray}h=1 r=1} When analyzing debt-related contexts, carefully distinguish between measurement tags that describe instrument characteristics (e.g., DebtInstrumentFaceAmount for original principal value, ...
\item {\ttfamily\bfseries\color{bulletblue}[sai-00104]} {\scriptsize\color{countgray}h=2 r=0} When analyzing ownership percentages, carefully distinguish between acquisition events and resulting ownership structures: use BusinessAcquisitionPercentageOfVotingInterestsAcquired for the ...
\item {\ttfamily\bfseries\color{bulletblue}[sai-00106]} {\scriptsize\color{countgray}h=0 r=1} When multiple tags could potentially apply to a financial context, always prefer the most specific tag that precisely matches the economic substance of the transaction or measurement. For example, ...
\item {\ttfamily\bfseries\color{bulletblue}[sai-00108]} {\scriptsize\color{countgray}h=0 r=0} For operating lease rent expenses, use OperatingLeasesRentExpenseNet when the context explicitly mentions 'rent expense for operating leases' or similar phrasing that specifically ties the rent ...
\item {\ttfamily\bfseries\color{bulletblue}[sai-00109]} {\scriptsize\color{countgray}h=0 r=0} For letters of credit under credit facilities, use LettersOfCreditOutstandingAmount when the context describes the actual drawn or outstanding balance, including initial amounts at issuance. ...
\item {\ttfamily\bfseries\color{bulletblue}[sai-00110]} {\scriptsize\color{countgray}h=2 r=0} For credit facility borrowing capacity, carefully distinguish three concepts: 1) Maximum authorized limit (LineOfCreditFacilityMaximumBorrowingCapacity), 2) Current capacity based on collateral ...
\item {\ttfamily\bfseries\color{bulletblue}[sai-00111]} {\scriptsize\color{countgray}h=1 r=4} For commitment fees on credit facilities, use LineOfCreditFacilityUnusedCapacityCommitmentFeePercentage when the context specifically mentions fees 'on undrawn amounts' or 'on unused capacity'. ...
\item {\ttfamily\bfseries\color{bulletblue}[sai-00112]} {\scriptsize\color{countgray}h=8 r=1} For business acquisition disclosures, carefully distinguish between total consideration transferred (BusinessCombinationConsiderationTransferred1) and net cash payments ...
\item {\ttfamily\bfseries\color{bulletblue}[sai-00113]} {\scriptsize\color{countgray}h=0 r=1} For stock repurchase disclosures, carefully distinguish between shares merely repurchased (use StockRepurchasedDuringPeriodShares) and shares both repurchased and retired (use ...
\item {\ttfamily\bfseries\color{bulletblue}[sai-00114]} {\scriptsize\color{countgray}h=0 r=0} For the weighted-average grant date fair value of stock options granted during a period, use ...
\item {\ttfamily\bfseries\color{bulletblue}[sai-00115]} {\scriptsize\color{countgray}h=0 r=0} For remaining performance obligation disclosures under ASC 606, use RevenueRemainingPerformanceObligation for both the total amount and any segment-specific portions. The same tag applies to the ...
\item {\ttfamily\bfseries\color{bulletblue}[sai-00116]} {\scriptsize\color{countgray}h=1 r=0} When the context specifies 'net of cash acquired' in business acquisition disclosures, use PaymentsToAcquireBusinessesNetOfCashAcquired. This tag specifically applies to the net cash outflow ...
\item {\ttfamily\bfseries\color{bulletblue}[sai-00117]} {\scriptsize\color{countgray}h=4 r=3} For debt instruments, carefully distinguish between 'outstanding' amounts and 'principal' amounts: 'outstanding' refers to the current carrying amount on the balance sheet (use ...
\item {\ttfamily\bfseries\color{bulletblue}[sai-00118]} {\scriptsize\color{countgray}h=0 r=0} When 'principal amount' is mentioned in credit facility contexts (e.g., 'aggregate principal amount of up to €65 million'), it refers to the face value of debt instruments that can be drawn under ...
\item {\ttfamily\bfseries\color{bulletblue}[sai-00119]} {\scriptsize\color{countgray}h=4 r=1} For depreciation expenses specifically related to property, plant and equipment (e.g., buildings, machinery, equipment), use the 'Depreciation' tag. This applies to both current period expense ...
\item {\ttfamily\bfseries\color{bulletblue}[sai-00120]} {\scriptsize\color{countgray}h=2 r=0} For goodwill amounts that represent the asset value recorded during acquisition transactions (not impairment charges), use the 'Goodwill' tag. This applies when the context describes goodwill ...
\item {\ttfamily\bfseries\color{bulletblue}[sai-00121]} {\scriptsize\color{countgray}h=3 r=0} For interest rate spreads on variable rate debt instruments, use DebtInstrumentBasisSpreadOnVariableRate1 regardless of the benchmark rate (e.g., LIBOR, prime, SOFR) specified in the context. The ...
\item {\ttfamily\bfseries\color{bulletblue}[sai-00122]} {\scriptsize\color{countgray}h=0 r=2} When tagging ownership percentages, carefully analyze legal structure and control indicators beyond just the percentage value. For limited liability companies (LLCs), joint ventures, or entities ...
\item {\ttfamily\bfseries\color{bulletblue}[sai-00123]} {\scriptsize\color{countgray}h=1 r=0} In fair value hierarchy disclosures, the dollar amounts presented represent fair value measurements even when the context mentions that fair value approximates carrying amount. The fair value ...
\item {\ttfamily\bfseries\color{bulletblue}[sai-00124]} {\scriptsize\color{countgray}h=0 r=0} When share-based compensation expense is explicitly broken down by specific award types (e.g., options, restricted stock awards, restricted stock units) with separate dollar amounts for each ...
\item {\ttfamily\bfseries\color{bulletblue}[sai-00125]} {\scriptsize\color{countgray}h=1 r=0} For share-based compensation fair value measurements, critically distinguish between total aggregate amounts and weighted average amounts: use 'TotalFairValue' tags (e.g., ...
\item {\ttfamily\bfseries\color{bulletblue}[sai-00126]} {\scriptsize\color{countgray}h=0 r=0} For equity interests issued in business acquisitions, use BusinessAcquisitionEquityInterestsIssuedOrIssuableNumberOfSharesIssued regardless of whether the context describes shares being issued or ...
\item {\ttfamily\bfseries\color{bulletblue}[sai-00127]} {\scriptsize\color{countgray}h=0 r=0} When the entity receives equity investments from third parties (rather than issuing its own stock), use EquityMethodInvestments instead of ProceedsFromIssuanceOfCommonStock. The key distinction is ...
\item {\ttfamily\bfseries\color{bulletblue}[sai-00128]} {\scriptsize\color{countgray}h=0 r=0} When a percentage appears in the name of a debt instrument (e.g., '6.75 \% Notes'), it typically refers to the stated interest rate that characterizes the instrument, not the face amount. Use ...
\item {\ttfamily\bfseries\color{bulletblue}[sai-00129]} {\scriptsize\color{countgray}h=3 r=2} For share-based compensation, precisely match the specific type of equity instrument mentioned in the context to the most specific tag available. Restricted share units (RSUs) and other equity ...
\item {\ttfamily\bfseries\color{bulletblue}[sai-00131]} {\scriptsize\color{countgray}h=1 r=1} For specific US GAAP tag applications: use BusinessCombinationAcquisitionRelatedCosts for acquisition-related costs (e.g., legal, accounting, advisory fees), ...
\item {\ttfamily\bfseries\color{bulletblue}[sai-00132]} {\scriptsize\color{countgray}h=0 r=0} For warrant quantity disclosures (e.g., '4.6 million warrants issued'), use StockIssuedDuringPeriodSharesNewIssues as warrants are equity instruments representing potential future shares issued ...
\item {\ttfamily\bfseries\color{bulletblue}[sai-00134]} {\scriptsize\color{countgray}h=0 r=0} For share-based compensation fair value measurements, critically distinguish temporal context: use grant-date weighted average fair value tags (e.g., ...
\item {\ttfamily\bfseries\color{bulletblue}[sai-00135]} {\scriptsize\color{countgray}h=0 r=0} For business acquisition disclosures, carefully distinguish between cash consideration and total consideration: use PaymentsToAcquireBusinessesGross for specific cash payments made to acquire ...
\item {\ttfamily\bfseries\color{bulletblue}[sai-00136]} {\scriptsize\color{countgray}h=0 r=0} For stock repurchase transactions, distinguish between authorization context ('up to \$X') which requires StockRepurchaseProgramAuthorizedAmount1 and acquisition context ('received X shares') which ...
\item {\ttfamily\bfseries\color{bulletblue}[sai-00137]} {\scriptsize\color{countgray}h=2 r=0} For loss contingency disclosures, always check if the context specifies a particular type of contingency (environmental, litigation, etc.) and prefer the most specific tag available. When ...
\item {\ttfamily\bfseries\color{bulletblue}[sai-00138]} {\scriptsize\color{countgray}h=0 r=0} For share-based compensation metrics, systematically match the specific measurement context (unrecognized compensation cost, weighted-average recognition period, grant date fair value per share, ...
\item {\ttfamily\bfseries\color{bulletblue}[sai-00139]} {\scriptsize\color{countgray}h=0 r=0} For warrants and similar instruments, carefully analyze whether periods until exercisability (e.g., 'exercisable after X months') should be tagged as part of the expiration period rather than ...
\item {\ttfamily\bfseries\color{bulletblue}[sai-00140]} {\scriptsize\color{countgray}h=0 r=0} For share-based compensation disclosures, critically distinguish between quantitative measurements (count/number of shares) and monetary values (fair value amounts). Pay close attention to ...
\item {\ttfamily\bfseries\color{bulletblue}[sai-00141]} {\scriptsize\color{countgray}h=0 r=0} For counts of operational groupings used in internal reporting, such as grouping tenants by activity segments or other management-defined categories, use NumberOfOperatingSegments. This tag ...
\item {\ttfamily\bfseries\color{bulletblue}[sai-00142]} {\scriptsize\color{countgray}h=0 r=0} When share-based compensation expense is recognized and allocated to specific income statement line items (e.g., cost of sales, selling, general and administrative expenses), use ...
\item {\ttfamily\bfseries\color{bulletblue}[sai-00143]} {\scriptsize\color{countgray}h=2 r=0} For debt issuance costs, carefully distinguish between initial capitalization and subsequent amortization: use DeferredFinanceCostsGross for the initial payment/capitalization of financing costs ...
\item {\ttfamily\bfseries\color{bulletblue}[sai-00144]} {\scriptsize\color{countgray}h=0 r=0} For equity share reservations, carefully distinguish between shares reserved for future issuance and proceeds from actual issuances: use CommonStockCapitalSharesReservedForFutureIssuance for ...
\item {\ttfamily\bfseries\color{bulletblue}[sai-00145]} {\scriptsize\color{countgray}h=0 r=0} For lease expense tagging, follow a specificity hierarchy: use the general LeaseAndRentalExpense tag when the context mentions 'rent expense' without specifying lease type (operating vs capital). ...
\item {\ttfamily\bfseries\color{bulletblue}[sai-00146]} {\scriptsize\color{countgray}h=0 r=0} When model predictions exactly match ground truth answers and environment feedback confirms no errors occurred, continue applying the same successful approach: maintain contextual analysis, ...
\item {\ttfamily\bfseries\color{bulletblue}[sai-00147]} {\scriptsize\color{countgray}h=1 r=0} For percentage values describing vesting conditions of share-based awards (e.g., '40\% of the restricted share units vest based on market conditions'), use ...
\item {\ttfamily\bfseries\color{bulletblue}[sai-00148]} {\scriptsize\color{countgray}h=2 r=0} For weighted average grant date fair value per unit measurements of equity instruments other than options (e.g., '\$19.28 per restricted share unit'), use ...
\item {\ttfamily\bfseries\color{bulletblue}[sai-00149]} {\scriptsize\color{countgray}h=0 r=0} For debt instruments issued at a discount or premium, the 'aggregate principal amount' mentioned in issuance contexts may be tagged as DebtInstrumentCarryingAmount when it refers to the amount ...
\item {\ttfamily\bfseries\color{bulletblue}[sai-00150]} {\scriptsize\color{countgray}h=0 r=0} When interpreting share-based compensation plan descriptions, carefully analyze modifying words that indicate maximum capacity: phrases like 'ceiling of X shares available for issuance,' 'maximum ...
\item {\ttfamily\bfseries\color{bulletblue}[sai-00151]} {\scriptsize\color{countgray}h=0 r=0} For operating lease rent expenses, prefer 'LeaseAndRentalExpense' when the context mentions 'rent expense' without indicating sublease income or netting. Reserve 'OperatingLeasesRentExpenseNet' ...
\item {\ttfamily\bfseries\color{bulletblue}[sai-00152]} {\scriptsize\color{countgray}h=0 r=0} For business acquisition disclosures, carefully distinguish between total consideration transferred (BusinessCombinationConsiderationTransferred1) and specific cash payments ...
\item {\ttfamily\bfseries\color{bulletblue}[sai-00153]} {\scriptsize\color{countgray}h=0 r=0} For debt-related transactions, carefully distinguish between instrument characteristics and repayment actions: use DebtInstrumentFaceAmount for describing the original principal value ...
\item {\ttfamily\bfseries\color{bulletblue}[sai-00154]} {\scriptsize\color{countgray}h=0 r=0} For cumulative effects of adopting new accounting principles (e.g., 'cumulative effect of adopting ASC 842'), use CumulativeEffectOfNewAccountingPrincipleInPeriodOfAdoption. This tag applies to ...
\item {\ttfamily\bfseries\color{bulletblue}[sai-00155]} {\scriptsize\color{countgray}h=3 r=1} For debt instrument measurements on the balance sheet, prioritize DebtInstrumentCarryingAmount over LongTermDebt when the context describes the carrying value of specific debt instruments (e.g., ...
\item {\ttfamily\bfseries\color{bulletblue}[sai-00156]} {\scriptsize\color{countgray}h=10 r=1} For revolving credit facilities, always use LineOfCreditFacilityMaximumBorrowingCapacity for the maximum authorized borrowing limit. Never use DebtInstrumentFaceAmount for revolving facilities, as ...
\item {\ttfamily\bfseries\color{bulletblue}[sai-00157]} {\scriptsize\color{countgray}h=0 r=0} For loss contingency disclosures, always check if more specific tags exist for particular contingency types (environmental, litigation damages, etc.) before defaulting to general tags. The phrase ...
\item {\ttfamily\bfseries\color{bulletblue}[sai-00158]} {\scriptsize\color{countgray}h=0 r=0} When shares are issued in business acquisitions or other transactions, prioritize the stock type context (common vs. preferred) over the transaction context for tag selection. Use ...
\item {\ttfamily\bfseries\color{bulletblue}[sai-00159]} {\scriptsize\color{countgray}h=0 r=0} When derivative instruments (e.g., warrants, hedges) have terms explicitly defined by reference to a primary financial instrument's characteristic, carefully analyze whether the numerical value ...
\item {\ttfamily\bfseries\color{bulletblue}[sai-00160]} {\scriptsize\color{countgray}h=0 r=0} For credit facilities with sub-facilities (e.g., letter of credit sub-facilities, swingline sub-facilities), carefully distinguish between the maximum authorized capacity of the sub-facility and ...
\item {\ttfamily\bfseries\color{bulletblue}[sai-00161]} {\scriptsize\color{countgray}h=0 r=0} For fair value measurements of debt instruments, default to DebtInstrumentFairValue when maturity context is absent or unspecified. Only use LongTermDebtFairValue when the debt is explicitly ...
\item {\ttfamily\bfseries\color{bulletblue}[sai-00162]} {\scriptsize\color{countgray}h=0 r=0} For loss contingency disclosures, critically distinguish between accrued amounts (already recorded on balance sheet) and estimates of possible losses (potential future exposures). Use accrual tags ...
\item {\ttfamily\bfseries\color{bulletblue}[sai-00163]} {\scriptsize\color{countgray}h=0 r=0} For stock repurchase transactions, carefully distinguish between shares repurchased and held as treasury stock versus shares repurchased and retired: use TreasuryStockSharesAcquired when shares ...
\item {\ttfamily\bfseries\color{bulletblue}[sai-00165]} {\scriptsize\color{countgray}h=0 r=0} For ownership percentages in collaborative arrangements, critically distinguish between consortiums and joint ventures: consortiums may involve subsidiary structures with minority interests ...
\item {\ttfamily\bfseries\color{bulletblue}[sai-00167]} {\scriptsize\color{countgray}h=0 r=0} When shares are repurchased under a formal stock repurchase program, they are typically retired as part of that program unless explicitly stated otherwise. Use ...
\item {\ttfamily\bfseries\color{bulletblue}[sai-00169]} {\scriptsize\color{countgray}h=0 r=0} For tax benefits derived from operating loss carryforwards that reduce income tax expense, use OperatingLossCarryforwards. This tag applies when the context describes tax benefits from net ...
\item {\ttfamily\bfseries\color{bulletblue}[sai-00170]} {\scriptsize\color{countgray}h=0 r=0} Maintain consistent tagging for the same financial metric across different periods, including when values are zero ('none' or '0'). The same US GAAP tag should be applied to represent the ...
\item {\ttfamily\bfseries\color{bulletblue}[sai-00171]} {\scriptsize\color{countgray}h=0 r=0} For deferred financing costs, carefully distinguish between gross and net carrying amounts: 'unamortized' costs refer to the net carrying amount after accumulated amortization, requiring ...
\item {\ttfamily\bfseries\color{bulletblue}[sai-00172]} {\scriptsize\color{countgray}h=0 r=0} For share-based compensation disclosures, critically analyze verb phrases to distinguish between similar numerical contexts: 'recognize over a period' indicates weighted-average recognition period ...
\item {\ttfamily\bfseries\color{bulletblue}[sai-00173]} {\scriptsize\color{countgray}h=0 r=0} When multiple numerical values appear in the same sentence context (e.g., '\$172.5 million aggregate principal amount of 3.25\% convertible senior notes'), carefully analyze which specific entity ...
\item {\ttfamily\bfseries\color{bulletblue}[sai-00174]} {\scriptsize\color{countgray}h=0 r=0} For debt issuance costs, critically distinguish between the initial cost amount (use DeferredFinanceCostsNet for the net carrying amount) and the periodic amortization expense (use ...
\item {\ttfamily\bfseries\color{bulletblue}[sai-00175]} {\scriptsize\color{countgray}h=0 r=1} For ownership percentages in consolidated subsidiaries, use MinorityInterestOwnershipPercentageByNoncontrollingOwners when the parent company consolidates the entity despite owning less than 100\%. ...
\item {\ttfamily\bfseries\color{bulletblue}[sai-00176]} {\scriptsize\color{countgray}h=0 r=0} For specific stock offering transactions (e.g., follow-on offerings, public offerings), prefer transaction-specific tags like SaleOfStockNumberOfSharesIssuedInTransaction and ...
\item {\ttfamily\bfseries\color{bulletblue}[sai-00178]} {\scriptsize\color{countgray}h=0 r=0} For ownership percentages, use EquityMethodInvestmentOwnershipPercentage for stakes typically between 20-50\% (indicating significant influence), and MinorityInterestOwnershipPercentageByParent for ...
\item {\ttfamily\bfseries\color{bulletblue}[sai-00179]} {\scriptsize\color{countgray}h=0 r=0} For tax rate reconciliation disclosures, carefully distinguish between reconciliation components and the final effective rate: use ...
\item {\ttfamily\bfseries\color{bulletblue}[sai-00180]} {\scriptsize\color{countgray}h=0 r=0} For loss contingency disclosures, critically distinguish between estimated ranges of possible outcomes and specific claimed amounts: use LossContingencyEstimateOfPossibleLoss for ranges of ...
\item {\ttfamily\bfseries\color{bulletblue}[sai-00181]} {\scriptsize\color{countgray}h=0 r=0} For intersegment revenues in segment reporting, use 'Revenues' rather than 'RevenueFromRelatedParties'. Intersegment revenues represent internal transactions between business segments and are ...
\item {\ttfamily\bfseries\color{bulletblue}[sai-00182]} {\scriptsize\color{countgray}h=0 r=0} For new stock issuances during a period (e.g., common stock offerings, equity raises), use 'StockIssuedDuringPeriodSharesNewIssues'. Reserve 'SaleOfStockNumberOfSharesIssuedInTransaction' for ...
\item {\ttfamily\bfseries\color{bulletblue}[sai-00184]} {\scriptsize\color{countgray}h=0 r=0} When the context explicitly states that operating segments 'represent our reportable segments' or similar phrasing that directly equates operating segments with reportable segments, use ...
\item {\ttfamily\bfseries\color{bulletblue}[sai-00185]} {\scriptsize\color{countgray}h=0 r=0} When term loans are presented as components of credit facilities within borrowing capacity context (e.g., 'Amended Credit Facilities... which includes the \$X million Term Loan Facility'), use ...
\item {\ttfamily\bfseries\color{bulletblue}[sai-00187]} {\scriptsize\color{countgray}h=0 r=0} For share-based compensation disclosures, critically distinguish between shares available within existing compensation arrangements and shares reserved from capital stock: use ...
\item {\ttfamily\bfseries\color{bulletblue}[sai-00188]} {\scriptsize\color{countgray}h=0 r=0} For business acquisition cash payments, use PaymentsToAcquireBusinessesGross when the context specifies cash paid at closing without mentioning netting against cash acquired (e.g., 'was paid at ...
\item {\ttfamily\bfseries\color{bulletblue}[sai-00189]} {\scriptsize\color{countgray}h=0 r=0} When identifying US GAAP tags for numerical entities, focus on the contextual meaning: property counts use NumberOfRealEstateProperties, share grant quantities use ...
\item {\ttfamily\bfseries\color{bulletblue}[sai-00190]} {\scriptsize\color{countgray}h=0 r=0} When identifying US GAAP tags for numerical entities, focus on the specific accounting context and purpose of the number rather than just its numerical value or general financial statement ...
\item {\ttfamily\bfseries\color{bulletblue}[sai-00191]} {\scriptsize\color{countgray}h=0 r=0} For intangible asset useful life disclosures, critically distinguish between general reporting (FiniteLivedIntangibleAssetUsefulLife) and business combination contexts: use ...
\item {\ttfamily\bfseries\color{bulletblue}[sai-00192]} {\scriptsize\color{countgray}h=0 r=0} For business acquisition cash payments, carefully distinguish between gross cash consideration and total consideration: use PaymentsToAcquireBusinessesGross for specific cash payments made to ...
\item {\ttfamily\bfseries\color{bulletblue}[sai-00193]} {\scriptsize\color{countgray}h=0 r=0} For credit arrangements described with 'not to exceed' language, this indicates maximum borrowing capacity rather than fixed principal amounts. Both revolving credit facilities and term loan ...
\item {\ttfamily\bfseries\color{bulletblue}[sai-00194]} {\scriptsize\color{countgray}h=0 r=2} For business acquisition contingent consideration, use BusinessCombinationContingentConsiderationLiability for both the fair value at acquisition date and subsequent measurement dates. The same ...
\item {\ttfamily\bfseries\color{bulletblue}[sai-00195]} {\scriptsize\color{countgray}h=0 r=0} In US GAAP financial reporting contexts, 'rent expense' disclosures typically refer to operating leases unless explicitly stated otherwise. Always prefer the more specific tag ...
\item {\ttfamily\bfseries\color{bulletblue}[sai-00196]} {\scriptsize\color{countgray}h=0 r=0} When backward references (e.g., 'such services', 'these transactions') clearly refer to previously described related party transactions, they should trigger RelatedPartyTransaction tags instead of ...
\item {\ttfamily\bfseries\color{bulletblue}[sai-00197]} {\scriptsize\color{countgray}h=1 r=0} For share-based compensation expense disclosures, verbs like 'recorded', 'recognized', or 'expensed' during a period specifically indicate expense allocation to the income statement, requiring ...
\item {\ttfamily\bfseries\color{bulletblue}[sai-00198]} {\scriptsize\color{countgray}h=1 r=0} For tax benefits specifically related to stock-based compensation arrangements (e.g., 'Income tax benefits related to stock-based compensation'), use ...
\item {\ttfamily\bfseries\color{bulletblue}[sai-00199]} {\scriptsize\color{countgray}h=0 r=0} When selecting between similar XBRL tags, prioritize the most specific tag that directly matches contextual language. If the context contains precise terminology (e.g., 'unused portion') that ...
\item {\ttfamily\bfseries\color{bulletblue}[sai-00200]} {\scriptsize\color{countgray}h=0 r=0} When multiple numerical values appear in close proximity, carefully analyze each value's specific context to determine the appropriate XBRL tag. For tax benefits, use ...
\item {\ttfamily\bfseries\color{bulletblue}[sai-00201]} {\scriptsize\color{countgray}h=0 r=0} For revenue recognition under ASC 606, critically distinguish between: 1) RevenueRemainingPerformanceObligation - for unsatisfied performance obligations that represent future revenue to be ...
\item {\ttfamily\bfseries\color{bulletblue}[sai-00202]} {\scriptsize\color{countgray}h=0 r=0} For interest expense amounts, default to the general InterestExpense tag unless the context explicitly requires distinguishing between different types of interest (e.g., debt interest vs. lease ...
\item {\ttfamily\bfseries\color{bulletblue}[sai-00203]} {\scriptsize\color{countgray}h=0 r=0} For contingent consideration liabilities, carefully distinguish between liability measurement contexts and accretion contexts: use BusinessCombinationContingentConsiderationLiability for the fair ...
\item {\ttfamily\bfseries\color{bulletblue}[sai-00204]} {\scriptsize\color{countgray}h=0 r=0} For stock issuance contexts, use SaleOfStockNumberOfSharesIssuedInTransaction when specific transaction details are present, including: named counterparties (e.g., 'Mill Road Capital'), explicit ...
\item {\ttfamily\bfseries\color{bulletblue}[sai-00205]} {\scriptsize\color{countgray}h=0 r=0} When encountering vague terms like 'these plans' without explicit specification, carefully analyze the broader context for clues about plan type. Annual expense amounts for employee benefit ...
\item {\ttfamily\bfseries\color{bulletblue}[sai-00206]} {\scriptsize\color{countgray}h=0 r=0} For business acquisition disclosures, critically distinguish between cash consideration and total consideration: use PaymentsToAcquireBusinessesGross when the context describes only cash payments ...
\item {\ttfamily\bfseries\color{bulletblue}[sai-00207]} {\scriptsize\color{countgray}h=0 r=0} When the context explicitly describes 'stock award expense' recognized in the income statement, use AllocatedShareBasedCompensationExpense rather than the general ShareBasedCompensation tag. The ...
\item {\ttfamily\bfseries\color{bulletblue}[sai-00208]} {\scriptsize\color{countgray}h=0 r=0} For unrecognized tax benefits, use UnrecognizedTaxBenefits when the context refers to the gross amount without any mention of impact on effective tax rate. Reserve ...
\item {\ttfamily\bfseries\color{bulletblue}[sai-00209]} {\scriptsize\color{countgray}h=0 r=0} For segment counts under ASC 280, use NumberOfReportableSegments when the context refers to the final aggregated segments presented in financial statements after applying quantitative thresholds ...
\item {\ttfamily\bfseries\color{bulletblue}[sai-00210]} {\scriptsize\color{countgray}h=0 r=0} Before selecting a US GAAP tag, always verify that the tag exists in the provided options list. If a more specific tag is recommended by context but not available, use the most appropriate general ...
\item {\ttfamily\bfseries\color{bulletblue}[sai-00212]} {\scriptsize\color{countgray}h=0 r=0} For share issuances during a period, default to StockIssuedDuringPeriodSharesNewIssues as the primary tag for all new shares issued, including those issued in business combinations. Reserve ...
\item {\ttfamily\bfseries\color{bulletblue}[sai-00213]} {\scriptsize\color{countgray}h=0 r=0} For share-based compensation, restricted stock units (RSUs) are treated identically to restricted stock awards as both are 'equity instruments other than options.' Use the same tags for RSUs as ...
\item {\ttfamily\bfseries\color{bulletblue}[sai-00214]} {\scriptsize\color{countgray}h=0 r=0} For business acquisition disclosures: use FiniteLivedIntangibleAssetUsefulLife for amortization periods of acquired intangible assets, Goodwill for recognized goodwill amounts from acquisitions, ...
\item {\ttfamily\bfseries\color{bulletblue}[sai-00215]} {\scriptsize\color{countgray}h=2 r=0} For debt issuance contexts, carefully distinguish between debt discounts and deferred financing costs: use DebtInstrumentUnamortizedDiscount for debt discounts (difference between face value and ...
\item {\ttfamily\bfseries\color{bulletblue}[sai-00216]} {\scriptsize\color{countgray}h=0 r=0} When analyzing debt-related contexts, critically distinguish between instrument characteristic tags (e.g., DebtInstrumentFaceAmount, DebtInstrumentUnamortizedDiscount) that describe features of ...
\item {\ttfamily\bfseries\color{bulletblue}[sai-00217]} {\scriptsize\color{countgray}h=0 r=0} For interest expense amounts, default to the general InterestExpense tag unless the context explicitly specifies debt instruments or requires distinction between different interest types. ...
\item {\ttfamily\bfseries\color{bulletblue}[sai-00218]} {\scriptsize\color{countgray}h=0 r=0} Before finalizing tag selection, always verify that the chosen tag exists in the provided options list. If a more specific tag is recommended by context but not available, select the most ...
\item {\ttfamily\bfseries\color{bulletblue}[sai-00219]} {\scriptsize\color{countgray}h=0 r=0} For share-based compensation and equity disclosures, critically distinguish between grant date measurements (e.g., weighted-average grant date fair value of options) and current market ...
\item {\ttfamily\bfseries\color{bulletblue}[sai-00220]} {\scriptsize\color{countgray}h=0 r=0} For warrants issued as compensation, treat them as share-based payment arrangements under ASC 718 rather than standalone derivatives. Use ...
\item {\ttfamily\bfseries\color{bulletblue}[sai-00221]} {\scriptsize\color{countgray}h=0 r=0} For tax authority claims involving unpaid taxes, penalties, and interest where the outcome is uncertain (e.g., show cause notices, regulatory assessments), use ...
\item {\ttfamily\bfseries\color{bulletblue}[sai-00222]} {\scriptsize\color{countgray}h=0 r=0} For proceeds from debt instrument issuances (including convertible notes, promissory notes, and other debt forms), always use 'DebtInstrumentFaceAmount' to represent the principal amount received. ...
\item {\ttfamily\bfseries\color{bulletblue}[sai-00224]} {\scriptsize\color{countgray}h=0 r=0} In US GAAP share-based compensation disclosures, the verb 'recognized' specifically indicates expense allocation to accounting periods, requiring the 'AllocatedShareBasedCompensationExpense' tag. ...
\item {\ttfamily\bfseries\color{bulletblue}[sai-00225]} {\scriptsize\color{countgray}h=0 r=0} For share-based compensation disclosures, carefully distinguish between vested fair value measurements ...
\item {\ttfamily\bfseries\color{bulletblue}[sai-00226]} {\scriptsize\color{countgray}h=0 r=0} For revolving credit facilities, carefully distinguish between initial establishment and ongoing characteristics: use DebtInstrumentFaceAmount for the original principal amount authorized when a ...
\item {\ttfamily\bfseries\color{bulletblue}[sai-00228]} {\scriptsize\color{countgray}h=0 r=0} For financing costs that are capitalized (added to an asset's value rather than expensed), use DeferredFinanceCostsNet regardless of whether the context mentions amortization. ...
\item {\ttfamily\bfseries\color{bulletblue}[sai-00229]} {\scriptsize\color{countgray}h=0 r=0} For share issuance disclosures, critically distinguish between general issuances (e.g., for compensation, acquisitions, conversions) and specific sale transactions (e.g., IPOs, secondary ...
\item {\ttfamily\bfseries\color{bulletblue}[sai-00230]} {\scriptsize\color{countgray}h=0 r=0} When distinguishing between LineOfCreditFacilityCurrentBorrowingCapacity and LineOfCreditFacilityMaximumBorrowingCapacity, carefully analyze verb tense and temporal context: past tense verbs like ...
\item {\ttfamily\bfseries\color{bulletblue}[sai-00231]} {\scriptsize\color{countgray}h=0 r=0} For convertible note transactions with Original Issue Discount (OID), consistently apply DebtInstrumentFaceAmount for the principal/face amount of consideration tranches and ...
\item {\ttfamily\bfseries\color{bulletblue}[sai-00232]} {\scriptsize\color{countgray}h=0 r=0} For revenue recognition under ASC 606, carefully distinguish between total revenue from contracts with customers and revenue recognized from contract liabilities: use ...
\item {\ttfamily\bfseries\color{bulletblue}[sai-00233]} {\scriptsize\color{countgray}h=0 r=0} For debt instruments, the original principal value is always tagged as DebtInstrumentFaceAmount, and the stated annual interest rate percentage is always tagged as ...
\item {\ttfamily\bfseries\color{bulletblue}[sai-00234]} {\scriptsize\color{countgray}h=0 r=0} For tax-related numerical entities, use unit context as a key differentiator: dollar amounts representing tax benefits or expenses require IncomeTaxExpenseBenefit, while percentage values ...
\item {\ttfamily\bfseries\color{bulletblue}[sai-00235]} {\scriptsize\color{countgray}h=0 r=0} For interest rate spreads on variable rate debt instruments, use DebtInstrumentBasisSpreadOnVariableRate1 for all margin components regardless of whether they represent minimum or maximum values ...
\item {\ttfamily\bfseries\color{bulletblue}[sai-00236]} {\scriptsize\color{countgray}h=0 r=0} For term loans, prioritize LongTermDebt as the primary tag for the outstanding amount on the balance sheet, as this represents the standard classification for aggregated term debt. Reserve ...
\item {\ttfamily\bfseries\color{bulletblue}[sai-00237]} {\scriptsize\color{countgray}h=0 r=0} For business acquisition disclosures, critically distinguish between cash consideration and total consideration: use PaymentsToAcquireBusinessesGross for specific cash payments made to acquire ...
\item {\ttfamily\bfseries\color{bulletblue}[sai-00238]} {\scriptsize\color{countgray}h=0 r=0} For ownership percentages, apply a clear threshold-based approach: use EquityMethodInvestmentOwnershipPercentage for investments with significant influence (typically 20-50\% ownership), and ...
\item {\ttfamily\bfseries\color{bulletblue}[sai-00239]} {\scriptsize\color{countgray}h=0 r=0} When interpreting share-based compensation plan descriptions, carefully analyze phrases like 'authorized for awards that may be granted' - this typically refers to shares currently available for ...
\item {\ttfamily\bfseries\color{bulletblue}[sai-00240]} {\scriptsize\color{countgray}h=0 r=0} For shares reserved for future issuance under stock plans, prefer the broader tag 'CommonStockCapitalSharesReservedForFutureIssuance' over ...
\item {\ttfamily\bfseries\color{bulletblue}[sai-00241]} {\scriptsize\color{countgray}h=0 r=0} For credit facility disclosures, recognize that 'available borrowing capacity' consistently refers to the remaining unused portion after deducting outstanding borrowings and letters of credit, ...
\item {\ttfamily\bfseries\color{bulletblue}[sai-00242]} {\scriptsize\color{countgray}h=0 r=0} When bank guarantees are issued under a credit facility as part of its components (alongside borrowings and available capacity), use LineOfCredit for the guarantee amount rather than ...
\item {\ttfamily\bfseries\color{bulletblue}[sai-00243]} {\scriptsize\color{countgray}h=0 r=0} For share price disclosures, prefer the general 'SharePrice' tag for per-share price measurements in various contexts including stock issuances, rather than transaction-specific tags like ...
\item {\ttfamily\bfseries\color{bulletblue}[sai-00244]} {\scriptsize\color{countgray}h=0 r=0} For share repurchase disclosures, critically distinguish temporal context: use StockRepurchaseProgramAuthorizedAmount1 for authorized amounts for future repurchases (e.g., 'entered into an ASR to ...
\item {\ttfamily\bfseries\color{bulletblue}[sai-00245]} {\scriptsize\color{countgray}h=0 r=0} For commitment fees on credit facilities, prefer the general LineOfCreditFacilityCommitmentFeePercentage tag as the standard choice, even when the fee calculation is based on unused amounts. The ...
\end{itemize}
\pbsection{COMMON MISTAKES TO AVOID}
\begin{itemize}[leftmargin=1.2em, itemsep=1pt, parsep=0pt, topsep=2pt]
\item {\ttfamily\bfseries\color{bulletblue}[err-00048]} {\scriptsize\color{countgray}h=0 r=0} Avoid using incomplete or abbreviated tag names. Always match the exact tag format from the provided US GAAP taxonomy list, including full compound words and proper suffixes (e.g., ...
\item {\ttfamily\bfseries\color{bulletblue}[err-00055]} {\scriptsize\color{countgray}h=3 r=2} Avoid selecting general tags when more specific tags exist that exactly match the context. For operating lease rent expenses, using OperatingLeaseExpense instead of OperatingLeasesRentExpenseNet ...
\item {\ttfamily\bfseries\color{bulletblue}[err-00059]} {\scriptsize\color{countgray}h=0 r=0} Avoid over-specifying tags when the context does not explicitly justify it. For example, prefer 'LeaseAndRentalExpense' for general rent expense contexts unless the sentence explicitly ...
\item {\ttfamily\bfseries\color{bulletblue}[err-00061]} {\scriptsize\color{countgray}h=0 r=0} Avoid treating derivative instruments as independent when their terms are directly derived from and identical to characteristics of an underlying primary financial instrument. For example, when ...
\item {\ttfamily\bfseries\color{bulletblue}[err-00079]} {\scriptsize\color{countgray}h=0 r=0} Avoid assuming that dollar amounts associated with fair value hierarchy disclosures (Level 1, 2, or 3) automatically represent fair values. Balance sheet amounts at specific dates typically ...
\item {\ttfamily\bfseries\color{bulletblue}[err-00080]} {\scriptsize\color{countgray}h=4 r=0} Avoid using DebtInstrumentFaceAmount for revolving credit facilities. The term 'revolving credit facility' specifically indicates a line of credit arrangement that requires ...
\item {\ttfamily\bfseries\color{bulletblue}[err-00087]} {\scriptsize\color{countgray}h=0 r=0} Avoid treating depreciation within restructuring contexts as regular depreciation. When depreciation is part of estimated restructuring program costs (e.g., 'Non-cash accelerated depreciation of ...
\item {\ttfamily\bfseries\color{bulletblue}[err-00094]} {\scriptsize\color{countgray}h=0 r=0} Avoid using OperatingLeasesRentExpenseNet when the context describes cash payments ('paid in rent') rather than expense recognition. The phrase 'paid...in rent' indicates cash outflow for rental ...
\item {\ttfamily\bfseries\color{bulletblue}[err-00103]} {\scriptsize\color{countgray}h=0 r=1} Avoid using DebtInstrumentFaceAmount for repayment transactions. The phrase 'principal repayment of \$X' describes an action (repayment transaction) requiring RepaymentsOfDebt, not a measurement of ...
\item {\ttfamily\bfseries\color{bulletblue}[err-00105]} {\scriptsize\color{countgray}h=0 r=2} Avoid using BusinessAcquisitionPercentageOfVotingInterestsAcquired for total ownership percentages in subsidiaries or joint ventures. This tag specifically applies to the percentage acquired in a ...
\item {\ttfamily\bfseries\color{bulletblue}[err-00107]} {\scriptsize\color{countgray}h=0 r=0} Avoid misclassifying balance sheet amounts as transactional events. When amounts represent outstanding balances or carrying values reported on the balance sheet (e.g., 'outstanding term loan of ...
\item {\ttfamily\bfseries\color{bulletblue}[err-00130]} {\scriptsize\color{countgray}h=0 r=2} Avoid using generic share-based compensation tags when more specific tags exist that distinguish between different types of equity instruments. For restricted share units (RSUs) and other equity ...
\item {\ttfamily\bfseries\color{bulletblue}[err-00133]} {\scriptsize\color{countgray}h=0 r=0} Avoid using share-based compensation tags (e.g., ShareBasedCompensationArrangementByShareBasedPaymentAwardOptionsGrantsInPeriodGross) for warrant quantity disclosures. Warrants are distinct equity ...
\item {\ttfamily\bfseries\color{bulletblue}[err-00164]} {\scriptsize\color{countgray}h=0 r=0} Avoid using tags that are not present in the provided US GAAP taxonomy list, even if they seem semantically correct based on context. Always verify that the selected tag exists in the available ...
\item {\ttfamily\bfseries\color{bulletblue}[err-00166]} {\scriptsize\color{countgray}h=0 r=0} Avoid assuming all ownership percentages in collaborative arrangements are equity method investments. For consortium structures described with ownership percentages below 50\%, do not default to ...
\item {\ttfamily\bfseries\color{bulletblue}[err-00168]} {\scriptsize\color{countgray}h=0 r=0} Avoid using 'CommonStockCapitalSharesReservedForFutureIssuance' for Employee Stock Purchase Plan (ESPP) shares. ESPP shares represent share-based compensation arrangements and should use ...
\item {\ttfamily\bfseries\color{bulletblue}[err-00177]} {\scriptsize\color{countgray}h=0 r=0} Avoid using EquityMethodInvestmentOwnershipPercentage for consolidated subsidiaries. When an entity is described as a subsidiary and consolidated in financial statements, the non-controlling ...
\item {\ttfamily\bfseries\color{bulletblue}[err-00183]} {\scriptsize\color{countgray}h=0 r=0} Avoid using 'SaleOfStockNumberOfSharesIssuedInTransaction' for new stock issuances that increase outstanding shares. This tag is more appropriate for treasury stock sales or specific disposal ...
\item {\ttfamily\bfseries\color{bulletblue}[err-00186]} {\scriptsize\color{countgray}h=0 r=0} Avoid automatically using DebtInstrumentFaceAmount for term loan amounts when they are presented as components within credit facility structures described with borrowing capacity language. Even ...
\item {\ttfamily\bfseries\color{bulletblue}[err-00211]} {\scriptsize\color{countgray}h=0 r=0} Avoid selecting US GAAP tags that are not present in the provided options list, even if they seem conceptually correct based on context or playbook recommendations. Always verify tag availability ...
\item {\ttfamily\bfseries\color{bulletblue}[err-00227]} {\scriptsize\color{countgray}h=0 r=0} Avoid applying absolute prohibitions against using DebtInstrumentFaceAmount for revolving credit facilities. While LineOfCreditFacilityMaximumBorrowingCapacity is appropriate for describing the ...
\item {\ttfamily\bfseries\color{bulletblue}[err-00246]} {\scriptsize\color{countgray}h=0 r=0} Avoid using LineOfCreditFacilityUnusedCapacityCommitmentFeePercentage for commitment fees unless the context explicitly requires distinguishing between fees on the entire facility versus only the ...
\end{itemize}
\pbsection{OTHERS}
\begin{itemize}[leftmargin=1.2em, itemsep=1pt, parsep=0pt, topsep=2pt]
\item {\ttfamily\bfseries\color{bulletblue}[misc-00001]} {\scriptsize\color{countgray}h=2 r=4} When shares are issued to acquire ownership interests in another entity (e.g., in an exchange offer), use BusinessAcquisitionEquityInterestsIssuedOrIssuableNumberOfSharesIssued instead of ...
\item {\ttfamily\bfseries\color{bulletblue}[misc-00002]} {\scriptsize\color{countgray}h=6 r=0} For cash payments in business acquisitions, use PaymentsToAcquireBusinessesGross when the context specifies 'cash paid' or 'purchase price' without mentioning netting against cash acquired. ...
\item {\ttfamily\bfseries\color{bulletblue}[misc-00003]} {\scriptsize\color{countgray}h=3 r=0} In business combination transactions, carefully analyze the form of consideration: equity interests issued, cash payments (gross vs net), and total consideration. Each has distinct tags - equity ...
\item {\ttfamily\bfseries\color{bulletblue}[misc-00004]} {\scriptsize\color{countgray}h=0 r=0} For mandatory convertible preferred stock and other hybrid instruments, use tags specific to convertible features (e.g., DebtInstrumentConvertibleConversionPrice1) rather than simple stock sale ...
\item {\ttfamily\bfseries\color{bulletblue}[misc-00005]} {\scriptsize\color{countgray}h=0 r=0} When dealing with preferred stock issuances, use broader stock issuance tags like StockIssuedDuringPeriodSharesNewIssues instead of SaleOfStockNumberOfSharesIssuedInTransaction, which is typically ...
\item {\ttfamily\bfseries\color{bulletblue}[misc-00057]} {\scriptsize\color{countgray}h=6 r=0} For amortization expense related to intangible assets, always use 'AmortizationOfIntangibleAssets' as the appropriate US GAAP tag. This tag specifically applies to the periodic expense recognition ...
\item {\ttfamily\bfseries\color{bulletblue}[misc-00058]} {\scriptsize\color{countgray}h=2 r=2} For shares available for issuance under share-based compensation plans, always use 'ShareBasedCompensationArrangementByShareBasedPaymentAwardNumberOfSharesAvailableForGrant'. This tag specifically ...
\item {\ttfamily\bfseries\color{bulletblue}[misc-00084]} {\scriptsize\color{countgray}h=1 r=0} For pension and benefit plan disclosures, critically distinguish between Defined Contribution (DC) and Defined Benefit (DB) plans as they have fundamentally different accounting treatments and ...
\item {\ttfamily\bfseries\color{bulletblue}[misc-00085]} {\scriptsize\color{countgray}h=1 r=0} Avoid using DefinedBenefitPlanContributionsByEmployer for Defined Contribution (DC) plan contexts. These are distinct plan types with separate accounting treatments and tags. The tag ...
\item {\ttfamily\bfseries\color{bulletblue}[misc-00223]} {\scriptsize\color{countgray}h=0 r=0} When multiple debt-related measurements appear together in disclosures, systematically distinguish between: (1) 'outstanding balance' or 'principal balance' at a point in time indicating carrying ...
\end{itemize}
\end{playbookbox}
\vspace{8pt}